\documentclass[reprint,twocolumn,aps,secnumarabic,nobibnotes,prx,showpacs,preprintnumbers,amsmath,amssymb,longbibliography,nofootinbib]{revtex4-1}
  
\usepackage[T1]{fontenc}    % use 8-bit T1 fonts
\usepackage{bm}
\usepackage{color}
\usepackage{xcolor}
\usepackage{empheq}
\usepackage{amsfonts}       % blackboard math symbols
\usepackage{nicefrac}       % compact symbols for 1/2, etc.
\usepackage{microtype}      % microtypography
\usepackage{url}            % simple URL typesetting
\usepackage{amssymb}
\usepackage{algorithm}
\usepackage{algorithmic}
\usepackage{newfloat}
\usepackage{afterpage}
\usepackage{citesort}
\usepackage{soul}
\usepackage{localmacros}        
\usepackage{pstricks}
\usepackage{pstricks-add}
\usepackage{float}
\usepackage{graphicx}
\usepackage[all]{nowidow}
\makeatletter\providecommand\href@noop{\@secondoftwo}\makeatother
\begin{document}

\title{A Deterministic and Generalized Framework for 
       Unsupervised Learning with
       Restricted Boltzmann Machines}

%%%%%%%% Author Block %%%%%%%%
\author{Eric W. Tramel}
  \altaffiliation{OWKIN, Inc., New York, New York, USA}
\author{Marylou Gabri\'e}
  \altaffiliation{Laboratoire de Physique Statistique,
                  \'Ecole Normale Sup\'erieure,
                  PSL University, Paris, France}
\author{Andre Manoel}
  \altaffiliation{OWKIN, Inc., New York, New York, USA}
\author{Francesco Caltagirone}
  \altaffiliation{Snips, Paris, France}
\author{Florent Krzakala}
  \altaffiliation{Laboratoire de Physique Statistique,
                  \'Ecole Normale Sup\'erieure,
                  PSL University, Paris, France}
  \altaffiliation{Universit\'e Pierre et Marie Curie,
                  Sorbonne Universit\'es, Paris, France}
  \altaffiliation{LightOn, Paris, France}
%%%%%%%%%%%%%%%%%%%%%%%%%%%%%%

\date{\today}

\begin{abstract}%
    Restricted Boltzmann machines (RBMs) are energy-based neural-networks
    which are commonly used as the building blocks for deep architectures
    neural architectures. In this work, we derive a deterministic
    framework for the training, evaluation, and use of RBMs 
    based upon the Thouless-Anderson-Palmer (TAP)
    mean-field approximation of widely-connected systems with weak
    interactions coming from spin-glass theory.  While the TAP approach
    has been extensively studied for fully-visible binary spin systems,
    our construction is generalized to latent-variable models, as well as
    to arbitrarily distributed real-valued spin systems with bounded
    support. In our numerical experiments, we demonstrate the effective 
    deterministic training of our proposed models and are able to show interesting
    features of unsupervised learning 
    which could not be directly observed with sampling.
    Additionally, we demonstrate how to utilize our TAP-based framework 
    for leveraging trained RBMs as joint priors in denoising problems.
\end{abstract}

\pacs{05.10.-a,05.90.+m }

\maketitle

%%%%%%%% Start Sections %%%%%%%%
\section{Introduction}
\label{sec:intro}
The past decade has witnessed a groundswell of research in machine learning,
bolstered by the deep learning revolution and the resurgence of neural 
networks \cite{CBH2015}. Since their inception, researchers have identified the
deep connection between neural networks and statistical mechanics. Perhaps the 
most well-known unsupervised neural models studied through the lens of statistical physics
have been the the Hopfield model \cite{Hop1982,AGS1985}
and the Boltzmann machine \cite{ackley1985learning}. These models were proposed 
from a connectionist perspective of cognitive science and were studied in the 
context emergent representation in unsupervised machine learning. 

We can look to the Hopfield model to directly observe some of the contributions
of physics to both machine learning and cognitive sciences. For example,
by applying techniques from the study of spin-glasses, Amit \emph{et al.} 
\cite{AGS1985} were famously able to derive the memory capacity of the Hopfield
model and provide a concrete understanding of the dynamics of the model via
the study of its phase transitions. This fundamental understanding of the 
behavior of the Hopfield model has provided insight into the complexities of
associative memory.

The closely related Boltzmann Machine is an undirected stochastic neural network 
which finds its physics parallel in Ising spin glass models \cite{MPV1986}.
Specifically, for this model, one is interested in the \emph{inverse}
problem: \emph{learning} the couplings between spins in order to generate a 
particular set of configurations at equilibrium. The process of learning
couplings, or \emph{training}, is often referred to as the inverse Ising 
problem in the physics literature 
\cite{sessak2009small,ricci2012bethe,ekeberg2013improved}. However, because 
couplings only exist between pairs of spins for the fully-visible Ising 
spin-glass, such models have limited practical application as they cannot 
successfully capture higher-order correlations which might exist in a set of 
training configurations. 

For this reason, the general Boltzmann machine
introduces a set of unobserved \emph{latent} spins. The effect of these 
latent spins is to abstract high-order correlations within the set of observed 
spins. While an optimal training of the 
couplings would
potentially lead to a very effective general model of high-dimensional joint
distributions, the intractability of this joint latent model confounds the 
practical application of general Boltzmann machines.

A {\it restricted} Boltzmann Machine (RBM) is a special case of the general
Boltzmann machine, where couplings only exist between latent and observed spins.
This bipartite structure is key to the efficient and effective training
of RBMs \cite{Hin2002}. 
RBMs have found many applications in machine
learning problems as diverse as dimensionality reduction
\cite{HS2006}, classification \cite{LB2008}, collaborative
filtering \cite{SMH2007}, feature learning \cite{CNL2011}, and topic
modeling \cite{HS2009}. Additionally, RBMs can be stacked into multi-layer neural
networks, which have played a historically fundamental role in 
pre-training deep network architectures \cite{HOT2006,SL2010}. These 
constructions, known as deep belief networks,
were the first truly \emph{deep} neural architectures, leading
to the current explosion of activity in deep
learning \cite{deepLearnBook}. 
While access to vast training datasets has made such pre-training
dispensable for certain tasks, RBMs remain a fundamental tool in the theory
of unsupervised learning. As such, a better understanding of RBMs
can be key to future developments in emergent machine intelligence.

To date, the most popular and effective approaches to training RBMs have centered
on differing flavors of short-chain Monte Carlo sampling \cite{Hin2002,Tie2008}, 
which we cover in detail in the sequel. While such techniques can yield trained
RBMs which produce sampled configurations very similar to the target dataset,
and can be used in a number of applications as detailed previously,
they do not bridge the gap in understanding \emph{what} the RBM has learned. 
Furthermore, understanding the modes, or internal representations, 
of the RBM with sampling-based frameworks have mostly consisted of subjective
comparisons of sampled configurations as well as a subjective analysis of the
couplings themselves, often referred to as \emph{receptive fields} in the 
machine learning literature.  

Additionally, comparing two trained models, or even monitoring the training of 
one model, becomes problematic when using sampling-based investigative tools. 
For example, annealed techniques \cite{Nea2001} can provide estimates of the 
log-likelihood of a model, but only at a large computational cost
\cite{SM2008,BGS2015}. 
% At a much
% lower computational cost, pseudo-likelihoods can be used to monitor training,
% but the estimates produced in this manner can be misleading \cite{SMB2010}.
At a much lower computational cost, pseudo-likelihoods can be used to monitor training,
but the estimates produced in this manner are inaccurate, as compared
to annealed importance sampling (AIS) \cite{Nea2001}, 
and even AIS can fail to detect model divergence in practice \cite{SMB2010}.

In the present work, we seek to address these concerns by developing a 
deterministic framework to train, compare, and analyze RBMs, as well as to
leverage their modeling power for inference tasks.
We accomplish this via statistical physics techniques 
through the use of the Thouless-Anderson-Palmer (TAP) formalism of spin-glass theory \cite{TAP1977,MPV1986,OW1997,BBOY2016}.
In this manner, we produce a model which no longer refers to a stochastic model
possessing an intractable Gibbs measure, but to a \emph{TAP machine}: 
an entirely self-consistent mean-field model which operates as a classical
RBM, but which admits deeper introspection via deterministic inference.
TAP machines also naturally handle non-binary variables as well as
deep architectures. While Deep Boltzmann Machines' (DBMs) \cite{SH2009} 
state-of-the-art training algorithms mix both Monte Carlo sampling and 
``na{\"i}ve'' mean-field approximation,  a deep TAP machine relies entirely on
the TAP mean-field approximation.

Under this interpretation, a TAP machine is not
a generative probabilistic model, but a deterministic model
defining a set of representational magnetizations for a given training
dataset. Advantageously, this learning output can be computed exactly
in finite time by converging a fixed-point iteration, in contrast to the
indeterminate stopping criterion of Markov-chain Monte Carlo sampling. 
This is a major distinction between the TAP machine and 
the classical RBM, for which the true probability density function is 
intractable. At its core, the TAP machine training consists of arranging the
minima, \emph{solutions}, 
in the proposed TAP-approximated free energy so as to maximize the correlation
between these solutions and the dataset. 
In our experiments, we demonstrate how
to track the growth and geometry of these solutions as a novel way to investigate
the progress of unsupervised learning. We also show how to use a trained 
TAP machine as a prior for inference tasks.

The paper is organized as follows. In Sec. \ref{sec:rbm} we formally describe 
the classical binary RBM and review the literature on RBM training and analysis.
Subsequently, in Sec. \ref{sec:grbm}, we describe our proposed modification of 
the binary RBM to a model with arbitrary real-valued distributions with
bounded support. Next, in
Sec. \ref{sec:abp}, we briefly describe how to apply belief-propagation to
perform inference in the setting of real-valued spins. The details of this
approach are pedagogically described in Appendices \ref{sec:bp} \& 
\ref{sec:rbp}. In Sec. \ref{sec:tap} we derive the TAP approximation of the
real-valued RBM via a high-temperature expansion of a two-moment Gibbs
free energy. Then, in Sec. \ref{sec:implementation}, we detail how to convert 
this approximation to a practical training algorithm. In Sec. 
\ref{sec:experiments}, we conduct a series of experiments on real datasets,
demonstrating how to use the properties of the TAP machine interpretation to 
provide insight into the unsupervised learning process. We additionally show
how to use a trained model for bit-flip correction as a simple example of
leveraging a TAP machine for inference tasks. Lastly, in Appendix 
\ref{sec:distributions}, we detail the derivations of necessary 
distribution-specific functions.

\section{Restricted Boltzmann Machines}
\label{sec:rbm}
Restricted Boltzmann Machines (RBMs) \cite{Smo1986} are latent-variable 
generative models often used in the context of unsupervised learning.
A set of weights and biases, the model parameters of the RBM, which correspond
to the \emph{couplings} and \emph{local fields} present in the system, 
constructs an energy as a function of the data points from which follows a 
Gibbs-Boltzmann probability density function. In the well-known \emph{binary}
RBM, for which all visible and latent variables are in $\bra{0,1}$, the RBM 
distribution is
\begin{equation}
    P(\mathbf{x}; W, \boldsymbol{\theta}) =
    \frac{1}{\mathcal{Z}\s{W,\boldsymbol{\theta}}}\sum_{\mathbf{h}} e^{\sum_{ij} \Wij x_i h_j + \sum_i b_i x_i + \sum_j c_j h_j}
    \label{eq:brbm}
\end{equation}
where $\boldsymbol{\theta} = \bra{\mathbf{b},\mathbf{c}}$ is the set of 
\emph{local potentials},
i.e. the set of values which define the biases
acting on each variable,
and 
\begin{equation}
    \mathcal{Z}\s{W,\boldsymbol{\theta}} = \sum_{\mathbf{x}}\sum_{\mathbf{h}}
        e^{\sum_{ij} \Wij x_i h_j + \sum_i b_i x_i + \sum_j c_j h_j}.
    \label{eq:rbmPartition}
\end{equation}
Here, we use the notation $\sum_{\mathbf{x}}$ and $\sum_{\mathbf{h}}$ to
refer to sums over the entire space of possible configurations of visible
and latent variables, respectively.
When taken with respect to the parameters of the model, 
$\mathcal{Z}\s{W,\boldsymbol{\theta}}$
is known as the 
\emph{partition function}. We give a factor-graph representation of the RBM
distribution in Fig. \ref{fig:visFactorGraph}. 

As evidenced by \eqref{eq:rbmPartition}, 
an exact computation of the normalizing partition function, and thus the
probability of a given high-dimensional data point, is inaccessible in practice.
Sophisticated Monte Carlo (MC) schemes relying on importance sampling 
\cite{SM2008,BGS2015} can produce estimates and bounds of the 
partition, but at the cost of substantial computation, running on 
the time scale of days or even weeks. 

Thankfully, a precise estimate of the normalization is unnecessary for many
RBM applications. Additionally, 
the bipartite structure of the RBM, which only admits couplings between
the hidden and visible variables, can be leveraged to construct efficient
sampling schemes. This approach was demonstrated in the 
contrastive divergence (CD) of \cite{Hin2002}, where very short-chain 
block-Gibbs sampling was shown to be sufficient for adequate RBM training. 
The CD approach consists of a sampling chain alternating between samples drawn
from the conditional probabilities of each layer, which are dependent on the 
conditional expectations at the previously sampled layer.
Specifically, the conditional probabilities for the hidden and visible units 
factorize as 
\begin{align}
    P(\mathbf{x}|\mathbf{h}) &= \prod_i {\rm sigm}\p{b_i + \sum_j \Wij h_j},\\
    P(\mathbf{h}|\mathbf{x}) &= \prod_j {\rm sigm}\p{c_j + \sum_i \Wij x_i},
\end{align}
where ${\rm sigm}(y) = \p{1 + e^{-y}}^{-1}$ is the logistic sigmoid function.
\def\svgwidth{0.5\textwidth}
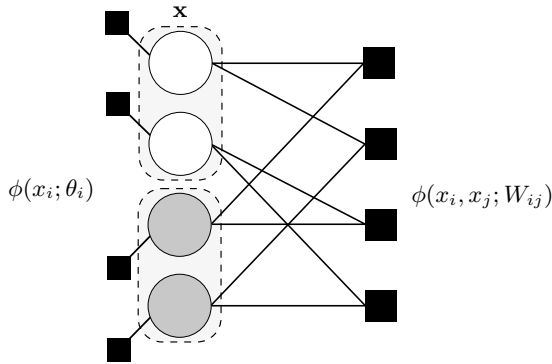
\begin{figure}[t!]
    \begin{center}
              \begingroup%
              \makeatletter%
              \providecommand\color[2][]{%
                \errmessage{(Inkscape) Color is used for the text in Inkscape, but the package `color.sty' is not loaded}%
                \renewcommand\color[2][]{}%
              }%
              \providecommand\transparent[1]{%
                \errmessage{(Inkscape) Transparency is used (non-zero) for the text in Inkscape, but the package `transparent.sty' is not loaded}%
                \renewcommand\transparent[1]{}%
              }%
              \providecommand\rotatebox[2]{#2}%
              \ifx\svgwidth\undefined%
                \setlength{\unitlength}{341.67988281bp}%
                \ifx\svgscale\undefined%
                  \relax%
                \else%
                  \setlength{\unitlength}{\unitlength * \real{\svgscale}}%
                \fi%
              \else%
                \setlength{\unitlength}{\svgwidth}%
              \fi%
              \global\let\svgwidth\undefined%
              \global\let\svgscale\undefined%
              \makeatother%
              \begin{picture}(1,0.58345911)%
                \put(0,0){\includegraphics[width=\unitlength,page=1]{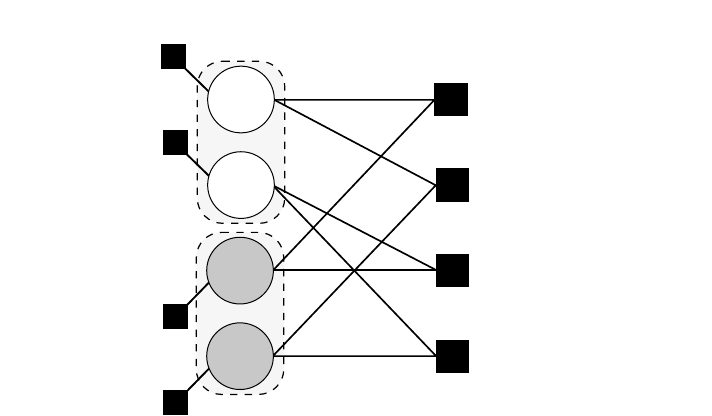}}%
                \put(0.33831966,0.51230333){\color[rgb]{0,0,0}\makebox(0,0)[b]{$\mathbf{x}$}}%
                \put(0.21136378,0.23754714){\color[rgb]{0,0,0}\makebox(0,0)[rb]{$\phi(x_i;\theta_i)$}}%
                \put(0.68103368,0.22846975){\color[rgb]{0,0,0}\makebox(0,0)[lb]{$\phi(x_i,x_j;\Wij)$}}%
              \end{picture}%
            \endgroup%
            
        \caption{Factor graph representation of the RBM distribution. Variables
                 are indicated by circles, with latent variables denoted by 
                 shaded circles. The shaded
                 rectangles indicated layer partitions within the RBM structure. 
                 Factors are represented by squares, with the right hand side
                 factors representing the pairwise relationships between
                 variables and the left hand side factors representing the 
                 influence of the localized prior distributions on the variables.
                 \label{fig:visFactorGraph}}
    \end{center}    
\end{figure}

In order to learn the parameters of the RBM for a given training dataset, one
looks to maximize the following log-likelihood,
\begin{align}
    \ln P(\mathbf{x};W,\boldsymbol{\theta}) &= 
    \ln \sum_{\mathbf{h}} e^{\sum_{\ij}\Wij x_i h_j + \sum_i b_i x_i + \sum_j c_j h_j} \notag\\
    &\quad- 
     \ln \mathcal{Z}\s{W,\boldsymbol{\theta}},
    \label{eq:rbm-ll}
\end{align}
via gradient ascent on the parameters $W$ and $\boldsymbol{\theta}$.
Commonly, one does not calculate these gradients for each data-point from the training
set, but instead calculates the gradients in average across $M$ data-points,
often referred to as a \emph{mini-batch}. At each mini-batch, the gradients of
\eqref{eq:rbm-ll} are given as,
\begin{align}
    \Delta \Wij &= \ang{x_i h_j}_\mathbf{X} - \ang{x_i h_j}_{\rm Sampled}, \label{eq:cdgw}\\
    \Delta b_i &= \ang{x_i}_\mathbf{X} - \ang{x_i}_{\rm Sampled}, \label{eq:cdgb}\\
    \Delta c_j &= \ang{h_j}_\mathbf{X} - \ang{h_j}_{\rm Sampled}, \label{eq:cdgc}
\end{align}
where $\ang{\cdot}_{\rm Sampled}$ refers to averages over particles sampled from
the model, and $\ang{\cdot}_\mathbf{X}$ refers to the so-called \emph{clamped}
expectations, where the values of $\mathbf{x}$ are fixed to the training data
samples in the mini-batch. In the case of the expectations involving hidden
units, which are unobserved and therefore have no training data, \cite{Hin2002}
originally proposed the use of configurations sampled from 
$P(\mathbf{h}|\mathbf{x};W,\mathbf{b},\mathbf{c})$. 
However, one could also use the exact conditional expectations directly to 
calculate these clamped averages; especially in cases where sampling from 
these conditionals may be problematic.
 
Since \cite{Hin2002}, there have been a number of proposed modifications to the
core sampling-based training scheme described above.
The persistent trick \cite{Tie2008} takes neatly advantage of the
iterative gradient ascent over mini-batches 
to quickly obtain thermalized Markov chains through Gibbs 
sampling at no extra computational cost over one step CD (CD-1).
Nevertheless, the probability density function of a trained RBM is typically 
highly multimodal, thus making this sampling inexact.
Indeed, in such glassy landscapes mixing becomes very slow as Markov chains 
become stuck in metastable states, leading to over- and 
under-represented, as well as missed, modes of the high-dimensional distribution.
This in turn produces high variance estimates of means and correlations. 
A more accurate sampling can be achieved using parallel tempering
\cite{DCB2010,Sal2010}, where particles are swapped
between multiple Markov chains running at differing temperatures.
This approach, however, requires not only the additional computational burden
of running more chains, but also requires further tuning of hyper-parameters,
such as the number of chains and at which temperatures to run them.

As accurate sampling-based inference on RBMs can be costly, it would seem that
their usefulness is limited. As learning of the RBM via gradient ascent is 
dependent upon this inference, the difficulty of training a generative model 
with a high-degree of accuracy is compounded. However,
RBMs have proven to be very useful in many applications where sampling 
from a full-fledged generative model is unneeded. For instance, RBMs can be used
as an unsupervised ``feature extraction'' pre-training for feed-forward networks
\cite{HOT2006,HS2006,BLP2007}. 
RBMs have also been used for data-imputation tasks, e.g. image in-painting,
label recovery \cite{LB2008}, or collaborative filtering \cite{SMH2007} 
by reconstructing missing data with a single visible-hidden-visible step.
In truth, the CD-k training algorithm which popularized 
RBMs, first with binary units \cite{Hin2002}, then Gaussian units  
\cite{HS2006,Cho2011} and finally with arbitrary units \cite{WRH2004}, 
does not use thermalized samples to evaluate means and correlations. 
Instead, it focuses on the region of the configuration space nearest to the 
training dataset \cite{DCB2010} by using short block-Gibbs Markov chains, 
starting from training data points, to get fast and low variance estimates of 
moments. However, CD-k is prone to learn spurious minima in configuration space 
far from the data as it does not explore this region during training \cite{DCB2010}. 
It also does not systematically increase the true likelihood of training data 
\cite{FI2010}. However, this training strategy has been found to be very 
efficient in the applications mentioned above, which consistently remain close 
to the dataset in configuration space. One finds that CD falls short for applications
which require long-chain MCMC sampling from the trained RBM, as this represents
a fundamental mismatch between the training and application of the RBM. In 
order to address some of theses shortcomings of sampling-based approaches,
we now turn our attention to deterministic mean-field approximations of the 
RBM.

The TAP approximation \cite{TAP1977} for disordered systems relies on the 
deterministic inference of approximated magnetizations, from which one can 
obtain estimators of all kinds of observables, starting from the log-partition 
or \emph{free energy}. TAP is derived from a small weight expansion of the
variational approach and can be considered as an extension of the na{\"i}ve 
mean-field (NMF) method (see~\cite{Ple1982,GY1999} for the original derivation,
and \cite{yedidia2001idiosyncratic,zamponi2010mean} for pedagogical expositions).
Previous works which have attempted to make use of the NMF
approximation of the RBM have shown negative results \cite{WH2002,Tie2008}. 

The TAP approximation was first considered for Boltzmann machines in the
context of small random models without hidden units in \cite{KR1998}.
In the recent work of \cite{GTZ2015}, this approximation was extended to a
practical training algorithm for full-scale binary RBMs which was shown to be 
competitive with Persistent Contrastive Divergence (PCD) \cite{Tie2008} when applied to real-world datasets.
In parallel, other works have used TAP, and the related Bethe approximation, 
to perform inference on binary Boltzmann machines \cite{WT2003,TDK2016,Ric2012,Mez2016}. 

In the next sections, we detail how to re-write the RBM model in the non-binary
case, for generalized distributions on the visible \emph{and} hidden units,
similar in spirit to \cite{WRH2004}. However, unlike earlier techniques, we will
approach the problem of estimating the normalization of the RBM model via the
tools of statistical mechanics, resulting in a fully-deterministic framework for
RBM inference, training, and application.

\section{General Distributions for RBMs}
\label{sec:grbm}
We now turn our attention to the case of the general RBM (GRBM),
where the distributions of the hidden and visible units are not fixed. 
We define the distribution of interest in the following manner,
\begin{align}
P(\vecx, \vech;W,\allparams) &= 
    \frac{1}{\Z\s{W,\allparams}} e^{\sum_{i,j} x_i \Wij h_j}
      \notag\\ 
    &\quad\times \prod_{i}
     P_i^{\rm v}(x_i;\visparam{i}) 
    \prod_{j} 
    P_j^{\rm h}(h_j;\hidparam{j})
\label{eq:visgrbm}                          
\end{align}
where the sum over $i,j$ indicates a sum over the all visible and hidden 
units in the model, and 
$\allparams = \{\visparam{1}, \dots, \visparam{\numvis}, 
\hidparam{1}, \dots, \hidparam{\numhid}\}$ 
are the parameters defining the 
local distributions, $P_i^{\rm v}$ and $P_j^{\rm h}$, on respectively the variables 
of $\vecx$ and the variables of $\vech$.
In the case that $\vecx \in \{\pm 1\}^{N_{\rm v}}$ and 
$\vech \in \{\pm 1\}^{N_{\rm h}}$, we can see that the distribution 
above reduces to a bipartite spin glass  model with $\allparams$ representing the local fields acting on the 
system spins; the fields $\mathbf{b}$ and $\mathbf{c}$ for binary spins as
described in Eq. \eqref{eq:brbm}. 
This specific case is simply a binary RBM, as described
in the previous section, and which we have already considered within an
extended mean-field framework in \cite{GTZ2015}.
The important distinction with the model
we evaluate here is that we do not assume a binary discrete distribution on the
variables, but instead allow for a formulation where the variables in the system
can possess distributions on discrete- or real-valued bounded support. 
By considering this more general class of models, one can include a wide range
of different models, including the Hopfield model and spike-and-slab RBMs \cite{CBB2011},
and data sets, such as images or genomic data, by varying the distributions
of the hidden and visible units.
The distribution of
visible variables is obtained by marginalizing out the latent variables, 
\begin{equation}
    P(\mathbf{x};W,\allparams) =
    \int\p{\prod_{j} {\rm d}h_j}   P(\vecx, \vech;W,\allparams)  ,
\end{equation}
giving the log-probability 
\begin{align}
\ln P(\mathbf{x};&W,\allparams) = 
    -\ln \Z\s{W,\allparams} 
    + \sum_{i}\ln P_i(x_i;\visparam{i}) \notag\\
    &\quad+ \sum_{j}\ln \integ{h_j} P_j(h_j;\hidparam{j}) e^{h_j \sum_{l}W_{lj}x_l} .
    \label{eq:genloglike}
\end{align}

If we take the gradients of \eqref{eq:genloglike} with respect to the model
parameters, in the case of distribution terms $\allparams$ we find,
\begin{equation}
\Delta\visparam{i} = 
        \ang{\pde{\visparam{i}} \ln P_i(x_i;\visparam{i})}_{\data} 
        - \pde{\visparam{i}}\ln\Z\s{W,\allparams}, \label{eq:paramGradVis}
\end{equation}
and
\begin{equation}
\Delta\hidparam{j} = 
        \ang{\pde{\hidparam{j}} \ln P_j(h_j;\hidparam{j})}_{\data} 
        - \pde{\hidparam{j}}\ln\Z\s{W,\allparams}, \label{eq:paramGradHid}
\end{equation}
which are generalizations of \eqref{eq:cdgb} and  \eqref{eq:cdgc}.
However, in the case of the gradient with respect to the couplings, we find
\begin{equation}
    \Delta\Wij = \ang{x_i\cdot f\p{\sum_{l}W_{lj}x_l;\hidparam{j} }}_{\data} - \pde{\Wij}\ln\Z\s{W,\allparams}, \label{eq:Wgrad}\\
\end{equation}
where the function 
\begin{equation}
f(B;\theta) = \frac{\integ{h} h\cdot P(h;\theta) e^{Bh}}{\integ{h} P(h;\theta) e^{Bh}}     
\label{eq:condexp}
\end{equation}
computes the conditional expectation of $h_l$ knowing the value of the 
visible units. The one-dimensional integral in \eqref{eq:condexp}
can be computed either analytically or numerically.
Note moreover that the data-dependent
term is tractable thanks
to the bipartite structure of the one-hidden layer RBM.

In contrast to the data-dependent terms,
the second terms of Eqs. \eqref{eq:paramGradVis}--\eqref{eq:Wgrad} 
require knowledge
of the partials of the log normalization w.r.t. the parameter of interest. 
However, this term cannot be written exactly as the explicit calculation of the
normalization is intractable. 
Rather than resorting to sampling, 
we will attempt to approximate the \emph{free energy}
 $\F = -\ln\Z\s{W,\allparams}$
in a parametric and deterministic way, as in \cite{GTZ2015}.
In the next section, we discuss how belief propagation (BP) can be used to 
estimate $\F$ and to conduct inference on RBMs.

\section{Approximation via Belief Propagation}
\label{sec:abp}
One method by which we might estimate the partition of $\F$ is
via belief-propagation (BP) \cite{Pea1982}, which we review in Appendix
\ref{sec:bp} for pairwise models such as the RBM. 
Essentially, given a factor graph for some joint statistical 
model, such as that of our RBM in Fig. \ref{fig:visFactorGraph}, 
the BP algorithm attempts to estimate a set of marginal distributions at 
each variable. In the case of tree-like graphs, BP provides an exact
calculation of these marginals. 
The 
application of BP to factor graphs containing cycles, \emph{loopy} BP, is
not guaranteed to provide accurate estimates of the marginals. However, often
these estimated marginals have significant overlap with the true ones 
\cite{YFW2001}.
Additionally, it is known that the solutions of loopy BP are the fixed-points 
of the 
Bethe free energy \cite{YFW2001}, which allows the construction of an 
approximation of $\F$. Applying this to the inverse learning problem,
one can compute the gradients of this Bethe free energy in terms of the 
parameters of the model, allowing a gradient ascent on Bethe-approximated 
log-likelihood of the training data. 

One significant hurdle in the application of loopy BP to RBM learning
for real-valued variables is that the messages propagated on the edges of the
factor graph are continuous PDFs. In the case of discrete variables, such as
Ising or Potts spins, BP messages can be written using magnetizations or the 
full discrete PMF, respectively. For binary variables, both BP and mean-field 
approximations of fully-connected Boltzmann machines 
were considered in \cite{WT2003} in the context of
inference with fixed parameters. A similar study of binary RBMs 
was conducted with loopy BP in \cite{HT2015}. It is important to note that
both of these studies investigated the properties of Boltzmann machines
with \emph{i.i.d.} random weights. 
While such studies permit many analytical tools for 
studying the behavior of the RBM, one cannot directly map these observations
to RBM inference in practice, where trained 
weights may exhibit strong correlations both within and between receptive
fields. 

In order to construct a BP algorithm for PDFs over real-valued support, 
one requires a finite memory description of the messages. Some examples of 
such descriptions are given 
in non-parametric BP \cite{SIF2010}, moment matching \cite{OW2005}, 
and relaxed BP (r-BP) \cite{Ran2010}. In Appendix \ref{sec:rbp}, following
the example of r-BP, we show how to arrive at a two-moment approximation of 
the continuous BP messages via a small-weight expansion on the RBM coupling 
parameters $W$. There, we also
show the r-BP approximated free energy of pairwise models, as well as 
demonstrating the need for distributions with bounded support in order to 
preserve bounded messages. 

In the next section, building upon this derivation, 
we consider mean-field approximations of the RBM via high-temperature Plefka
expansion. 

\section{TAP Approximation for Pairwise Models}
\label{sec:tap}
While one could utilize the r-BP approach in order to estimate the 
free energy of a generalized real-valued spin model, as detailed in the
earlier section, such an approach might not be desirable in practice. 
Specifically, if one wishes to solve the inverse learning problem, 
estimating model parameters from a given dataset, it is necessary to
estimate the gradients of the model parameters w.r.t. the model likelihood
for each parameter update. Using a steepest-ascent approach, as detailed
in Sec. \ref{sec:grbm}, requires one to estimate these gradients many 
thousands of times. 
For systems of large size $N$, the r-BP scales quite poorly. Estimating
a gradient requires the iteration of the r-BP equations on $O(N^2)$ 
messages. Additionally, one must distinguish between cavity terms 
$(i\To j)$ and marginal terms $(\To i)$. If the final $O(N)$ 
gradients we 
desire can be estimated using the marginal terms alone, then 
requiring an iteration on the $O(N^2)$ set of messages is an extremely
costly operation. 

Instead, one can turn to a mean-field approach, writing the free
energy, and its stationary points, in terms
of the marginals alone. This can be done by including certain
correction terms, up to a specified degree in the weights. 
In the context of RBMs,
such approaches have been proposed at both the 1st order, the naive
mean-field \cite{WH2002}, and the 2nd order, using the so-called
Thouless-Anderson-Palmer (TAP) \cite{TAP1977} equations for introducing
an additional correction term
\cite{WT2003,lesieur2015phase,LKZ2015,Mez2016,GTZ2015}.  In the case
of a GRBM with arbitrary distributions on each unit, however, we must
re-derive the TAP approximation in terms of parameters of these
distributions as well as the approximate marginalized distribution at
each site, up to their first two moments. This task turns out to be
closely related to the TAP approach to low-rank matrix factorization
\cite{lesieur2015phase,LKZ2015,DM2014,SF2012,LKZ2017}.

While it is possible to derive the stationarity conditions for the
inferred marginals from the r-BP messages directly by Taylor
expansion, we rather focus on the free energy directly that will
provide the gradients we require for training the
GRBM parameters via a high-temperature expansion we present below.

%%% Apparently the \hl command doesn't like to play well with cite and ref commands
%
Lastly, we point out that the TAP free energy  second-order (TAP) term depends 
on the statistical properties of the weight distribution. The derivation presented 
below assumes independent identically distributed weights, scaling as $O(1/\sqrt{N})$. 
This assumption is a simplification, as in practice the weight distribution cannot
be known \emph{a priori}. The distribution depends on the training data and changes throughout
the learning process according to the training hyper-parameters. 
The adaptive TAP (adaTAP) formalism \cite{OW2001} 
attempts to correct this assumption by allowing one
to directly compute the ``correct'' second-order correction term 
for a realization of the $W$  matrix without any hypothesis on how its entries are 
distributed. Although this algorithm is the most principled approach, its computational 
complexity almost rules out its implementation. 
Moreover, practical learning experiments indicate that training using 
adaTAP does not differ significantly from TAP assuming i.i.d. weights. 
A more detailed discussion of the computational complexity and learning
performance is described in Appendix \ref{sec:adatap}.

\subsection{Derivation of the TAP Free Energy}
We now discuss the main steps of the derivation of the TAP free energy,
which was originally performed in \cite{Ple1982, GY1999}. We do not aim
to perform it in full detail; a more pedagogical and comprehensive
derivation can be found in the Appendix B of \cite{LKZ2017}.

In the limit $N\To \infty$, if we assume that the entries of $W$ 
scale as $O(1/\sqrt{N})$ and that all sites are widely connected, on the
order of the size of the system, then we can apply the TAP approximation
-- a high temperature expansion of the Gibbs free energy  
up to second-order \cite{Ple1982,GY1999}. In the case of a Boltzmann distribution,
the global minima of the Gibbs free energy, its value at equilibrium, matches
the Helmholtz free energy $\F$ \cite{Yed2000}. We will derive a two-variable
parameterization of the Gibbs free energy derived via the Legendre transform
\cite{WJ2008}.
Additionally, we will show that this two-variable Gibbs free energy
is both variational and attains the Helmholtz free energy at its 
minima. 
For clarity of notation,
we make our derivation in terms of a
pairwise interacting Hamiltonian without enforcing any specific structure on
the couplings; the bipartite structure of the RBM is reintroduced in
Section \ref{sec:implementation}.

We will first
introduce the inverse temperature term $\beta$ to facilitate our expansion as
$\beta \To 0$,
$P(\vecx;\beta,W,\allparams) = 
    e^{-\beta (\mathcal{H} - \F_\beta)}$,
where
\begin{equation}
    \mathcal{H} = -\sum_{(i,j)} \Wij x_i x_j - \frac{1}{\beta}\sum_i \ln P_i(x_i;\theta_i).
\end{equation}
Note that $\beta$ can be interpreted as the weight scaling, i.e.
one can rescale the weights so that $W \leftarrow \beta W$.

We wish to derive our two-variable Gibbs free energy for this system in terms of the 
first two moments of the marginal distributions at each site. To accomplish
this, we proceed as in \cite{GY1999,OW2001} by first defining an augmented system
under the effect of \emph{two} auxiliary fields,
\begin{equation}
    -\beta\F_\beta(\boldsymbol{\lambda},\boldsymbol{\xi})
    = 
    \ln \integ{\boldsymbol{x}} e^{-\beta\mathcal{H} + \sum_i \lambda_i x_i + \sum_i \xi_i x_i^2},
\end{equation}
where we see that as the fields disappear, 
$\F_\beta(\boldsymbol{0},\boldsymbol{0}) = \F_\beta$, and we recover the true Helmholtz
free energy. 

We additionally note the following identities for the augmented 
system, namely,
\begin{align}
    \pde{\lambda_i}\s{-\beta\F_\beta(\boldsymbol{\lambda},\boldsymbol{\xi})} &= 
    \ang{x_i}_{\boldsymbol{\lambda},\boldsymbol{\xi}},\\
    \pde{\xi_i}\s{-\beta\F_\beta(\boldsymbol{\lambda},\boldsymbol{\xi})} &= 
    \ang{x_i^2}_{\boldsymbol{\lambda},\boldsymbol{\xi}},
\end{align}
where $\ang{\cdot}_{\boldsymbol{\lambda},\boldsymbol{\xi}}$ is the average over 
the augmented system for the given auxiliary fields. Since the partial 
derivatives of the field-augmented Helmholtz free energy generate the 
cumulants of the Boltzmann distribution, it can be shown that the Hessian of 
$-\beta \F_\beta(\boldsymbol{\lambda},\boldsymbol{\xi})$ is simply a 
covariance matrix and, subsequently, positive semi-definite. Hence, 
$-\beta \F_\beta(\boldsymbol{\lambda},\boldsymbol{\xi})$ is a convex function 
in terms of $\boldsymbol{\lambda}\times\boldsymbol{\xi}$.
This convexity is shown to be true for all log partitions of exponential family
distributions in \cite{WJ2008}.

We now take the Legendre transform of 
$-\beta\F_\beta(\boldsymbol{\lambda},\boldsymbol{\xi})$, introducing the conjugate 
variables $\boldsymbol{a}$ and $\boldsymbol{c}$,
\begin{align}
    &-\beta \mathbb{G}_\beta(\boldsymbol{a},\boldsymbol{c}) =\notag\\
    &\quad
    -\beta \sup_{\boldsymbol{\lambda},\boldsymbol{\xi}}
    \bra{\F_\beta(\boldsymbol{\lambda},\boldsymbol{\xi})
         + \frac{1}{\beta}\sum_i \big[ \lambda_i a_i + \xi_i (c_i + a_i^2)
         \big]},
    \label{eq:legendreGibbs}         
\end{align}
where we define the solution of the auxiliary fields at which 
$\mathbb{G}_\beta(\boldsymbol{a},\boldsymbol{c})$ is defined as 
$\boldsymbol{\lambda}^* \defas \boldsymbol{\lambda}^*(\beta,\boldsymbol{a},\boldsymbol{c})$,
$\boldsymbol{\xi}^* \defas \boldsymbol{\xi}^*(\beta,\boldsymbol{a},\boldsymbol{c})$, where
we make explicit the dependence of the auxiliary field solutions on the values
of the conjugate variables. Looking at the stationary points of these 
auxiliary fields, we find that
\begin{align}
\pde{\lambda^*_i}\s{ -\beta \mathbb{G}_\beta(\boldsymbol{a},\boldsymbol{c})} &=
        \pde{\lambda^*_i}\s{-\beta\F\p{\boldsymbol{\lambda}^*,\boldsymbol{\xi}^*}}
        - a_i, \\
        \therefore a_i &= \ang{x_i}_{\boldsymbol{\lambda}^*,\boldsymbol{\xi}^*},
\end{align}
and 
\begin{align}
\pde{\xi^*_i}\s{ -\beta \mathbb{G}_\beta(\boldsymbol{a},\boldsymbol{c})} &=
        \pde{\xi^*_i}\s{-\beta \F\p{\boldsymbol{\lambda}^*,\boldsymbol{\xi}^*}}
        - (c_i + a_i^2), \\
        \therefore c_i &= \ang{x_i^2}_{\boldsymbol{\lambda}^*,\boldsymbol{\xi}^*} - \ang{x_i}^2_{\boldsymbol{\lambda}^*,\boldsymbol{\xi}^*}.
\end{align}
The implication of these identities is that 
$\mathbb{G}_\beta(\boldsymbol{a},\boldsymbol{c})$ cannot be valid unless it 
meets the self-consistency constraints that $\mathbf{a}$ and $\mathbf{c}$ are 
the first and second (central) moments of the marginal distributions of
the augmented system.

Now, we wish to show the correspondence of 
$\mathbb{G}_\beta(\boldsymbol{a},\boldsymbol{c})$
to the Helmholtz
free energy at its unique minimum. First, let us look at the stationary points of 
$-\beta \mathbb{G}_\beta(\boldsymbol{a},\boldsymbol{c})$ with respect to its 
parameters. We take the derivative with careful application of the chain rule to find,
\begin{align}
    &\pde{a_i}\s{-\beta\mathbb{G}_\beta(\boldsymbol{a},\boldsymbol{c})} =
    -\lambda_i^* - 2\xi_i^*a_i \, + \notag \\
    &\kern1em + \sum_j \frac{\partial}{\partial \lambda_j^\ast} [-\beta \mathbb{G}] \,
    \frac{\partial \lambda_j^\ast}{\partial a_i}
    + \sum_j \frac{\partial}{\partial \xi_j^\ast} [-\beta \mathbb{G}] \,
    \frac{\partial \xi_j^\ast}{\partial a_i} \notag \\
    &= -\lambda_i^* - 2\xi_i^*a_i,
\end{align}
with the terms inside the sums going to zero as the derivatives of
$\mathbb{G}$ with respect to $\lambda_j^\ast$ and $\xi_j^\ast$ are zero
by definition.
Carrying through a very similar computation for the derivative with respect to 
$c_i$ provides
$\pde{c_i}\s{-\beta\mathbb{G}_\beta(\boldsymbol{a},\boldsymbol{c})} = \xi_i^*$.
This shows that, at its solution, the Gibbs free energy must satisfy 
\begin{align}
    \xi_i^* &= 0, \notag\\
    \lambda_i^* + 2 \xi_i^* a_i &= 0,
    \label{eq:gibbs-soln}
\end{align}
which can only be true in the event that the solutions of the auxiliary fields
are truly $\boldsymbol{\lambda}^* = 0$ and $\boldsymbol{\xi}^* = 0$. 
Looking at the \emph{inverse} Legendre transform of the Gibbs free energy 
for $\boldsymbol{\lambda} = \boldsymbol{0}$, $\boldsymbol{\xi} = \boldsymbol{0}$, 
we find that 
$\inf_{\boldsymbol{a},\boldsymbol{c}} \mathbb{G}_\beta(\boldsymbol{a},\boldsymbol{c})
= \F_\beta\p{\boldsymbol{0},\boldsymbol{0}} = \F_\beta$, which implies that 
the minimum of the Gibbs free energy is equivalent to the Helmholtz free energy. 
This holds since $-\beta \mathbb{G}_\beta(\boldsymbol{a},\boldsymbol{c})$ is
convex, as the Legendre transform of a convex function is itself convex. 
Since the Gibbs free energy can therefore only possess a single solution,
then its minimum must satisfy \eqref{eq:gibbs-soln}, and therefore, must be
$\F_\beta\p{\boldsymbol{0},\boldsymbol{0}} = \F_\beta$.

Finally, 
we can now rewrite the Gibbs free energy defined in 
Eq. \eqref{eq:legendreGibbs} as a function of the moments 
$\alla$, $\allc$ and parameterized by $\beta$ and the GRBM parameters
$W$ and $\allparams$,
\begin{equation}
-\beta\gibbs_\beta(\alla,\allc;\allparams) = 
    \ln \integ{\vecx} e^{-\beta\widetilde{\mathcal{H}}},
\end{equation}
where
\begin{align}
\widetilde{\mathcal{H}} &\defas 
    \mathcal{H} -
    \sum_i 
    \frac{\lambda_i^*(\beta)}{\beta}(x_i - a_i)
    -
    \frac{\xi_i^*(\beta)}{\beta}(x_i^2 - c_i - a_i^2),
\end{align}
where the Lagrange multipliers are given as 
functions of the temperature in order to make clear the order in which we will
apply $\beta \To 0$ later.

As this exact form of the Gibbs free energy is just as intractable as the 
original free energy, we will apply a Taylor expansion in order to generate an
approximate Gibbs free energy \cite{Ple1982}. We make this expansion at 
$\beta = 0$, as in the limit of infinite temperature, all interactions between
sites vanish and the system can be described only in terms of individual sites
and their relationship to the system average and their local potentials,
allowing the Gibbs free energy to be decomposed into a sum of independent terms.
Specifically, if we take the expansion up to $s$ terms,
\begin{equation}
-\beta\gibbstaylor{s}\p{\beta,\alla,\allc; \allparams} = 
    \at{\ln \widetilde{\mathcal{Z}}_{\beta}}{\beta = 0} 
    + \sum_{p = 1}^{s}
            \frac{\beta^p}{p!} 
            \pden{\beta}{p}\s{\ln\widetilde{\mathcal{Z}}_{\beta}}_{\beta = 0},                                            
    \label{eq:taylorexp}
\end{equation}
where $\widetilde{\mathcal{Z}}_{\beta}$ is the normalization of the Boltzmann
distribution defined by $\widetilde{\mathcal{H}}$ at temperature $\beta$.
At $\beta = 0$ we can find the first term of the expansion directly
\begin{align}
\ln\widetilde{\mathcal{Z}}_{0} &= 
    -\sum_i \lambda_i^*(0) a_i %\notag\\
    -\sum_i \xi_i^*(0) (a_i^2 + c_i)  \notag\\
&\quad    + \sum_i \ln \integ{x_i} P_i(x_i;\theta_i) e^{\lambda_i^*(0)x_i-\xi_i^*(0)x_i^2},                   
\end{align}                   
where we recognize that the last term is simply the normalization of the 
Gaussian-product distribution whose moments were defined in 
\eqref{eq:fa}, \eqref{eq:fc}.

We define the TAP free energy by writing 
the remainder of the expansions terms in the 
specific case of $s = 2$ \cite{Ple1982,GY1999},
\begin{align}
-\beta\gibbstaylor{2}_\beta\p{\alla,\allc; \allparams} &= 
     \sum_i \ln Z_i(\lambda_i^*(0),\xi_i^*(0); \theta_i) \notag\\    
    &\quad - \sum_i \lambda_i^*(0) a_i
    - \sum_i \xi_i^*(0) (a_i^2 + c_i) \notag \\
    &\quad
    + \beta \sum_{(i,j)} W_{ij} a_i a_j
    + \frac{\beta^2}{2} \sum_{(i,j)} W_{ij}^2 c_i c_j,
    \label{eq:tapFE}
\end{align}
where
\begin{equation}
    Z_i(B,A;\theta) \defas \integ{x} P_i(x;\theta) e^{\frac12 A x^2 - B x}.
    \label{eq:Z_i}
\end{equation}
Note that the last two terms in \eqref{eq:tapFE} come from the
Taylor expansion in \eqref{eq:taylorexp}, and are related to the
derivatives of $\boldsymbol{\lambda}^\ast$ and $\boldsymbol{\xi}^\ast$
evaluated at 0.

We still need to determine the
values of $\boldsymbol{\lambda}^*(0)$ and
 $\boldsymbol{\xi}^*(0)$, which is done by taking the 
stationarity of the expanded Gibbs free energy with respect to $\alla$
and $\allc$
\begin{align}
A_i &\defas -2\xi_i(0)^* =  -\beta^2\sum_{j\in\partial_i} \Wij^2 c_j,\\
B_i &\defas \lambda_i(0)^* =  A_i a_i + \beta \sum_{j\in\partial_i} \Wij a_j,
\end{align}
where we make the definitions of $\boldsymbol{A}$ and $\boldsymbol{B}$ for
convenience and as a direct allusion to the definitions of the cavity sums
for BP inference, given in Eqs. \eqref{eq:Bij}, \eqref{eq:Aij}.

Conversely, by deriving the stationarity conditions of the 
auxiliary fields, we obtain the
self-consistency equations for $\alla$, $\allc$, which show us that the TAP
free energy is only valid when the following self-consistencies hold,
\begin{align}
a_i = \fa(B_i, A_i; \theta_i), \quad
c_i = \fc(B_i, A_i; \theta_i),
\end{align}
where $\fa$ and $\fc$ are defined from \eqref{eq:Z_i} via $\fa
\defas \frac{\partial}{\partial B_i} \log Z_i$ and $\fc \defas
\frac{\partial^2}{\partial B_i^2} \log Z_i$.

Substituting these values closes the free energy on the marginal distribution 
moments $\alla$, $\allc$ and completes our derivation of a free energy 
approximation which is defined by $O(N)$ elements versus the $O(N^2)$ values
required by r-BP.

\subsection{Solutions of the TAP Free Energy}
\label{subsec:tapsolutions}
As given, the TAP free energy is only valid
when the self-consistency equations are met at its stationary points. Thus, 
only a certain set of $\alla$, $\allc$ can have any physical meaning. 
Additionally, we know that only at the minima of the exact Gibbs free energy 
will we have a correspondence with the original exact Helmholtz free energy. 

While the exact Gibbs free energy in terms
of the moments $\alla$, $\allc$, is convex for exponential family 
$P_0(\vecx;\theta)$ such that
$e^{-\beta \widetilde{\mathcal{H}}} \geq 0$, 
the TAP free energy can possess multiple stationary points whose number
increases rapidly as $\beta$ grows \cite{MPV1986}. Later
in Sec. \ref{sec:experiments}, we show 
that as GRBM training progresses, so does the number of identified 
TAP solutions. This can be explained due to the variance of the
weights $W$ growing with training. For fixed $\beta = 1$, as
we use in our practical GRBM implementation,
the variance of the weights serves as an effective inverse temperature,
and its increasing magnitude has an identical effect to the system cooling as 
$\beta$ increases.

Additionally, while the
Gibbs free energy has a correspondence with the Helmholtz free energy at its
minimum, this is not necessarily true for the TAP free energy. The
approximate nature of the 
second-order expansion removes this correspondence. 
Thus, it may not be possible to ascertain an accurate estimate of the Helmholtz
free energy from a \emph{single} set of inferred $\alla$, $\allc$, as shown in 
Fig. \ref{fig:freeEnergies}. In the case of the na\"ive mean-field estimate of 
the Gibbs free energy, it is true that
$\F_\beta \leq \widetilde{\gibbs}^{(1)}_\beta$.
This implies that one should attempt to find the
minima of $\widetilde{\gibbs}^{(1)}_\beta$ in order to find a more accurate estimate of 
$\F_\beta$, a foundational principle in variational approaches. However, while 
the extra expansion term in the TAP free energy should improve its 
accuracy in modeling $\gibbs_\beta$ over $\alla$, $\allc$, it does not provide a lower
bound, and so an estimate of $\F_\beta$ from the TAP free energy could be an 
under- \emph{or} over-estimate.

\def\svgwidth{0.5\textwidth}
\begin{figure}
    \begin{center}
        \begingroup%
  \makeatletter%
  \providecommand\color[2][]{%
    \errmessage{(Inkscape) Color is used for the text in Inkscape, but the package 'color.sty' is not loaded}%
    \renewcommand\color[2][]{}%
  }%
  \providecommand\transparent[1]{%
    \errmessage{(Inkscape) Transparency is used (non-zero) for the text in Inkscape, but the package 'transparent.sty' is not loaded}%
    \renewcommand\transparent[1]{}%
  }%
  \providecommand\rotatebox[2]{#2}%
  \ifx\svgwidth\undefined%
    \setlength{\unitlength}{377.55328369bp}%
    \ifx\svgscale\undefined%
      \relax%
    \else%
      \setlength{\unitlength}{\unitlength * \real{\svgscale}}%
    \fi%
  \else%
    \setlength{\unitlength}{\svgwidth}%
  \fi%
  \global\let\svgwidth\undefined%
  \global\let\svgscale\undefined%
  \makeatother%
  \begin{picture}(1,0.5641105)%
    \put(0,0){\includegraphics[width=\unitlength,page=1]{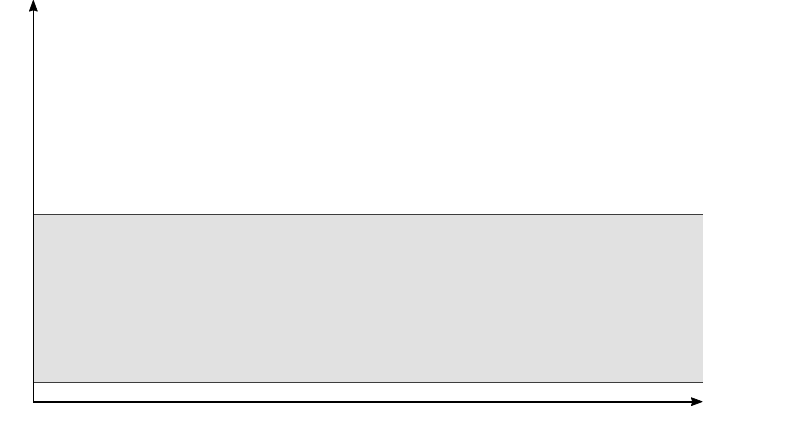}}%
    \put(0.42894904,0.00120016){\color[rgb]{0,0,0}\makebox(0,0)[lb]{$\br{\alla,\allc}$}}%
    \put(0,0){\includegraphics[width=\unitlength,page=2]{fig2.pdf}}%
    \put(0.051836,0.2237331){\color[rgb]{0,0,0}\makebox(0,0)[lb]{$\F_\beta$}}%
    \put(0,0){\includegraphics[width=\unitlength,page=3]{fig2.pdf}}%
    \put(0.20466311,0.44723154){\color[rgb]{0,0,0}\makebox(0,0)[lb]{$\gibbs_\beta$}}%
    \put(0,0){\includegraphics[width=\unitlength,page=4]{fig2.pdf}}%
    \put(0.78428529,0.30345999){\color[rgb]{0,0,0}\makebox(0,0)[lb]{$\widetilde{\gibbs}^{(2)}_\beta$}}%
    \put(0,0){\includegraphics[width=\unitlength,page=5]{fig2.pdf}}%
    \put(0.78736548,0.12054078){\color[rgb]{0,0,0}\makebox(0,0)[lb]{$\an{\widetilde{\gibbs}^{(2)}_\beta}_*$}}%
  \end{picture}%
\endgroup%
        \caption{Cartoon description for estimating the Helmholtz free energy (\emph{dotted}) via the
                 Gibbs (\emph{blue dash}) and TAP (\emph{red}) free energies. For this
                 example of a convex Gibbs free energy, there exists one
                 unique minimum over the moments ${\alla,\allc}$, and 
                 the Gibbs free energy here matches $\F_\beta$. The range of TAP
                 free energies (\emph{gray box}) gives a boundary on the location of 
                 $\F_\beta$. Averaging the TAP free energies (\emph{dash-dot}) provides an 
                 estimate of $\F_\beta$.
                 \label{fig:freeEnergies}}
    \end{center}    
\end{figure}

Instead, one might attempt to obtain an estimate of the Helmholtz free energy by
utilizing either all or a subset of the equilibrium solutions of the TAP free 
energy. Since there is no manner by which we might distinguish the equilibrium
moments by their proximity to the unknown $\F_\beta$, averaging 
the TAP free energy across its solutions, denoted as 
$\an{\widetilde{\gibbs}^{(2)}_\beta}_*$, can serve as a simple estimator of 
$\F_\beta$
\cite{MPV1986}. In \cite{DY1983} a weighting was introduced to the
average, correcting the Helmholtz free energy estimate at low 
temperature by removing the over-influence of the exponential number of
high-energy solutions. The weights in this approach are proportional to the
exponents of each solution's TAP free energy, placing much stronger emphasis on
low-energy solutions.

However, such an approach is not well-justified in the our general setting
of $P_i$,  where we expect large deviations from the expectations
derived for the SK model. Additionally, while this weighting scheme is
shown across the \emph{entire set} of solutions for a particular
random SK model, in our case, we are interested in the solution space 
centered on the particular dataset we wish to model. 
Since the solutions are computed by iterating the TAP self-consistency 
equations, we can easily probe this region by initializing the iteration 
according to the training data.
Subsequently, we do not encounter a band of high-energy solutions that
we must weight against. Instead, we obtain a set of solutions over
a small region of the support of the TAP free energy. Due to the uniformity of
these solutions, un-weighted averaging across the solutions seems the 
best approach in terms of efficiency. In the subsequent section, we
explore some of these properties numerically for trained RBMs.

\section{RBMs as TAP Machines}
\label{sec:implementation}
To utilize the TAP inference of Sec. \ref{sec:tap}, we need to write the TAP 
free energy in terms of the variables of the RBM.  To clarify the
bipartite structure of the GRBM, we rewrite the TAP free energy in terms of the 
hidden and visible variables at fixed temperature $\beta = 1$,
\begin{align}
&-\mathbb{F}_{RBM}\p{\allav,\allcv,\allah,\allch; \allparams} = \notag\\
     &\quad\sum_i \bra{\ln Z_i^{\rm v}(B_i^{\rm v},A_i^{\rm v}; \theta_i^{\rm v})     
    - B_i^{\rm v} a^{\rm v}_i
    + \frac{1}{2} A_i^{\rm v} ((a_i^{\rm v})^2 + c_i^{\rm v})} 
    \notag\\
    &+ \sum_j \bra{\ln Z_j^{\rm h}(B_j^{\rm h},A_j^{\rm h}; \theta_j^{\rm h})     
    - B_j^{\rm h} a^{\rm h}_j
    + \frac{1}{2} A_j^{\rm h} ((a_j^{\rm h})^2 + c_j^{\rm h})} 
    \notag\\
    &+ \sum_{ij}\bra{ W_{ij} a_i^{\rm v} a_j^{\rm h}
    + \frac{1}{2} W_{ij}^2 c_i^{\rm v} c_j^{\rm h}},
\end{align}
where $\bra{\allav, \allcv}$ and $\bra{\allah,\allch}$ are the means 
and variances of the visible and hidden variables, respectively.

As in Sec. \ref{sec:tap}, solutions of the TAP GRBM free energy can be found by
a fixed-point iteration, as shown in Alg. \ref{alg:rbmInf}, which bears much
resemblance to the AMP iteration derived in the context
of compressed sensing \cite{MM2009,KMS2012} and 
matrix factorization \cite{LKZ2015,LKZ2017,SF2012}. 
We note that rather than updating over the entire system at each time step, 
fixing one side at a time has the effect of stabilizing the fixed-point iteration. 
For clarity, Alg. \ref{alg:rbmInf} is written for a single initialization of the visible
marginals. However, as noted in Sec. \ref{subsec:tapsolutions}, there exist a
large number of initialization-dependent solutions to the TAP free energy. Thus,
in order to capture the plurality of modes present in the TAP free energy 
landscape, one should run this inference independently for many different
initializations. 

If the use case of the GRBM requires that we only train the GRBM
tightly to the data space (e.g. data imputation), it makes sense to fix the 
initializations of the inference to points drawn from the dataset, 
\begin{align}
  \mathbf{a}^{\rm v, (0)} &= \mathbf{x}^{(m)} \quad \text{where} \quad m \in \bra{1,\dots,M}, \label{eq:init_a} \\
  \mathbf{c}^{\rm v, (0)} &= \mathbf{0}. \label{eq:init_c}
\end{align}
In order to train the GRBM more holistically, structured
random initializations can help probe modes outside of the data space.
In this work we do not employ this strategy, restricting ourselves to
a deterministic initialization.

For a set of TAP solutions 
$\bra{\mathbf{a}_k, \mathbf{c}_k, \mathbf{B}_k,\mathbf{A}_k}$ for 
$k \in \bra{1,\dots,K}$ at fixed GRBM parameters $\bra{W,\allparams}$, the
TAP-approximated log-likelihood of can be written as
\begin{align}
&\ln P(\mathbf{x};W,\theta) \approx 
  \sum_i \ln P_i(x_i;\visparam{i}) \notag\\
  &\quad\quad+ \sum_j \ln Z_j^{\rm h}(\sum_i \Wij x_i, 0;\hidparam{j}) \notag \\
  &\quad\quad + \frac{1}{K} \sum_k \mathbb{F}_{RBM}\p{\allav_k,\allcv_k,\allah_k,\allch_k; \allparams},
  \label{eq:tapll}
\end{align}
where $Z_j^{\rm h}(B,0;\theta)$ is the normalization of the
conditional expectation of Eq. \eqref{eq:condexp}, since
$\fa(B,0;\theta) = f(B;\theta)$.

After re-introducing an averaging of the log-likelihood over the samples in the 
mini-batch, the gradients of the TAP-approximated GRBM log-likelihood w.r.t. 
the model parameters are given by
\begin{align}
\label{eq:gradupdate1}
\Delta W_{ij} \approx
  \frac{1}{M}&\sum_{m}x_i^{(m)} \fa^{\rm h}(\sum_i \Wij x_i^{(m)},0;\hidparam{j}) \notag\\
  -
  \frac{1}{K}&\sum_{k} \bra{a_{i,k}^{\rm v} a_{j,k}^{\rm h}
                       + \Wij c_{i,k}^{\rm v} c_{j,k}^{\rm h}},\\
\label{eq:gradupdate2}
\Delta \hidparam{j} \approx 
  \frac{1}{M} &\sum_m  \pde{\hidparam{j}}\s{\ln Z_j^{\rm h}(\sum_i \Wij x_i^{(m)}, 0;\hidparam{j})}\notag\\
  -
  \frac{1}{K} &\sum_k \pde{\hidparam{j}}\s{\ln Z_j^{\rm h}(B_{j,k}^{\rm h}, A_{j,k}^{\rm h}; \hidparam{j})}, \\
\label{eq:gradupdate3}
\Delta \visparam{i} \approx
  \frac{1}{M} &\sum_m \pde{\visparam{i}}\s{\ln P_i^{\rm v}(x_i^{(m)};\visparam{i})} \notag\\
  -
  \frac{1}{K} &\sum_k \pde{\visparam{i}}\s{\ln Z_i^{\rm v}(B_{i,k}^{\rm v}, A_{i,k}^{\rm v}; \visparam{i})}.
\end{align}

\begin{figure}
\begin{algorithm}[H]
  \caption{TAP Inference for GRBMs \label{alg:rbmInf}}
  \begin{algorithmic}
    \STATE \emph{Input}: $W$, $\allparams$
    \STATE \emph{Initialize}: $t=0$, $\mathbf{a}^{\rm v, (0)}$, $\mathbf{c}^{\rm v, (0)}$
    \REPEAT
      \STATE \emph{Hidden Side Updates}\Vhrulefill
      \STATE $A_j^{\rm h, (t+1)} = -\sum_i \Wij^2 c_i^{\rm v, (t)}$
      \STATE $B_j^{\rm h, (t+1)} = A_j^{\rm h, (t+1)} a_j^{\rm h, (t)} + \sum_i \Wij a_i^{\rm v, (t)}$
      \STATE $a_j^{\rm h, (t+1)} = \fa^{\rm h}\p{B_j^{\rm h, (t+1)}, A_j^{\rm h, (t+1)};\hidparam{j}}$
      \STATE $c_j^{\rm h, (t+1)} = \fc^{\rm h}\p{B_j^{\rm h, (t+1)}, A_j^{\rm h, (t+1)};\hidparam{j}}$
      \STATE 
      \STATE \emph{Visible Side Updates}\Vhrulefill
      \STATE $A_i^{\rm v, (t+1)} = -\sum_j \Wij^2 c_j^{\rm h, (t+1)}$
      \STATE $B_i^{\rm v, (t+1)} = A_i^{\rm v, (t+1)} a_i^{\rm v, (t)} + \sum_j \Wij a_j^{\rm h, (t+1)}$
      \STATE $a_i^{\rm v, (t+1)} = \fa^{\rm v}\p{B_i^{\rm v, (t+1)}, A_i^{\rm v, (t+1)} ;\visparam{i}}$
      \STATE $c_i^{\rm v, (t+1)} = \fc^{\rm v}\p{B_i^{\rm v, (t+1)}, A_i^{\rm v, (t+1)} ;\visparam{i}}$
      \STATE
      \STATE $t = t+1$
    \UNTIL{Convergence}
  \end{algorithmic}
\end{algorithm}
\end{figure}
In the presented gradients, we make the point that the set of data samples and 
the set of TAP solutions can have different cardinality. For example, one might
employ a mini-batch strategy to training, where the set of data samples used in
the gradient calculation might be on the order $10^2$. However, depending on the
application of the GRBM, one might desire to probe a very large number of 
TAP solutions in order to have a more accurate picture of the representations 
learned by the GRBM. In this case, one might start with a very large number of 
initializations, resulting in a very large number, $K\gg M$, of unique TAP solutions.
Or, contrary, while one might start with a number of initializations equal to 
$M$, the number of unique solutions might be $K\ll M$, especially early in 
training or when the number of hidden units is small.

Using these gradients, a simple gradient ascent with a fixed or monotonically 
decreasing step-size $\gamma$ can be used to update these GRBM parameters. We
present the final GRBM training algorithm in Alg. \ref{alg:rbmTrain}.

\begin{figure}[ht]
\begin{algorithm}[H]
  \caption{GRBM Training \label{alg:rbmTrain}}
  \begin{algorithmic}
    \STATE \emph{Input}: $\mathbf{X}$, $T$, $M$, $K$, $R(\cdot)$
    \STATE \emph{Initialize}: $W_{ij}^{(0)} \sim \mathcal{N}(0,\sigma)$, $\allparams^{(0)} \leftarrow \mathbf{X}$
    \REPEAT
      \FOR{All mini-batches of size $M$}
        \STATE $\mathbf{a}^{(t+1)}, \mathbf{c}^{(t+1)}, \mathbf{B}^{(t+1)}, \mathbf{A}^{(t+1)} \leftarrow$ Alg. \ref{alg:rbmInf}$(W^{(t)},\allparams^{(t)}) \times K$
        \STATE $W_{ij}^{(t+1)} \leftarrow W_{ij}^{(t)} + \gamma \Delta W_{ij} ^{(t)} + \gamma\epsilon {\rm R}(W_{ij} ^{(t)}) + \eta \Delta W_{ij}^{(t-1)}$
        \STATE ${\hidparam{j}}^{,(t+1)} \leftarrow {\hidparam{j}}^{,(t)} + \gamma \Delta {\hidparam{j}}^{,(t)}$
        \STATE ${\visparam{i}}^{,(t+1)} \leftarrow {\visparam{i}}^{,(t)} + \gamma \Delta {\visparam{i}}^{,(t)}$        
      \ENDFOR
      \STATE $t \leftarrow t+1$
    \UNTIL{$t > T$}
  \end{algorithmic}
\end{algorithm}
\end{figure}

Besides considering non-binary units, another natural extension of traditional
RBMs is to consider additional hidden layers, as in Deep Boltzmann Machines (DBMs).
It is possible to define and train \emph{deep} TAP machines, as well. 
Probabilistic DBMs are substantially harder to train than RBMs as the
data-dependent (or clamped) terms  of the gradient updates (\ref{eq:gradupdate1}-\ref{eq:gradupdate3}) become intractable with depth. 
Interestingly, state-of-the-art training algorithms retain a Monte Carlo evaluation 
of other intractable terms, while introducing a na{\i}ve mean-field approximation 
of these data-dependent terms. For \emph{deep} TAP machines, we consistently utilize
the TAP equations.  The explicit definition and training algorithm are fully 
described in Appendix \ref{sec:dbm}.

\section{Experiments}
\label{sec:experiments}
\subsection{Datasets}
\textbf{\textit{MNIST --- }}
The MNIST handwritten digit dataset \cite{LBB1998} consists of both a 
training and testing set, each with 60,0000 and 10,000 samples, respectively.
The 
data samples are real-valued $28\times 28$ pixel 8-bit grayscale images which we
normalize to the dynamic range of $[0,1]$. The images themselves are centered 
crops of the digits `0' through `9' in roughly balanced proportion. We construct
two separate versions of the MNIST dataset. The first, which we refer to as 
\emph{binary-MNIST}, applies a thresholding such that pixel values $> 0.5$ are set to
1 and all others to 0. The second, \emph{real-MNIST}, simply refers to the normalized 
dataset introduced above.

\textbf{\textit{CBCL --- }} 
The CBCL face database \cite{cbcl} consists of both face and non-face 8-bit
grayscale $19\times 19$ pixel images. For our experiments, we utilize only the
face images. The database contains 2,429 training and 472 testing samples of 
face images. For our experiments, we normalize the samples to the dynamic range
of $[0,1]$. 

\begin{figure}[ht]
\begin{table}[H]
\centering
\begin{tabular}{| r || c | c | c |}
\hline
~ & binary-MNIST & real-MNIST & CBCL \\
\hline\hline
$N_{v}$ & 784 & 784 & 361 \\
$N_{h}$ & $\bra{25,50,100,500}$ & $\bra{100,500}$ & 256 \\
$M$ & 100 & 100 & 20 \\
$K$ & 100 & 100 & 20 \\
Prior Vis. & B. & Tr. Gauss.-B. & Tr. Gauss. \\
Prior Hid. & B. & B. & B. \\
$R(\cdot)$ & $\ell_2$ & $\ell_2$ & $\ell_2$ \\
$\gamma$ & 0.005 & $\s{10^{-2},10^{-5}}$ & 0.005 \\
$\epsilon$ & 0.001 & 0.001 & 0.01 \\
$\eta$ & 0.5 & 0.5 & 0.5 \\
\hline
\end{tabular}
\caption{Parameter settings for GRBM training.
\label{tab:params}}
\end{table}
\end{figure}

\begin{figure}
\centering
  (a) \emph{binary-MNIST}\\
  \includegraphics[width=0.475\textwidth]{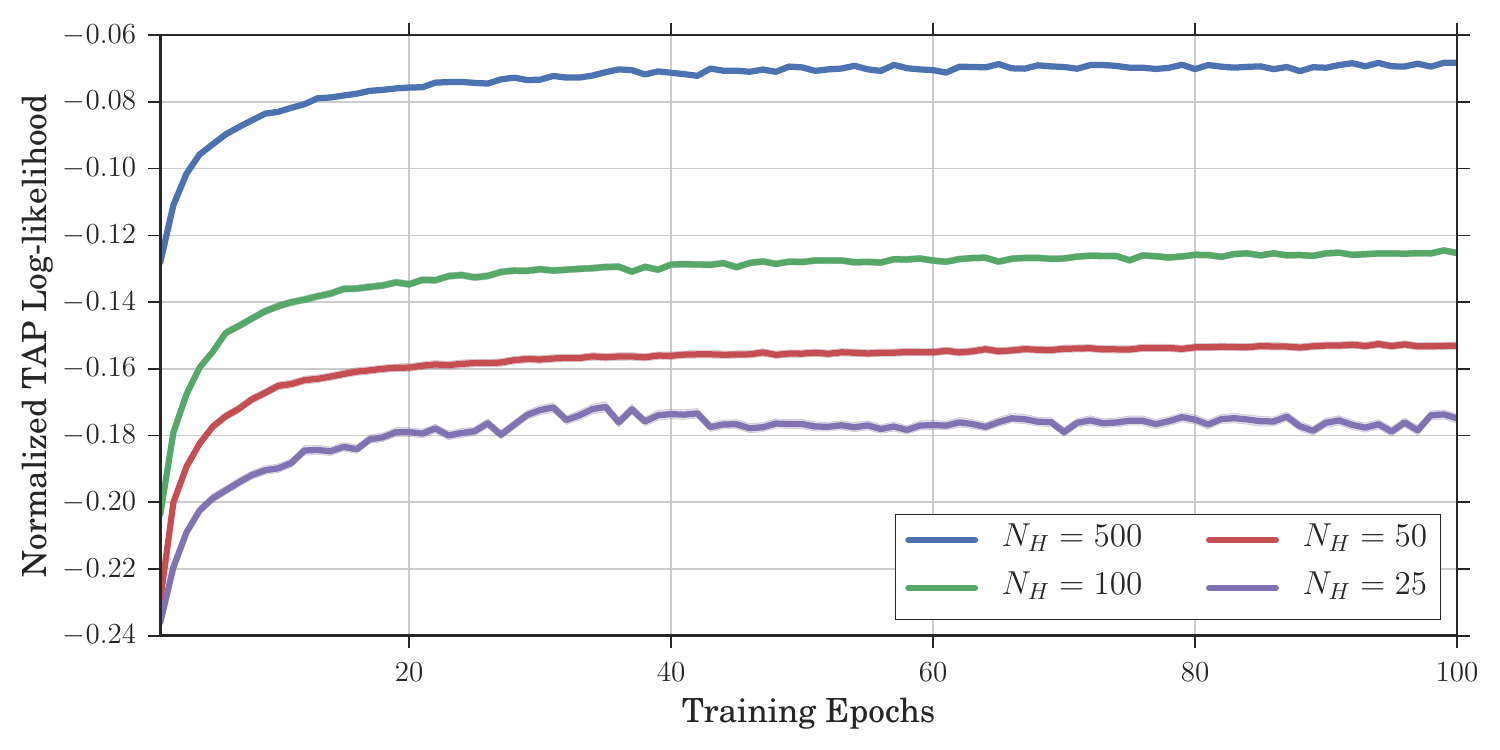}\\
  (b) \emph{real-MNIST}\\
  \includegraphics[width=0.475\textwidth]{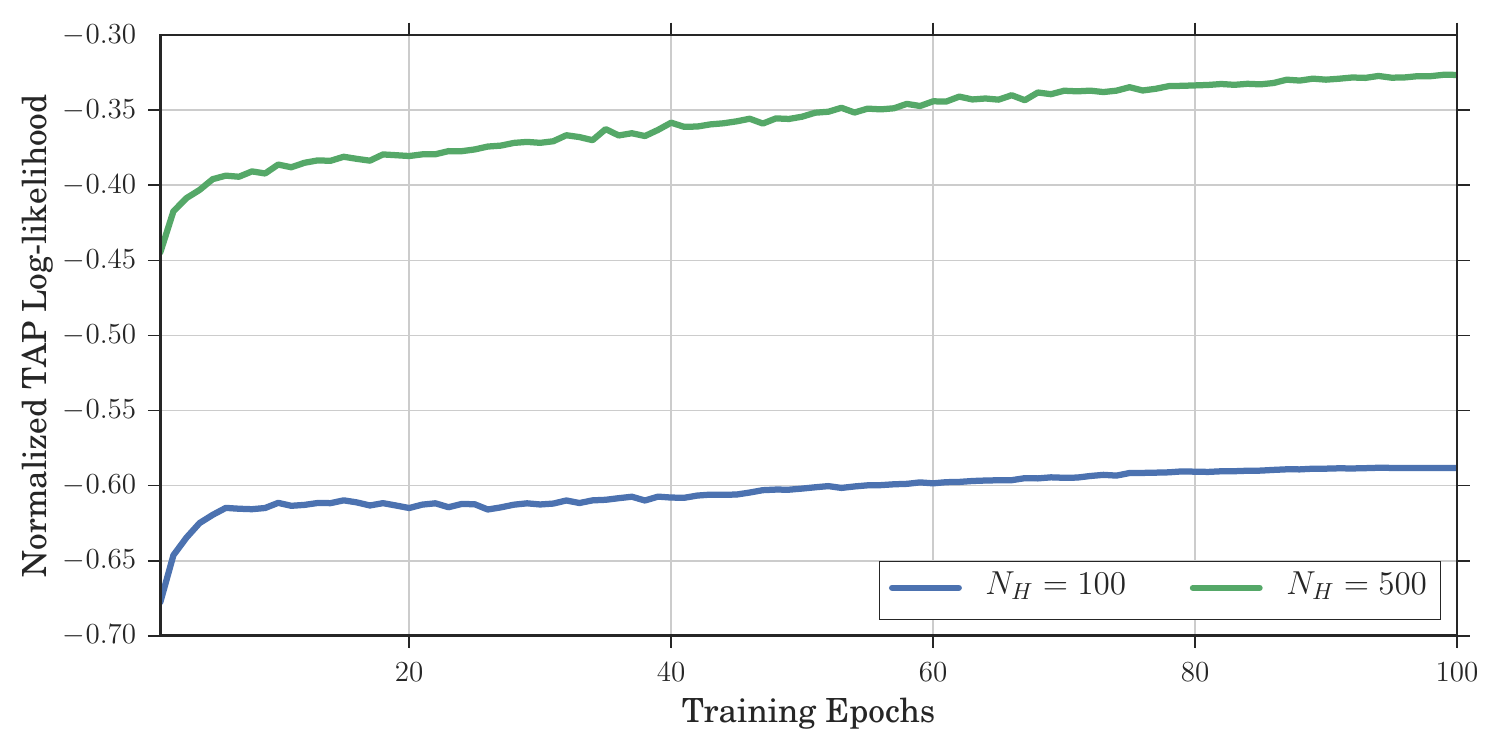}\\
  (c) \emph{CBCL}\\
  \includegraphics[width=0.475\textwidth]{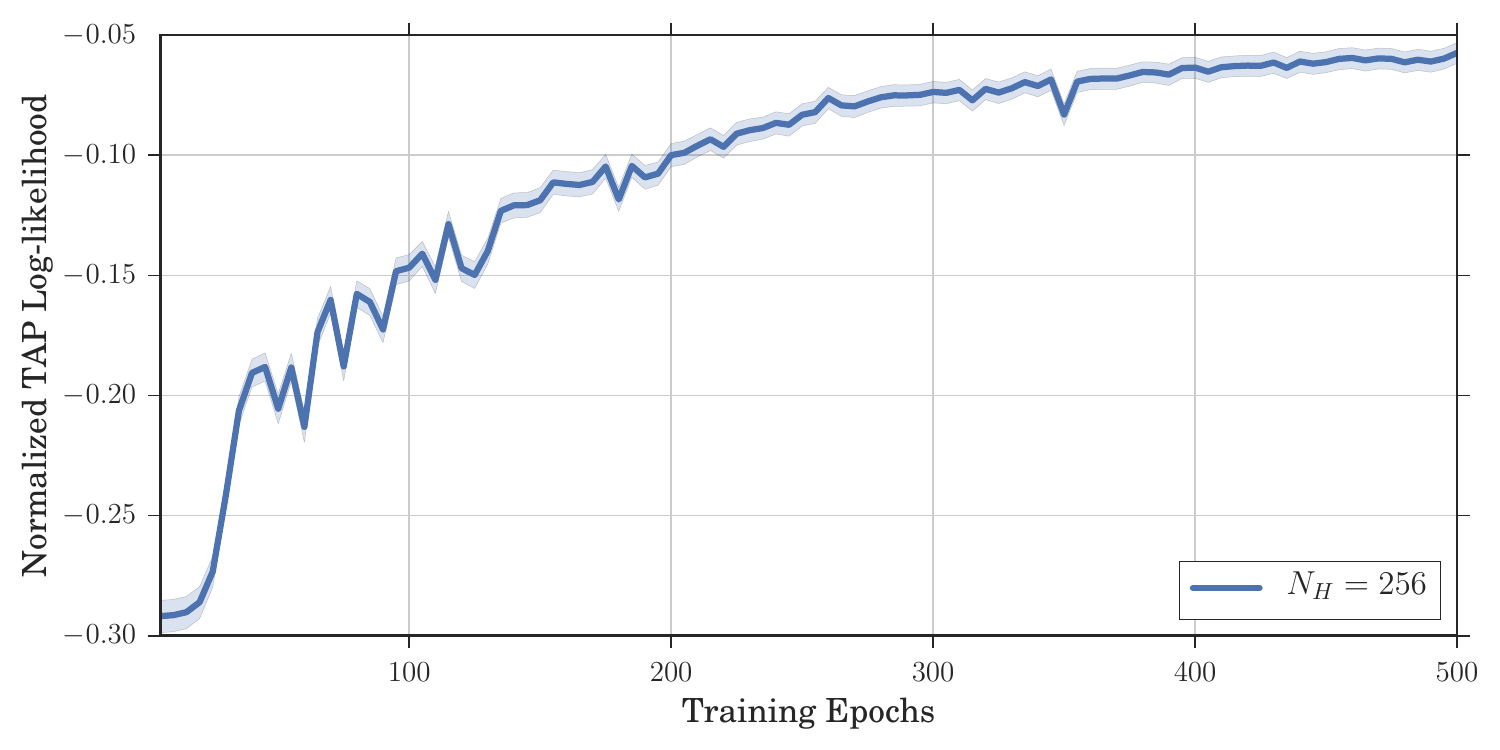}
  \caption{\label{fig:bmnist-tapLL} 
            Training performance over 100 epochs for the tested datasets 
            over varying numbers of hidden units, $N_h$. 
            Performance is measured in terms of the normalized (per-unit) TAP log-likelihood
            estimate computed for 10,000 training data samples. The TAP free 
            energy is estimated using the unique TAP solutions 
            thermalized from initial conditions drawn from the data samples,
            as in \eqref{eq:init_a}--\eqref{eq:init_c}. 
            Thermalization is determined by the convergence of the 
            magnetizations up to a difference of $10^{-8}$ in MSE between 
            iterations. 
            Solid lines indicate the average normalized TAP log-likelihood over
            the tested training samples and shaded regions indicate standard 
            error.
            }
\end{figure}

\subsection{Learning Dynamics}
We now investigate the behavior of the GRBM over the course of the learning
procedure, looking at a few metrics of interest: the TAP-approximated
log-likelihood of the training dataset, the TAP free energy, and the number of
discovered TAP solutions. We note that each of these metrics is unique to the 
TAP-based model of the GRBM. 

% \pagebreak
% \onecolumngrid
% \begin{center}
\begin{figure}[ht]
\centering
  \begin{minipage}[c]{0.49\textwidth}
    (a) \emph{binary-MNIST}\\
    \includegraphics[width=\textwidth]{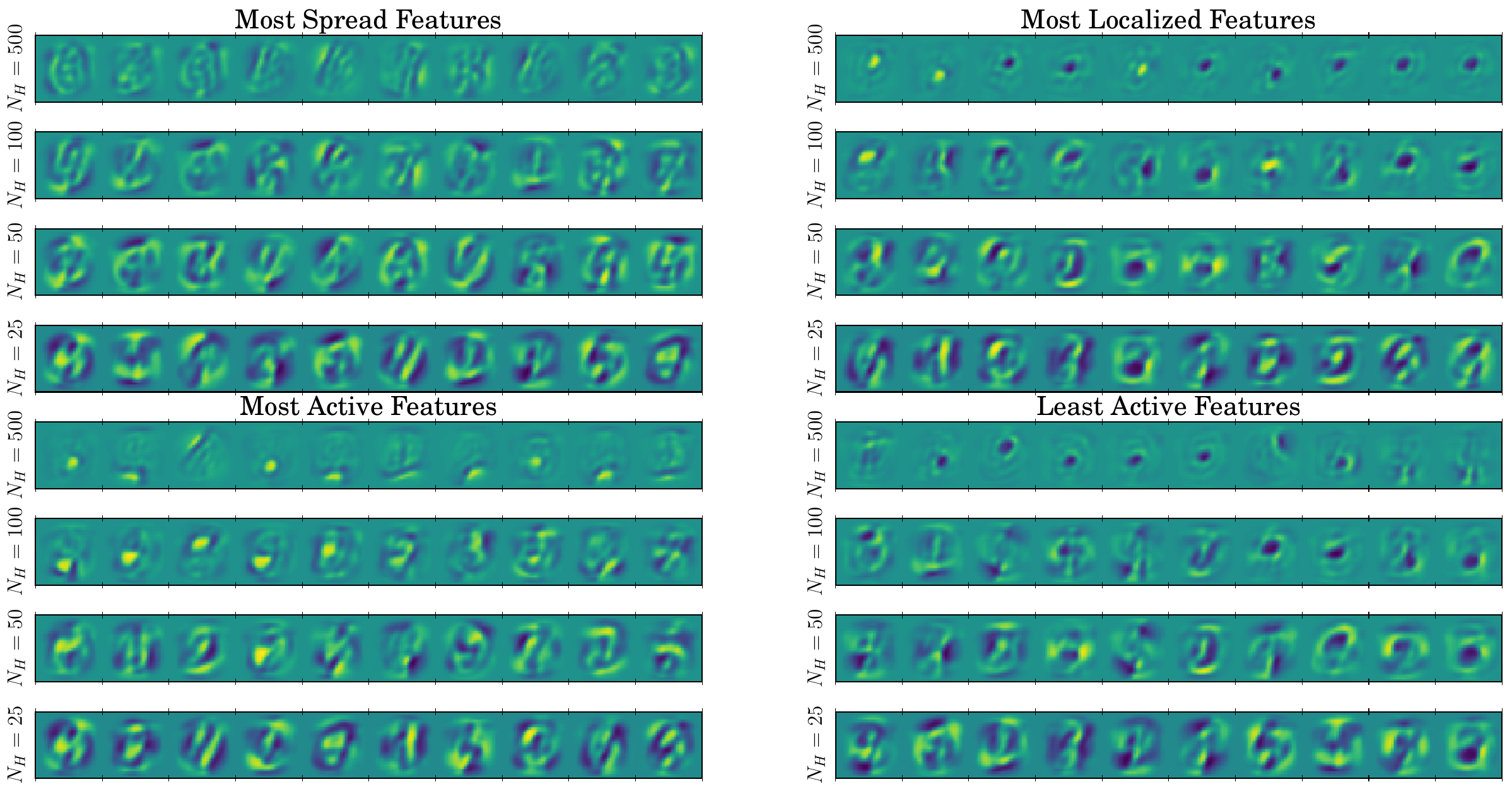}
  \end{minipage}
  ~
  \begin{minipage}[c]{0.49\textwidth}
    (b) \emph{real-MNIST}\\
    \includegraphics[width=\textwidth]{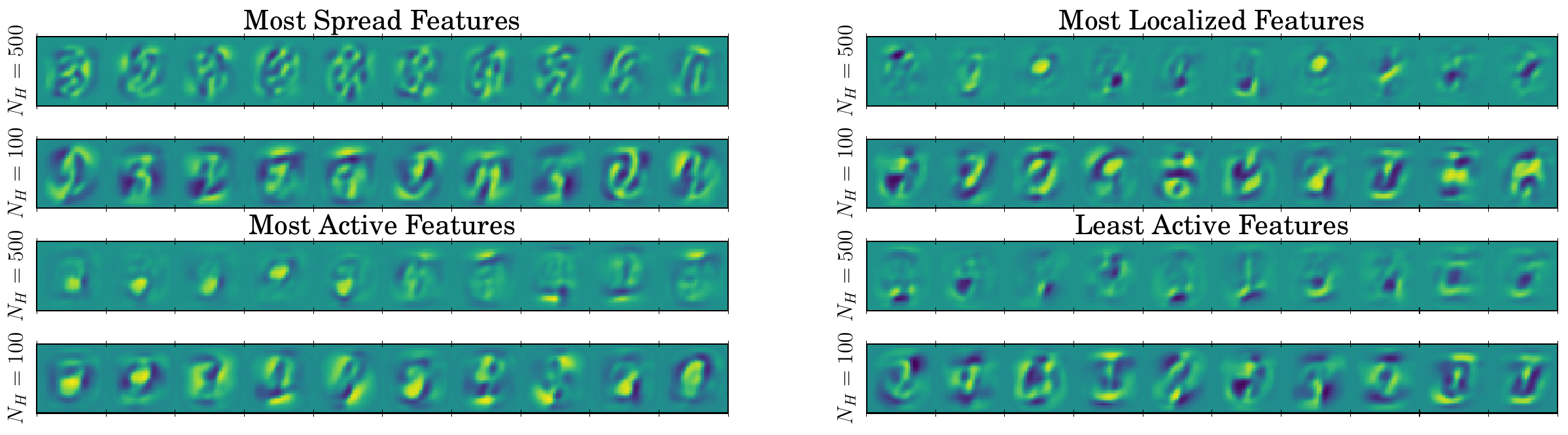}\\
    (c) \emph{CBCL}\\
    \includegraphics[width=\textwidth]{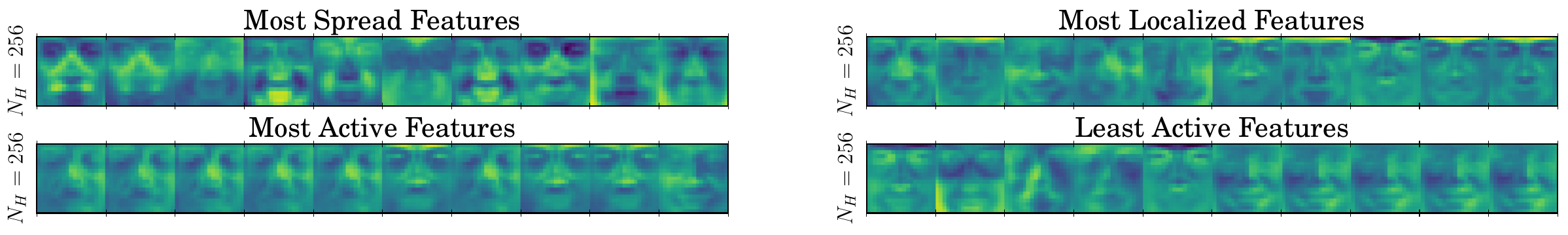}
  \end{minipage}
  \caption{\label{fig:bmnist-rf} 
            Subsets of the final receptive fields, i.e.  the columns of $W$, 
            obtained by TAP training of GRBM models with varying numbers of
            hidden units, $N_h$.
            For the receptive fields,
            dark blue and yellow are mapped to $-1$ and $+1$, respectively,
            and green indicates a value of $0$. Receptive fields are ranked
            according to two criteria. First, spread, and conversely localization,
            as measured by the $\ell_p$-norm of the receptive field, for $p = 0.1$.
            Second, by activation level, as measured by the mean activation of each
            receptive field's corresponding hidden unit averaged across the 
            training dataset.}
\end{figure}
% \end{center}
% \twocolumngrid

While it was empirically shown in \cite{GTZ2015} that CD does indeed
increase the TAP log-likelihood in the case of binary RBMs, the specific
construction of CD is entirely independent from the TAP model of the GRBM.
Thus, it is hard to say that a CD or TAP-trained GRBM is ``better'' in a
general case.  At present, we present comparisons between TAP GRBMs of
varying complexity trained under fixed hyper-parameters settings, as
indicated in Table \ref{tab:params}.

In Fig. \ref{fig:bmnist-tapLL} we see a comparison of the TAP log-likelihood as
a function of training \emph{epochs} for \emph{binary-MNIST} for binary RBMs consisting
of differing numbers of hidden units. As the gradient-ascent on the log-likelihood
is performed batch-by-batch over the training data, we define one \emph{epoch}
to be a single pass over the training data: every example has been presented to
the gradient ascent once. The specifics of this particular experiment
are given in the caption. We note that for equal comparison across
varying model complexity, this log-likelihood is normalized over the number of
visible and hidden units present in the model. In this way, we observe a 
``per-unit'' TAP log-likelihood, which gives us a measure of the concentration
of representational power encapsulated in each unit of the model. Increasing
values of the normalized TAP log-likelihood indicate that the evaluated training
samples are becoming more likely given the state of the GRBM model parameters. 

It can be observed that at each level of complexity, the TAP log-likelihood of the
data rapidly increases as the values of $W$ quickly adjust from random 
initializations to receptive fields correlated with the training data. However,
across each of the tested models, by about the $20^{\rm th}$ epoch the rate of
increase of the TAP log-likelihood tapers off to a constant rate of improvement.

For reference, we also show a subset of the trained receptive
fields, i.e. the rows  of $W$, for each of the tested experiments. Since the full 
set of receptive fields would be too large to display, we attempt to show some 
representative samples in Fig. \ref{fig:bmnist-rf} by looking at the extreme 
samples in terms of spatial spread/localization and activity over the training
set. We observe that the trained GRBMs, in the case of both \emph{binary-MNIST}
and \emph{real-MNIST}, are able to learn both the localized and stroke features
commonly observed in the literature for binary RBMs trained on the MNIST dataset
\cite{Hin2002,CRI2013}.
It is interesting to note that even in the case of \emph{real-MNIST}, where we 
are using the novel implementation of truncated Gauss-Bernoulli visible units (see
Appendix C\ref{sec:apdx_trunc_gb}),
we are able to observe similar learned features as in the case of 
\emph{binary-MNIST}. We take this as an empirical indication that the proposed 
framework of GRBM learning is truly learning correlations present in the dataset
as intended. 
Finally, we see feature localization increase with the number of hidden units.

To date, understanding ``what'' an RBM learns from the unlabelled data has mostly
been a purely subjective exercise in studying the receptive fields, as shown
Fig. \ref{fig:bmnist-rf}. However, the interpretation of the GRBM as a TAP
machine can provide us a novel insight in the nature and dynamics of GRBM 
learning via the stationary points of the TAP free energy, which we 
detail in the next section.

\pagebreak
\onecolumngrid
\begin{center}
\begin{figure}[t]
\centering
  \begin{tabular}{c c c}
  \includegraphics[width=0.333\textwidth]{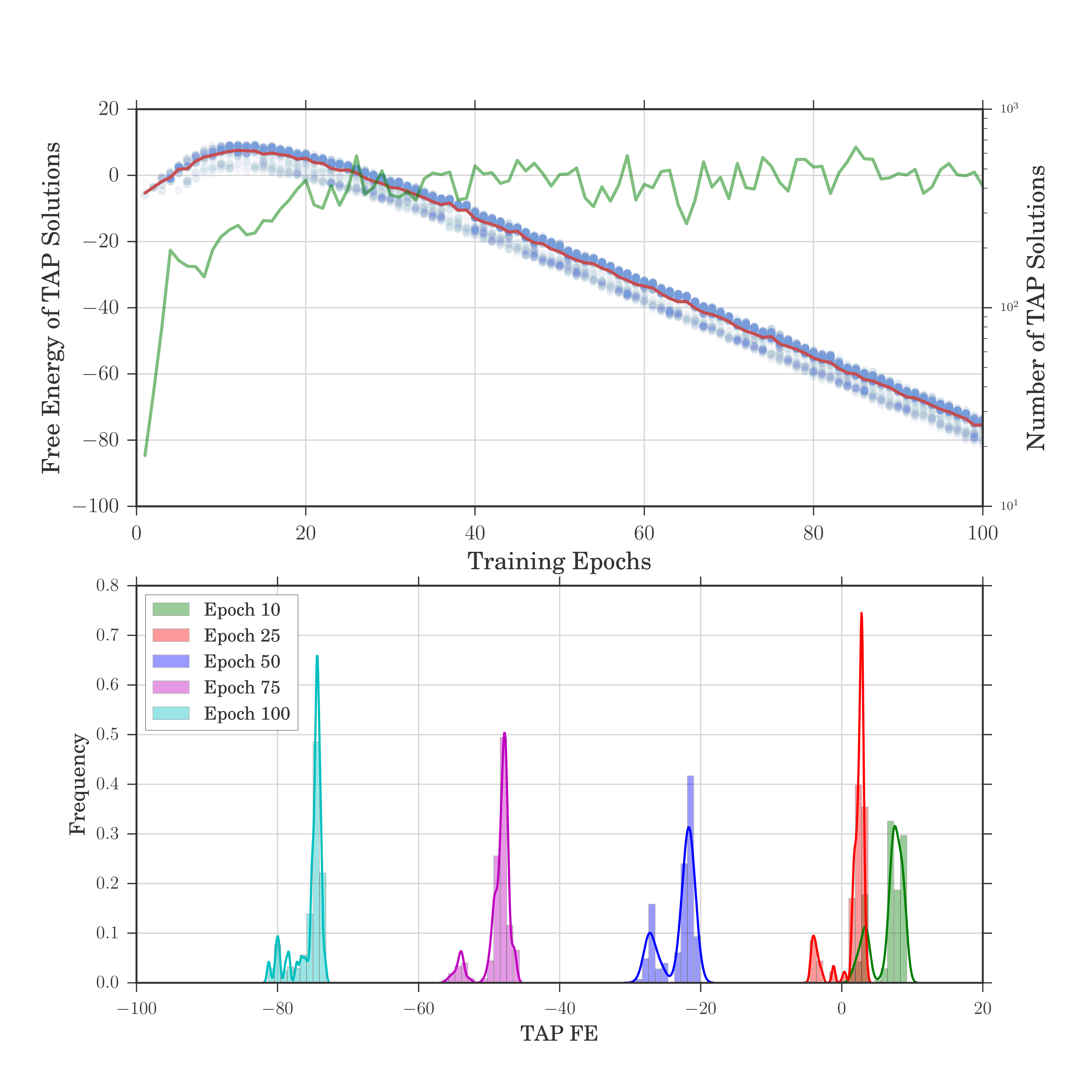}
  &
  \includegraphics[width=0.333\textwidth]{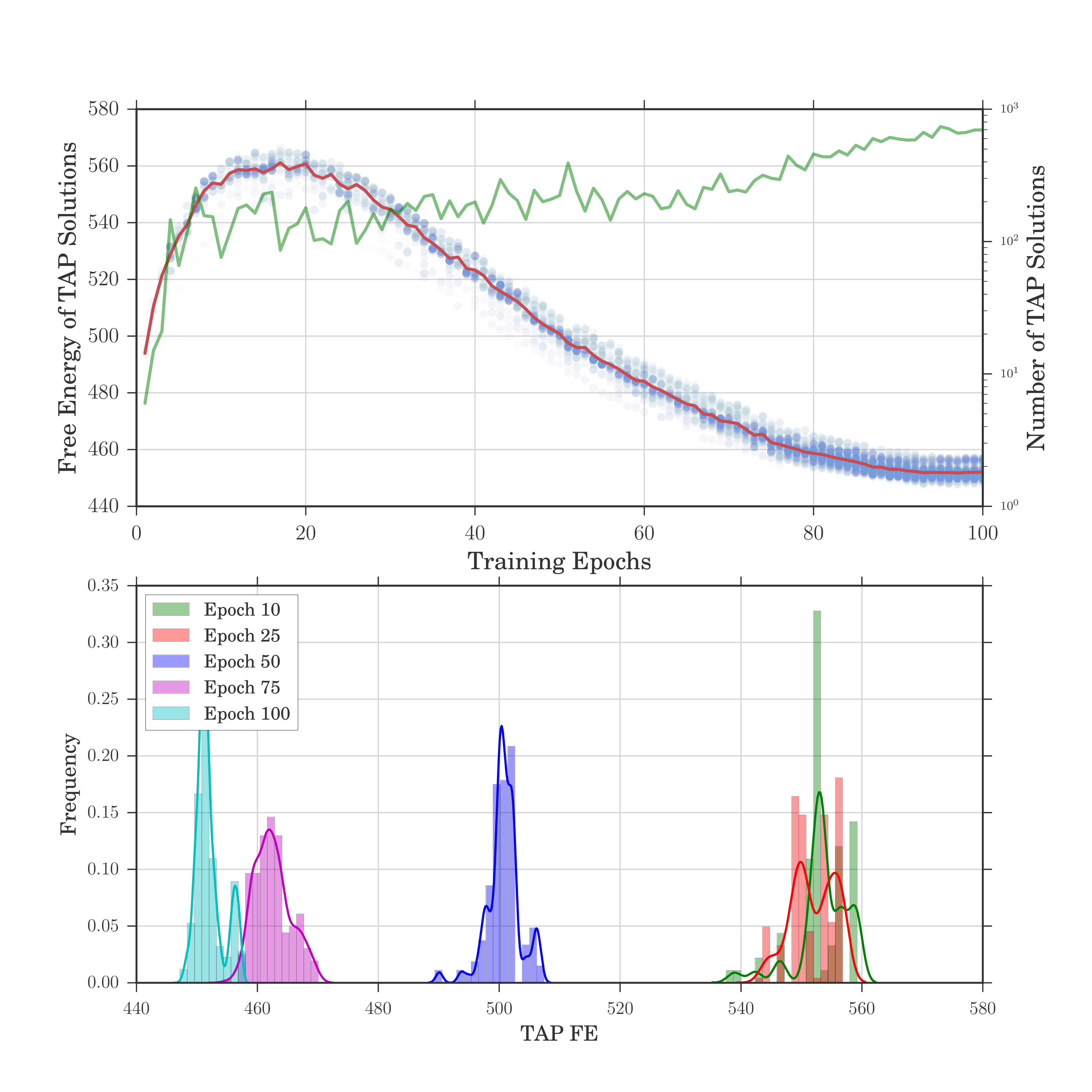}
  &
  \includegraphics[width=0.333\textwidth]{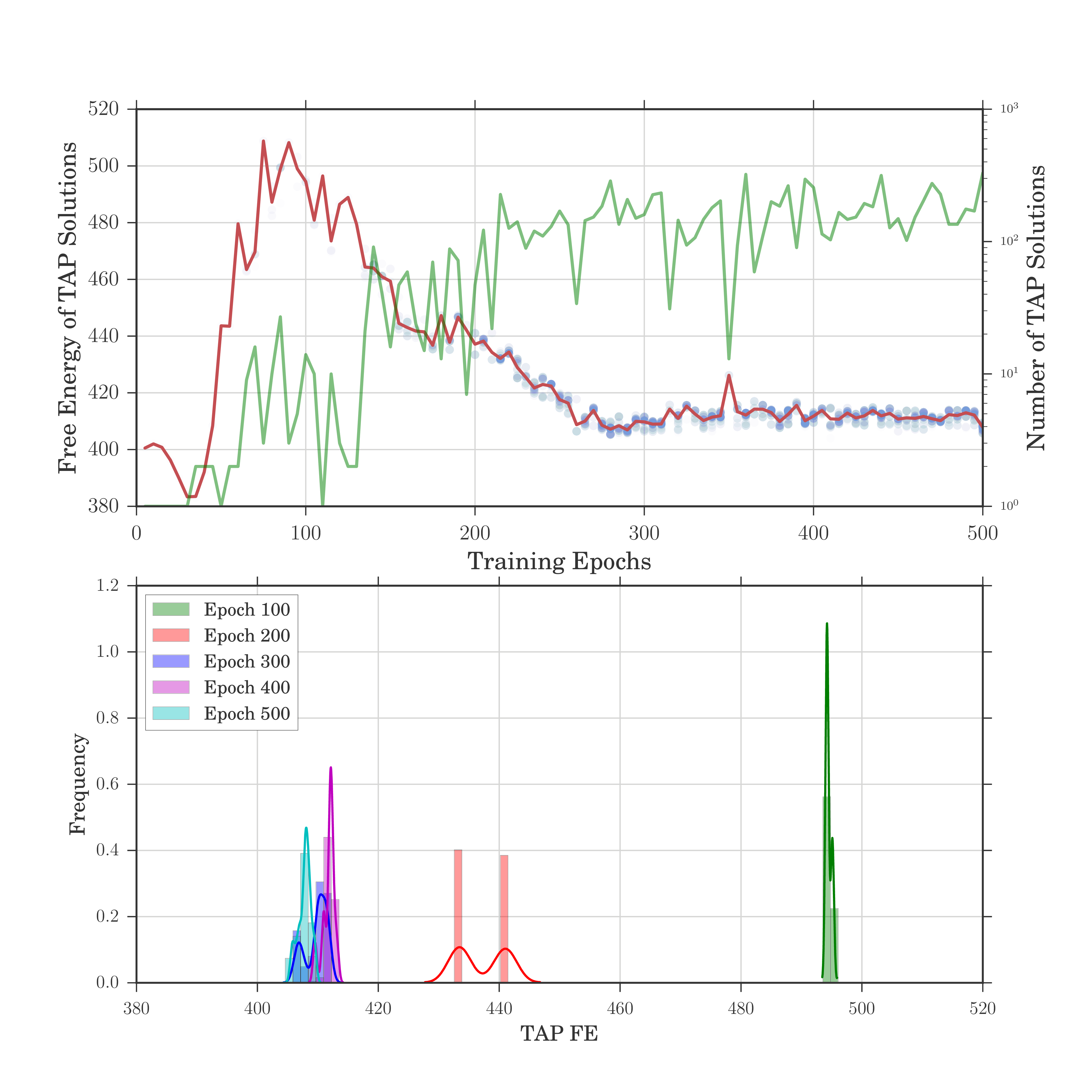}\\
  (a) \emph{binary-MNIST} & (b) \emph{real-MNIST} & (c) \emph{CBCL} 
  \end{tabular}
  \caption{\label{fig:fedist} 
           Distribution of free energy estimates of TAP solutions as a function 
           of training epochs for the three different datasets. In the case of
           the two MNIST experiments, the number of hidden units is the same
           $N_h = 100$ and $10,000$ samples drawn from the training data are
           used as initial conditions. For \emph{CBCL}, $2,400$ training samples
           are used. \textbf{Top Row:} TAP Free
           energy for all unique TAP solutions (\emph{transparent blue dots}), 
           and the Helmholtz free energy estimate via uniform averaging 
           (\emph{red line}). The number of unique TAP solutions is also given 
           (\emph{green line}). \textbf{Bottom Row:} Detail of 
           TAP free energy distributions for slices of training. Histograms are
           given as bars, while kernel density estimates of the TAP free energy
           distribution are given as curves.}          
\end{figure}
\end{center}
\twocolumngrid

\subsection{Probing the GRBM}
Given the deterministic nature of the TAP framework,
it is possible to investigate the structure of the modes 
which a given set of GRBM parameters produces in the free 
energy landscape. Understanding the nature and concentration of 
these modes gives us an intuition on the representational power
of the GRBM. 

To date, observing the modes of a given GRBM model could
only be approached via long-chain sampling. Given enough sampling chains
from a diverse set of initial conditions, thermalizing these chains
produces a set of samples from which one could attempt to derive 
statistics, such as concentrations of the samples in their high-dimensional
space, to attempt to pinpoint the likely modes in the model. However, the number
of required chains to resolve these features 
increases with the dimensionality of the space and the number of potential 
modes which might exist in the space. Because of this, the numerical 
evaluation we carry out here would be impractical with sampling techniques.

The r-BP and mean-field models of the RBM
allow us to directly obtain the modes of the model
by running inference to solve the direct problem. Given a diverse set 
of initial conditions, such as a given training dataset, running r-BP or
TAP provides a deterministic mapping between the initial conditions drawn
from the data, as in \eqref{eq:init_a}--\eqref{eq:init_c},
and the ``nearest'' solution of the TAP free energy. If the initial point was
drawn from the dataset, then this solution can be interpreted as the RBM's 
best-matching internal representation for the data point.

If a large number of structurally diverse 
data points map to a single solution, then this may be an indicator 
that the GRBM parameters are not sufficient to model the diverse 
nature of the data, and perhaps further changes to the model parameters
or hyper-parameters are required. Conversely, if the number of 
solutions explodes, being roughly equivalent to the number of initial
data points, then this indicates a potential spin-glass phase, that the
specific RBM is over-trained, perhaps memorizing the original 
data samples during training. Additionally, when in such a phase, the large
set TAP solutions may be replete with spurious solutions which convey very 
little structural information about the dataset.
In this case, hyper-parameters of the
model may need to be tuned in order to ensure that the model 
possess a meaningful generalization over the data space.

To observe these effects, we obtain a subset of the TAP solutions by
initializing the TAP iteration with initial conditions drawn from the data
set, running the iteration until convergence, and then counting the 
unique TAP solutions. We present some measures on these solutions in 
Fig. \ref{fig:freeEnergies}. Here, we both count the number of unique 
TAP solutions, as well as the distribution of the TAP free energy
over these solutions, across training epochs. There are a few common features
across the tested datasets. First, the early phase of training shows
a marked increase of the TAP free energy, which then gradually declines as
training continues. Comparing the point of inflection in the TAP free energy
against the normalized TAP log-likelihood shown in Fig. \ref{fig:bmnist-tapLL}
shows that the early phase of GRBM training is dominated by the reinforcement
of the empirical moments of the training data, 
with the GRBM model correlations playing a small  
role in the the gradient of \eqref{eq:Wgrad}. This makes sense, as the random initialization of 
$W \sim \mathcal{N}(0,\sigma)$ for $\sigma \approx 10^{-3}$ implies that 
the hidden units are almost independent of the training data. Thus, the 
TAP solutions at the early stage of learning are driven by, and correlated with,
the local potentials on the hidden and visible variables. 

The effect of this 
influence is that the TAP free energy landscape possesses very few modes in the
data space. Fig. \ref{fig:fedist} shows this very clearly, as the number of 
TAP solutions starts at $1$ and then steadily increases with training. 
Because the positive data-term of \eqref{eq:Wgrad} is dominant, the GRBM 
parameters do not appear to minimize the TAP free energy, as we would expect. 
However, as more TAP solutions appear, the data and model terms of the gradient
become balanced, and the TAP free energy is minimized. It is at this point of 
inflection that we see leveling off of the normalized TAP log-likelihood.

Second, we observe free-energy bands in the TAP solutions. This feature is
especially pronounced in the case of the \emph{binary-MNIST} experiment. Here,
at all training epochs, there exist two significant modes in the free 
energy distribution over the TAP solutions. We see this effect more clearly in 
the training-slice histograms shown in the bottom row of Fig. \ref{fig:fedist}(a).
In the case of the \emph{real-MNIST} experiment, we see that the 
free energy distributions do not exhibit such tight banding, but they do show 
the presence of some high- and low-energy solutions which persist across 
training. The main feature across experiments is the multi-modal structure 
of the free energy distribution. Finally, we note that for both \emph{real-MNIST} 
and \emph{binary-MNIST}, in the case of $N_h = 100$, we don't empirically 
observe an explosion of TAP solutions, a potential indicator a spin-glass phase, 
since the proportion of unique TAP solutions to the initial data points remains
less than 10\%.

\begin{center}
\begin{figure}
\centering
  \includegraphics[width=0.95\textwidth]{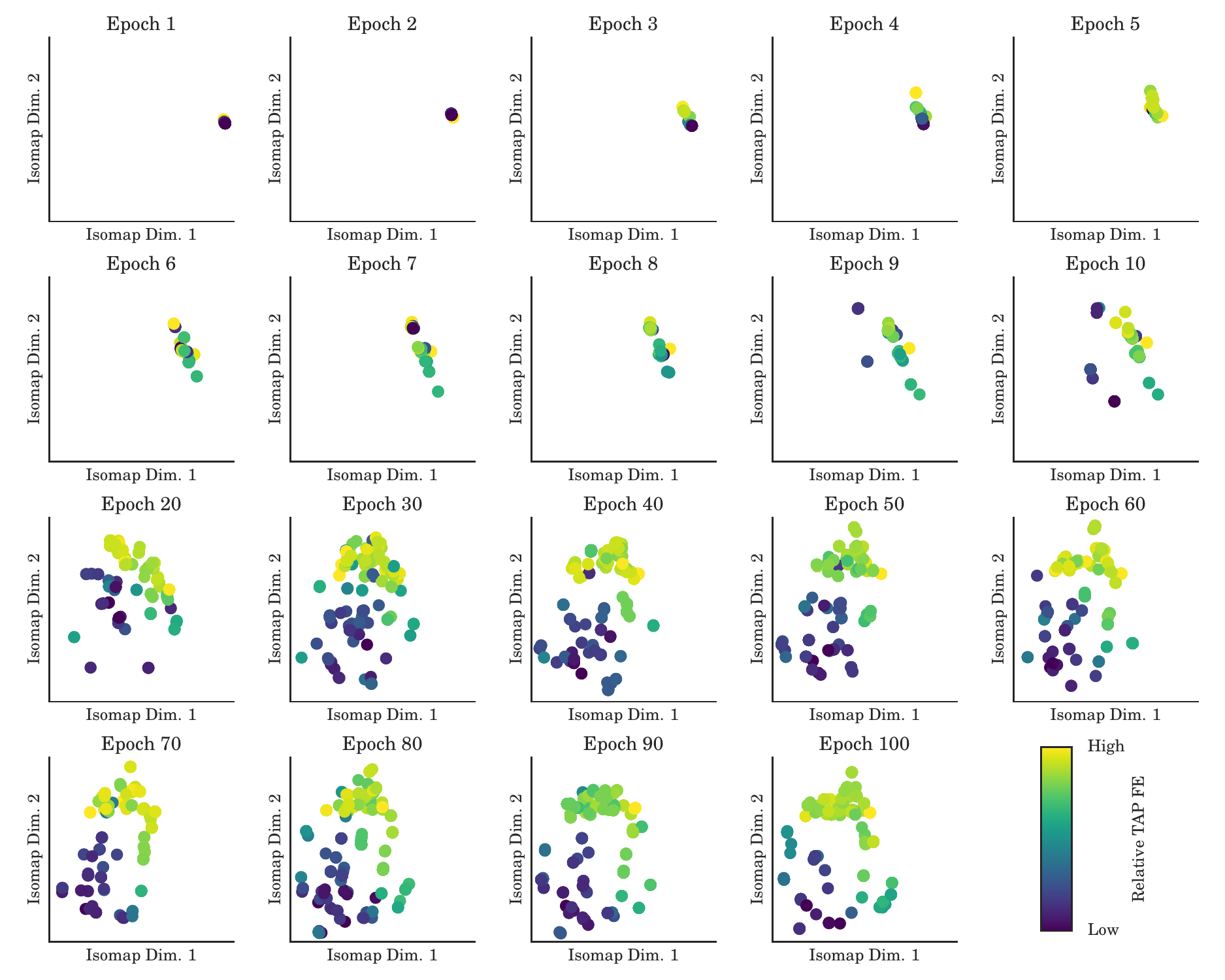}\\
  \caption{\label{fig:solnvis} 
           Isomap visualization of TAP solutions for \emph{binary-MNIST}
           over training epochs for $N_h = 100$. All TAP solutions are
           mapped to the same two-dimensional embedding via an Isomap transform
           fitted to the TAP solutions of Epoch 100. The embedding is performed 
           on both the hidden and visible inferred expectations, $\mathbf{a}^{\rm v}$
           and $\mathbf{a}^{\rm h}$.
           The color mapping
           corresponds to the TAP free energy values of each solution, with the
           range of colors normalized between the minimum and maximum free
           energies of the solution at each training epoch.
          }          
\end{figure}
\end{center}

In order to investigate whether the modes in the TAP free energy distributions 
are randomly assigned over configuration space, or exist in separate continuous
partitions of the configuration space, we need to look at the proximity of the
solutions in the configuration space. Because this space cannot be observed in
its ambient dimensionality, we project the configuration space into a 
two-dimensional embedding in Fig. \ref{fig:solnvis}. Here, we utilize the
well known Isomap \cite{TSL2000b} algorithm for calculating a two-dimensional
manifold which approximately preserves local neighborhoods present in the 
original space. Using this visualization we observe that as training progresses
the assignment of high and low free energy to TAP solutions does not appear 
random in nature, but seems to be inherent to the structure of the solutions
themselves, that is, their location in the configuration space. Additionally,
in Fig \ref{fig:solnvis} we can see the progression from few TAP solutions
to many, and how they spread across the configuration space. It is interesting
to note how the solutions start from a highly correlated state and then 
proceed to diversify. 

\begin{center}
\begin{figure}
  \begin{tabular}{c c c c}
    \multicolumn{4}{c}{(a) \emph{binary-MNIST} ($N_h = 100$)}\\
    Epoch 1 & Epoch 3 & Epoch 25 & Epoch 100 \\
    \includegraphics[width=0.22\textwidth]{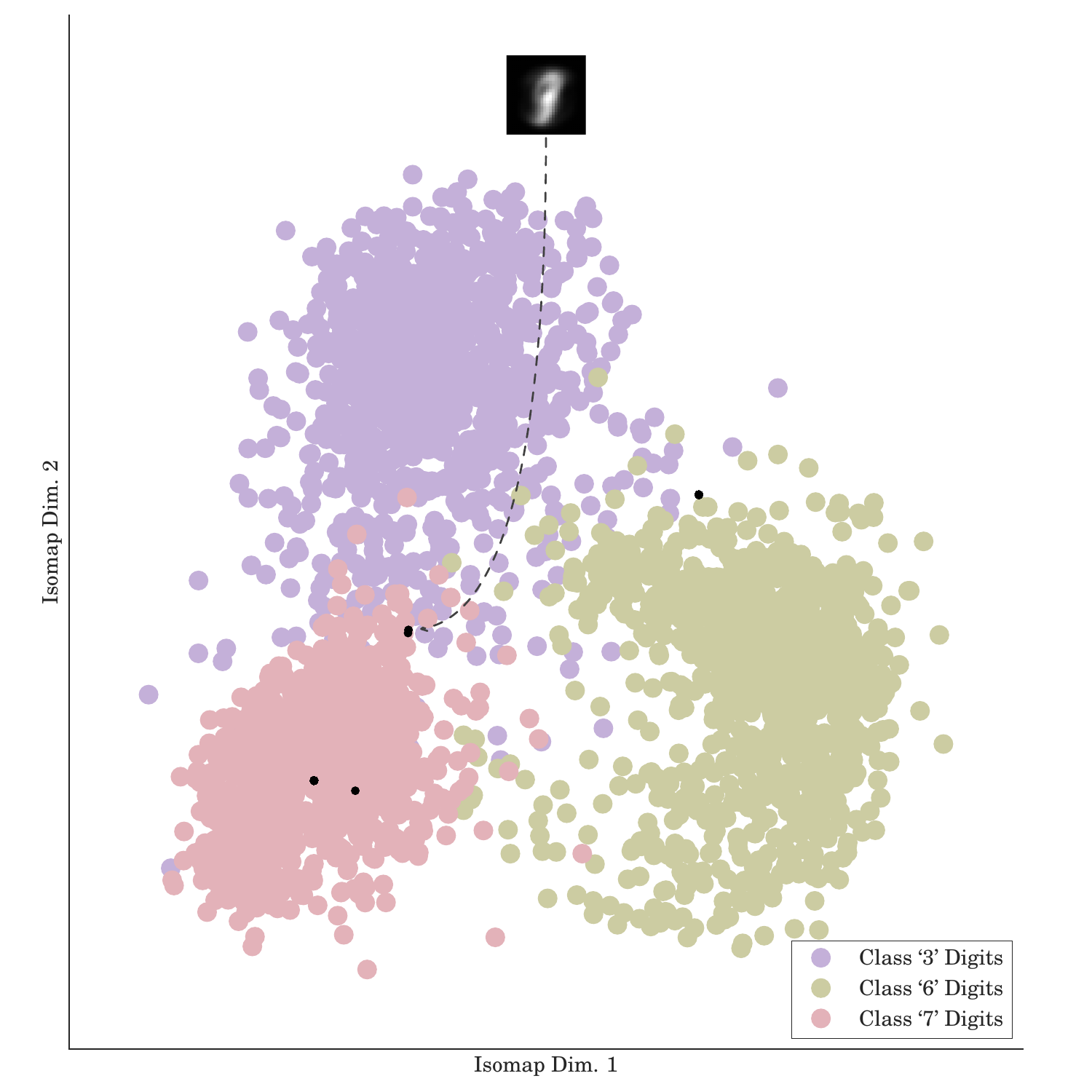} &
    \includegraphics[width=0.22\textwidth]{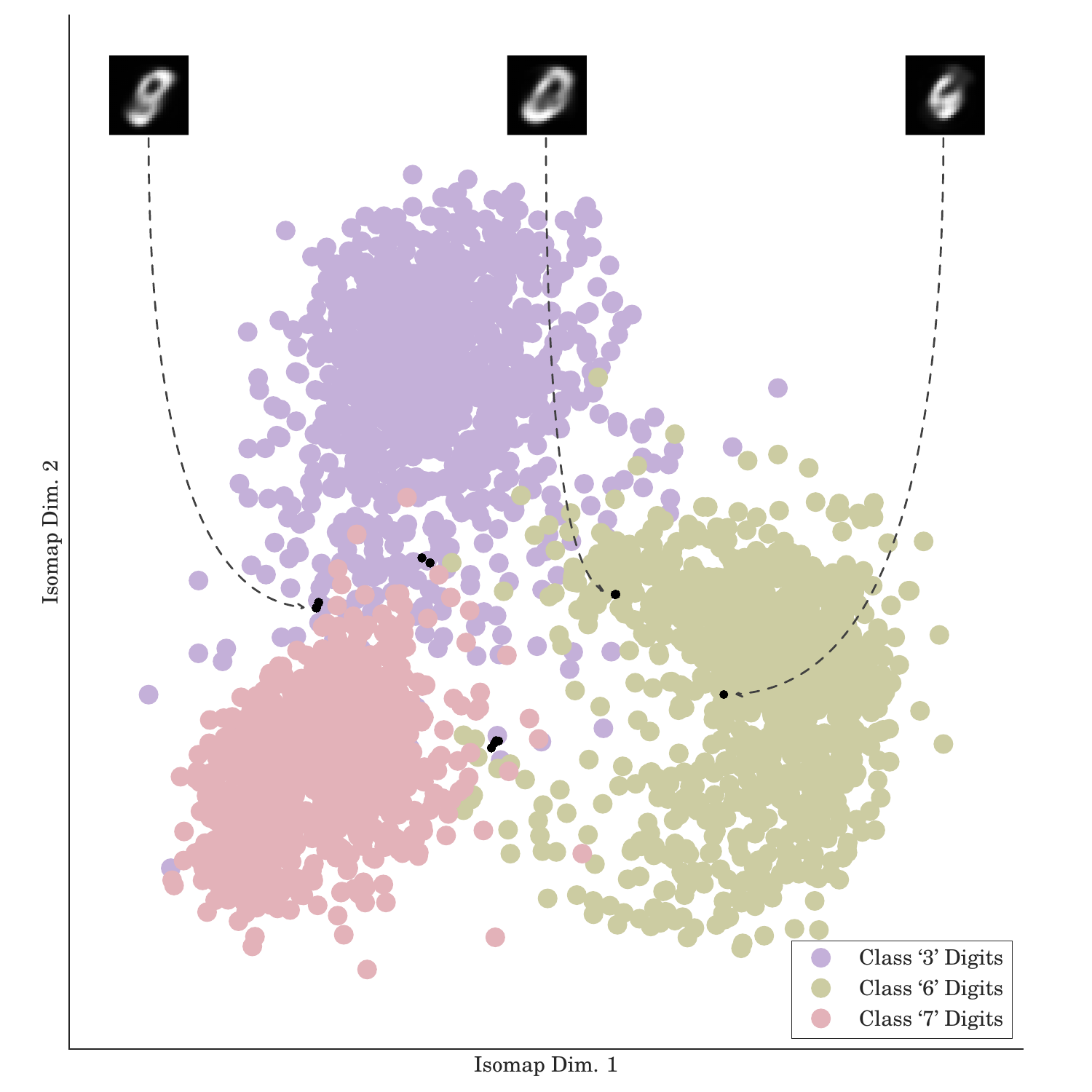} &
    \includegraphics[width=0.22\textwidth]{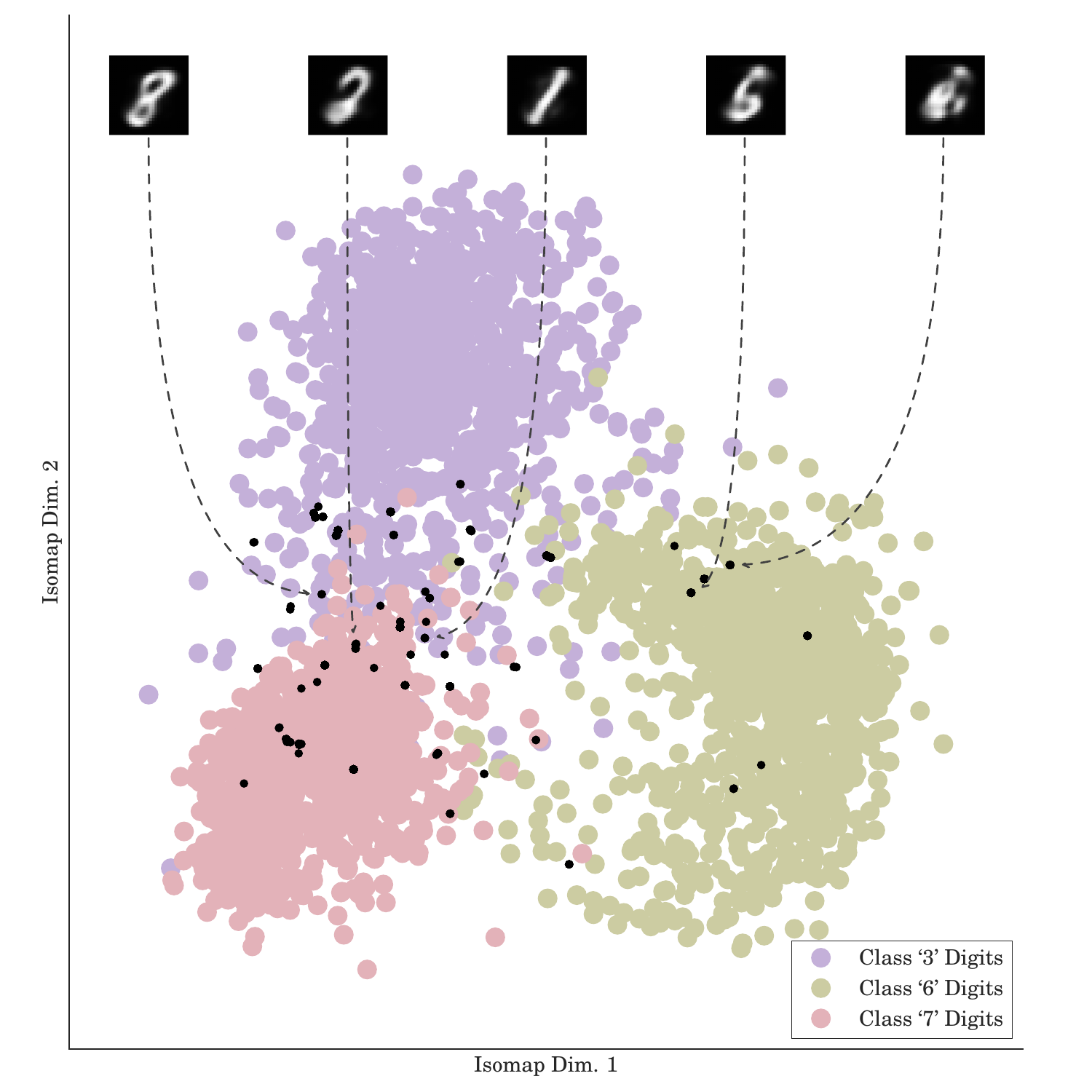} &
    \includegraphics[width=0.22\textwidth]{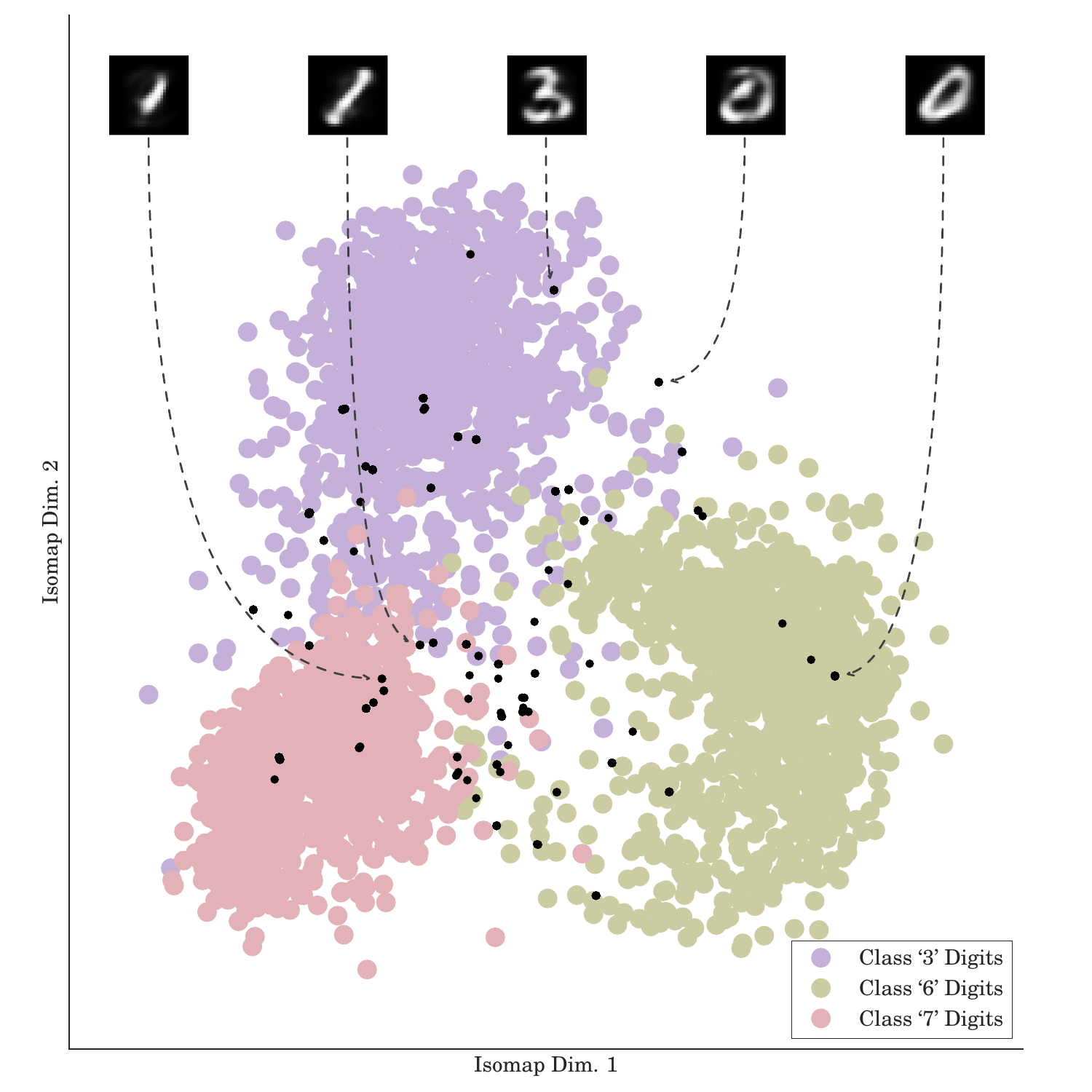} \\
    \multicolumn{4}{c}{(b) \emph{real-MNIST} ($N_h = 500$)}\\
    Epoch 1 & Epoch 3 & Epoch 25 & Epoch 100 \\
    \includegraphics[width=0.22\textwidth]{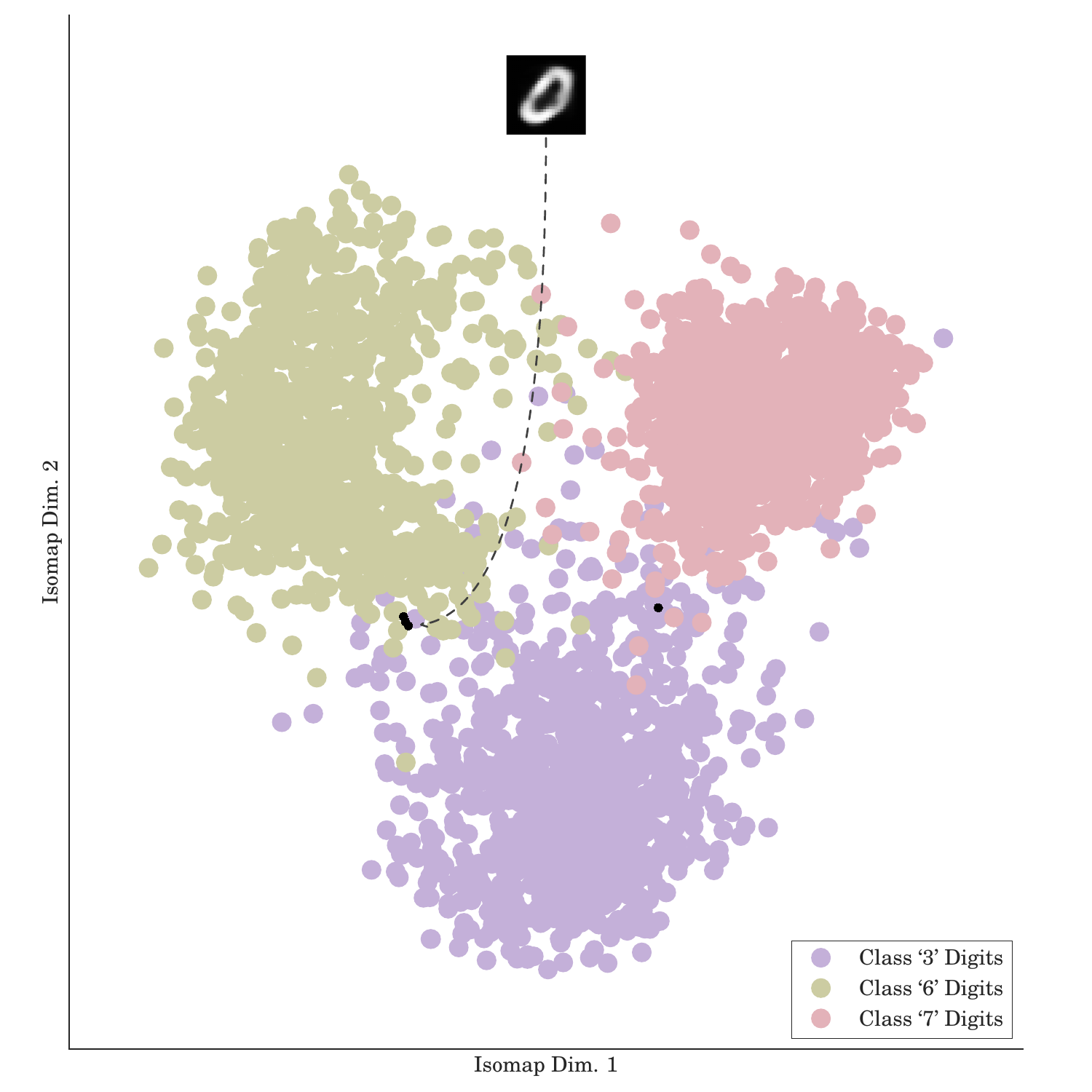} &
    \includegraphics[width=0.22\textwidth]{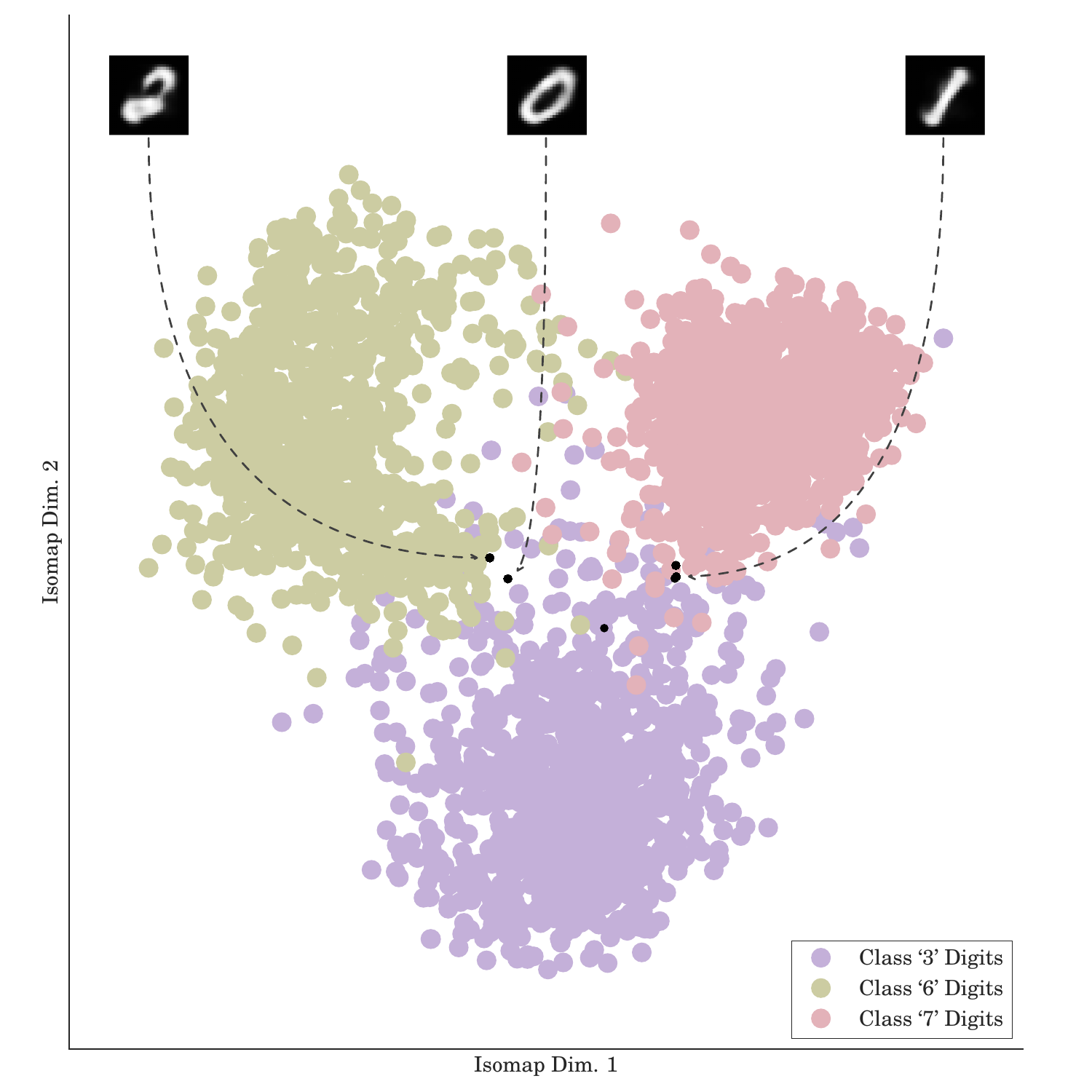} &
    \includegraphics[width=0.22\textwidth]{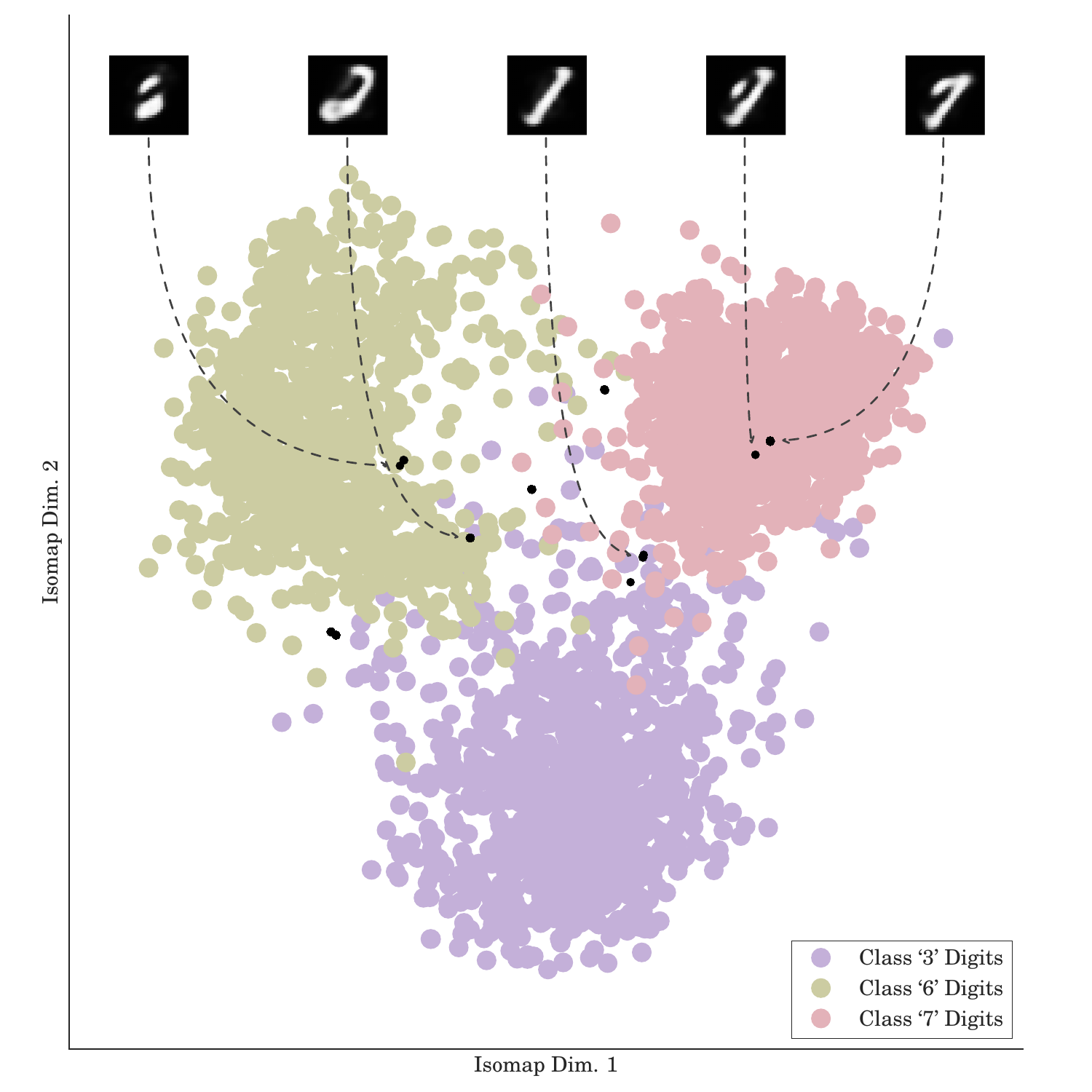} &
    \includegraphics[width=0.22\textwidth]{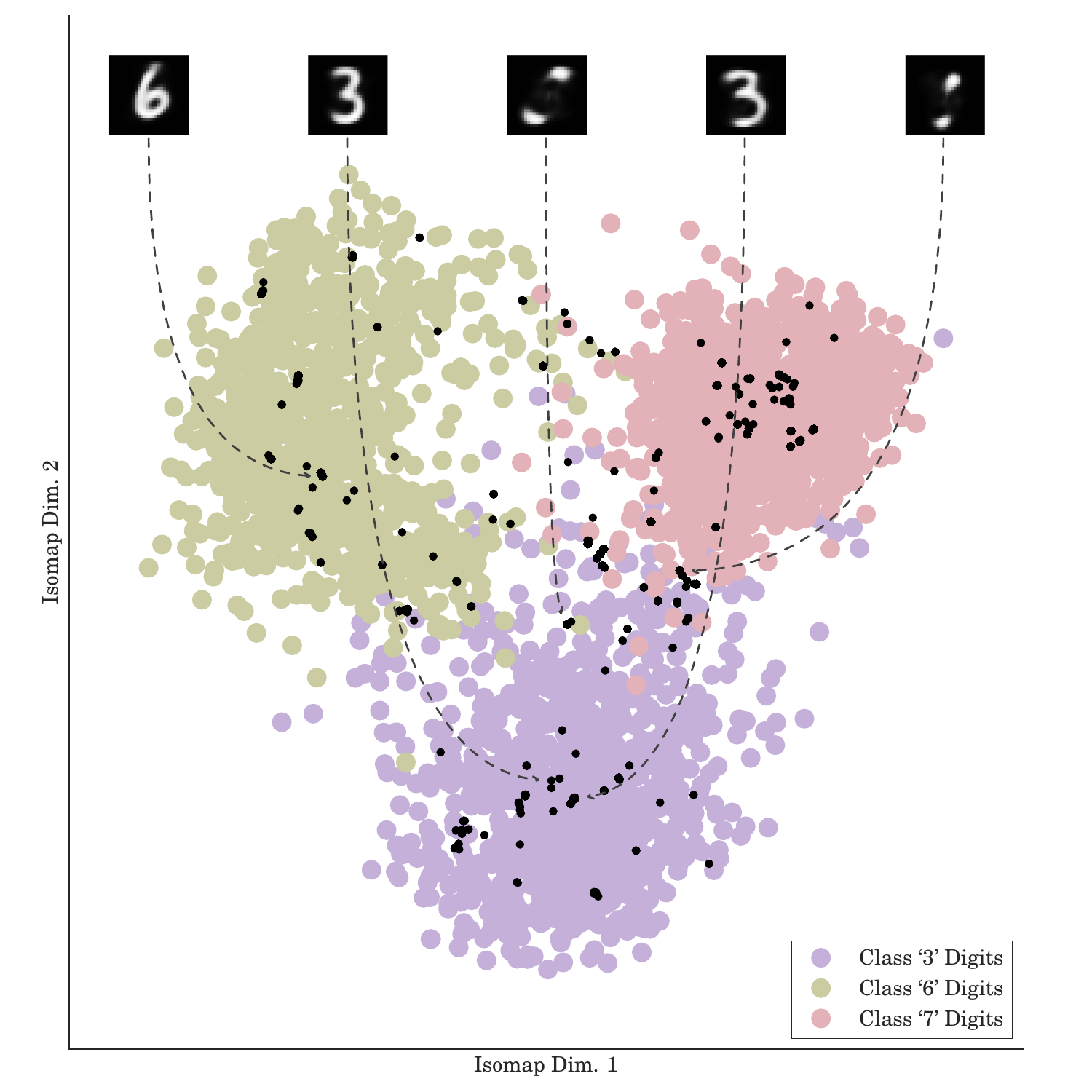} \\
    \multicolumn{4}{c}{(c) \emph{CBCL} ($N_h = 256$)}\\
    Epoch 5 & Epoch 100 & Epoch 200 & Epoch 500 \\
    \includegraphics[width=0.22\textwidth]{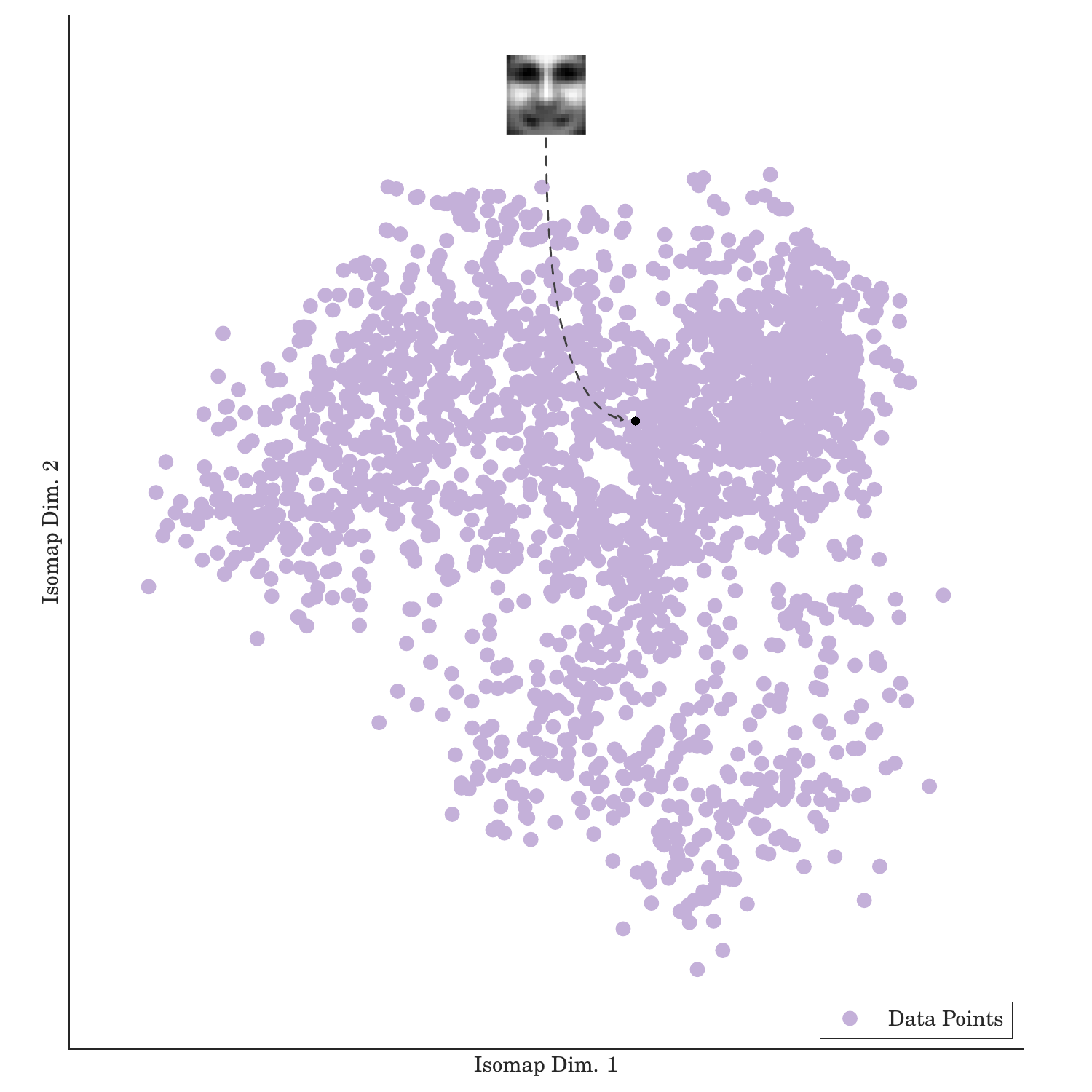} &
    \includegraphics[width=0.22\textwidth]{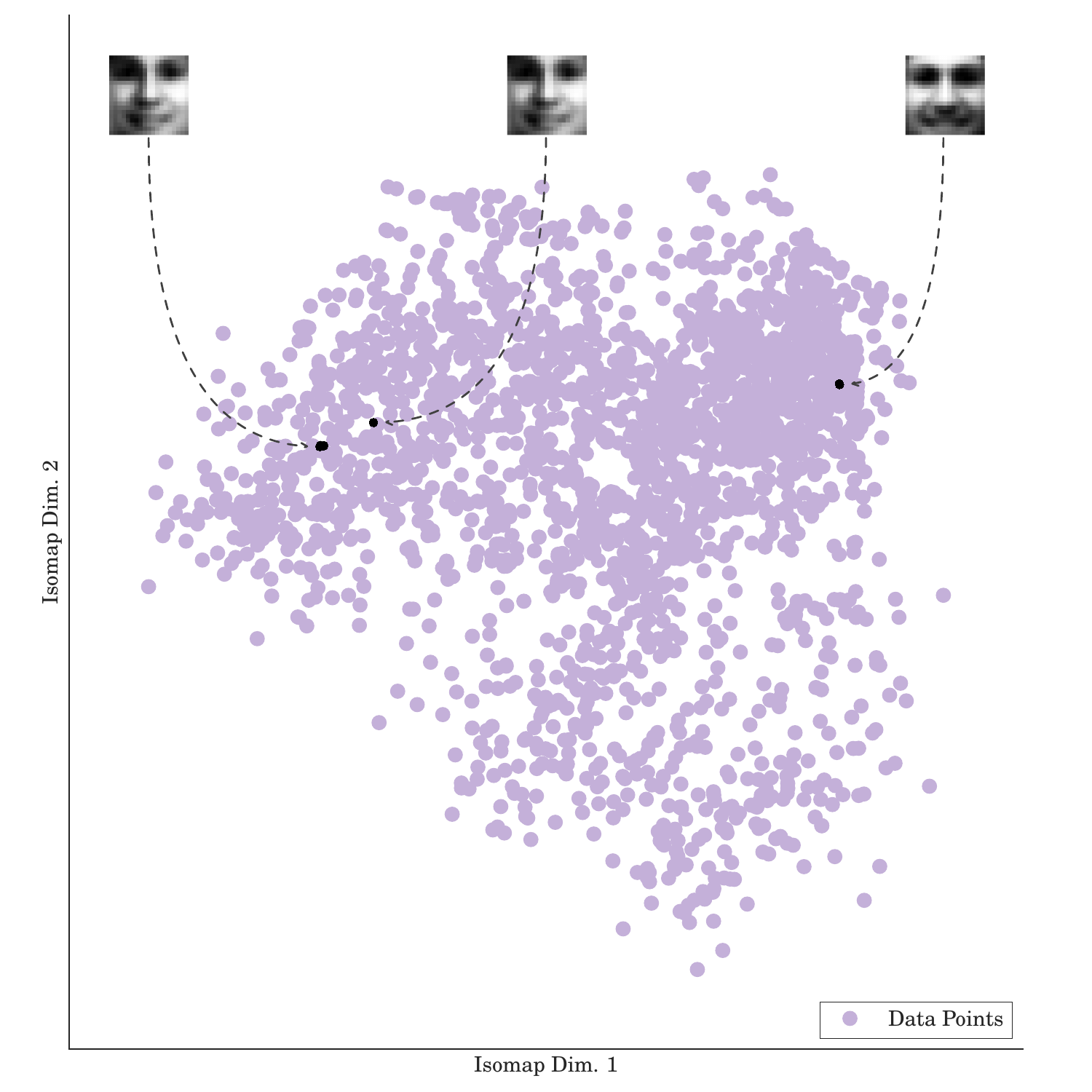} &
    \includegraphics[width=0.22\textwidth]{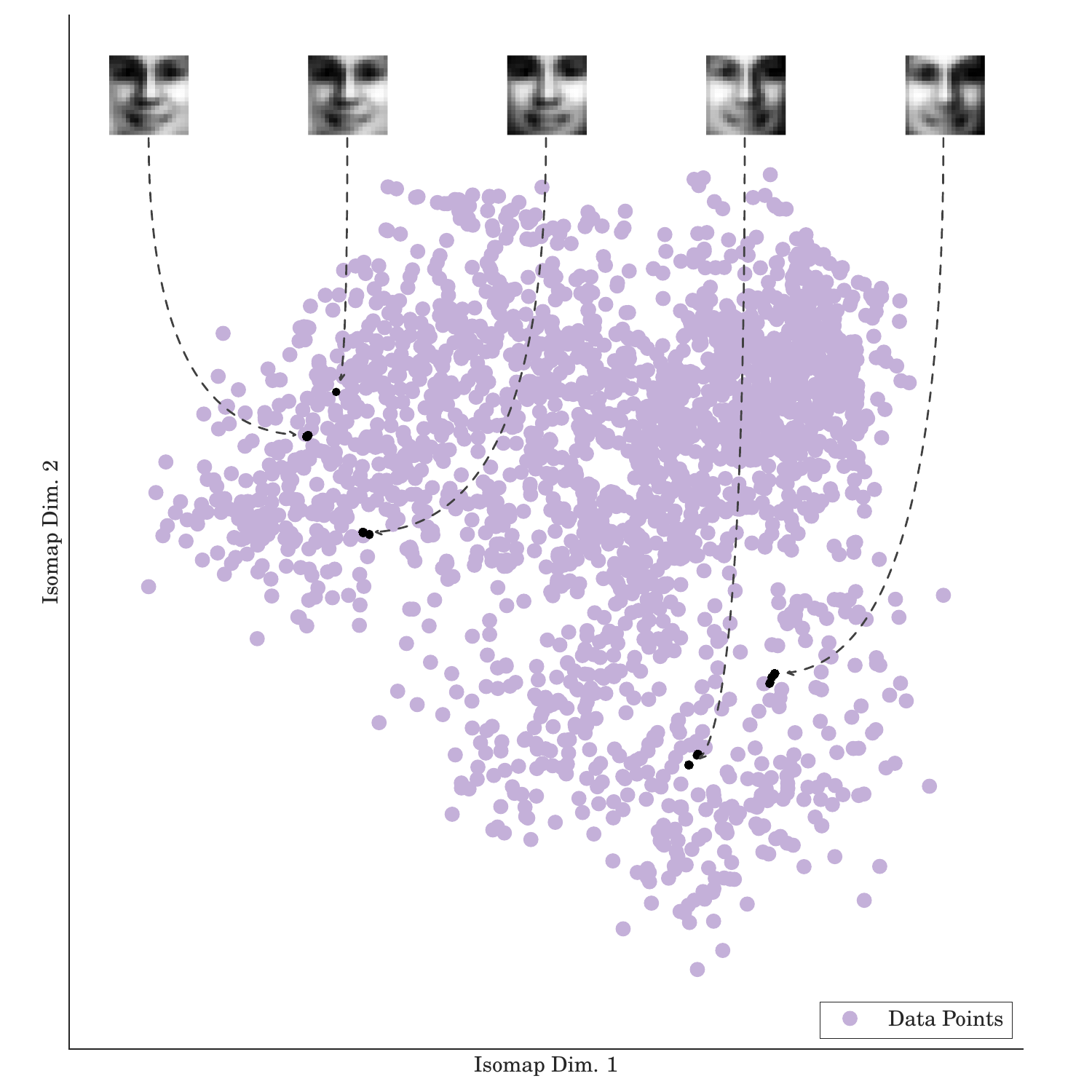} &
    \includegraphics[width=0.22\textwidth]{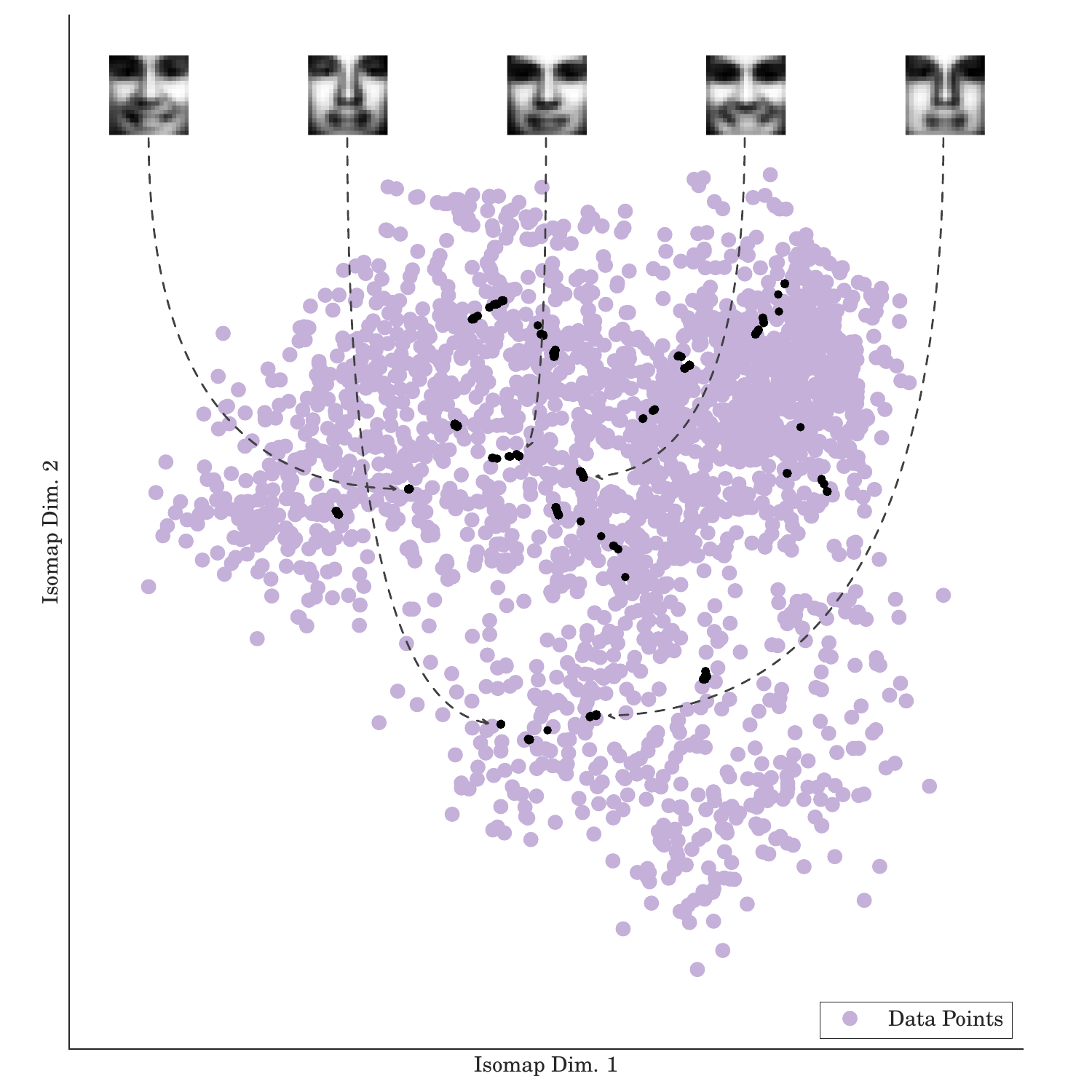} \\
  \end{tabular}
  \caption{Comparison of initial conditions for TAP equilibration (\emph{colored dots})
           compared to 
           converged TAP solutions (\emph{black dots}) for the tested datasets
           at different stages of training. For each dataset, a 
           two-dimensional Isomap 
           embedding is calculated over the initialization data. Subsequently,
           the magnetizations of the TAP solutions are embedded in the same 
           space. In each case, all initial variances are set to 0, as in Eq. 
           $\ref{eq:init_c}$. Also, a random selection of TAP solution 
           magnetizations are chosen to provide some context for the 
           representations the RBM is learning.
           \textbf{(a) \emph{binary-MNIST:}}~%
           Here, the $\sim$3,100 digits corresponding to
           the classes `3', `6', and `7' are drawn from the first 10,000 training
           samples of the binarized MNIST dataset as initializations. A reduced
           set of labels is used for readability. 
           \textbf{(b) \emph{real-MNIST:}}~%
           The same initializations are used as in (a),
           however, the initializations are not binarized. 
           \textbf{(c) \emph{CBCL:}}~%
           All available training face images are used as 
           initializations.} 
  \label{fig:compareInitSolns}
\end{figure}
\end{center}

We can also observe the TAP solutions with respect to the initializations 
which produced them, as shown in Fig. \ref{fig:compareInitSolns}. In 
these charts, we use a similar approach as Fig. \ref{fig:solnvis}, mapping
all high-dimensional data points, as well as TAP magnetizations, into a 
2D embedding using Isomap. This allows us to see, in an approximate way, how
the TAP solutions distribute themselves over the data space. We also show
how the number of TAP solutions grows from few to many over training, 
and how they maintain
a spread distribution over the data space. This demonstrates how the training 
procedure is altering the parameters of the model so as to place TAP solutions
within dense regions of the data space. For the sake of clarity, we have not
included lines indicating the attribution of an initial data point to its
resultant TAP solution. However, as training progresses, one sees that the TAP
solutions act as attractors over the data space, clustering together 
data points which the TAP machine recognizes as similar.

\subsection{Inference for Denoising}
\label{subsec:denoising}
Serving as a prior for inference is one particular use case
for the TAP machine interpretation of the GRBM. As a simple demonstration, we 
turn to the common signal processing task of \emph{denoising}. Specifically,
given a planted signal, one observes a set of noisy observations which are 
measures of the true signal corrupted by some stochastic process. Denoising
tasks are ubiquitous in signal processing, both at an analog level (e.g.
additive and shot noise), and at the level of digital communications (e.g. 
binary symmetric and erasure channels). The goal of this task is to produce the
most accurate estimate of the unknown signal. In the analog case, this 
may be a measure of mean-square-error (MSE) between the estimate and the
true
signal. In the binary case, this may be a measure of accuracy, counting the 
number of incorrect estimates, or some other function of the binary confusion
matrix, such as the F1-score or Matthews correlation coefficient (MCC).

For a fixed set of observations and channel parameters, if we assume that the 
original signal was drawn from some unknown and  intractable generating
distribution,
then as we construct more and more accurate tractable approximate priors, 
the more accurately we can construct an estimate of the original signal.

In other words, the more we know about the structure and content
of the unknown signal \emph{a priori}, the closer our estimate can be. 
Often, as in the
case of wavelet-based image denoising, statistics are gathered on the transform
coefficients of particular images classes and heuristic denoising approaches
are designed by-hand accordingly \cite{SS2002}.
By-hand derivation of denoising algorithms works well in practice owing to its
generality. Specific \emph{a priori} information about the original signal is 
not required, beyond its signal class (e.g. natural images, human speech, radar
return timings). However, meaningful features must be assumed or investigated by
practitioners before successful inference can take place. 

\clearpage
\onecolumngrid
\twocolumngrid

\clearpage
\onecolumngrid
\twocolumngrid
\clearpage

\subsubsection{Denoising the Binary Symmetric Channel}
\begin{center}
\begin{figure}
  \centering
  \includegraphics[width=0.45\textwidth]{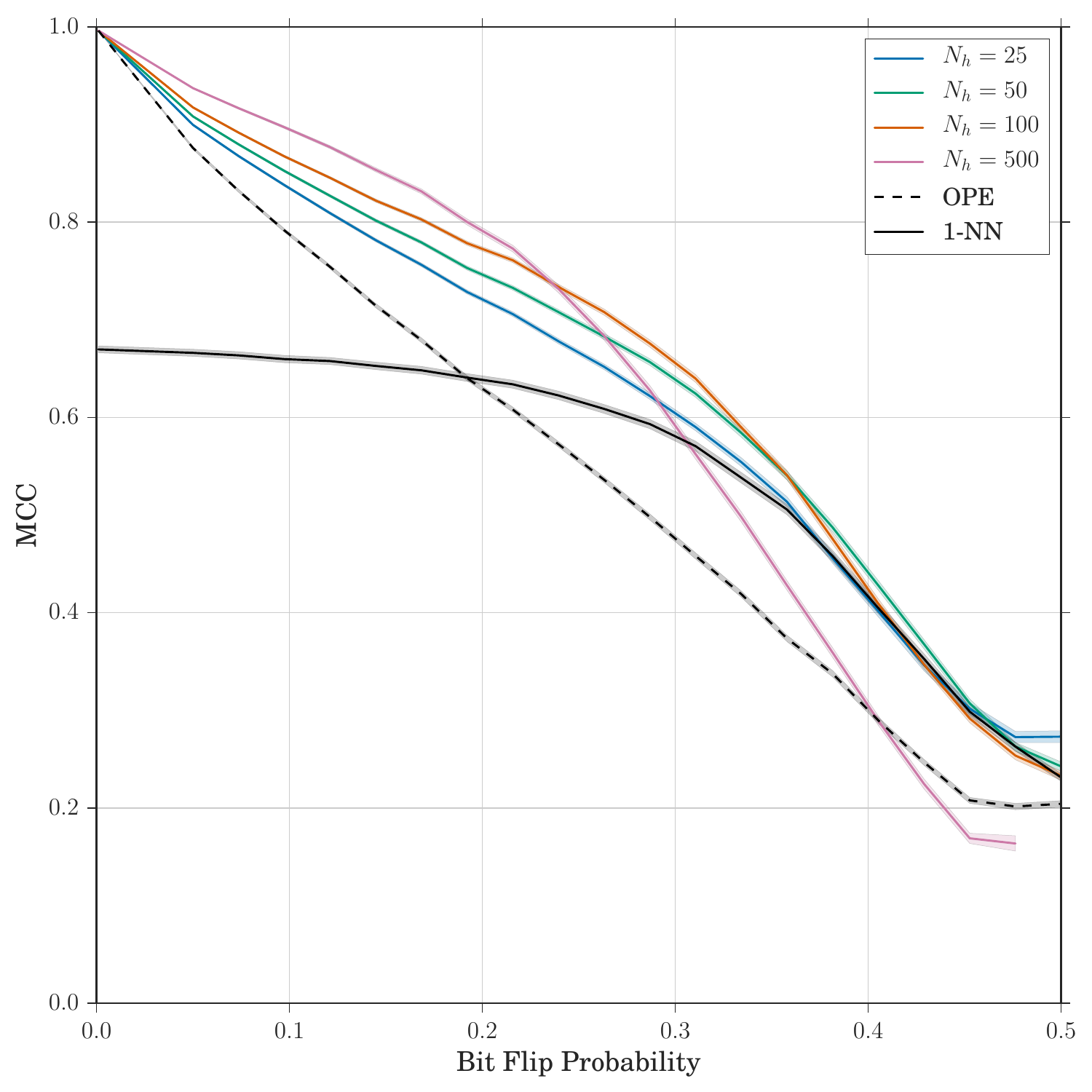}\\  
  \caption{\label{fig:bmnist-denoise}
    Average denoising performance for reconstruction from 
    bit-flip errors on \emph{binary-MNIST} over the probability of a bit to 
    be flipped. Denoising via inference on the binary-binary
    RBM is denoised by the varying numbers of hidden units 
    ($N_h = \bra{25,50,100,500}$).
    Also shown as baseline comparisons are
    the OPE given the empirical factorized 
    magnetizations at each site (\emph{dashed black}), and a 1-NN
    matching from the training set (\emph{solid black}). All experiments were
    run over the same 1,000 data samples drawn from the held-out test set and
    compared using the MCC. Binary estimates are obtained for the OPE and 
    TAP inferred estimates by rounding the resulting magnetizations.}
\end{figure}
\end{center}

For binary denoising problems, we assume a binary symmetric channel (BSC) 
defined in the following manner. Given some binary signal 
$\mathbf{x}\in\bra{0,1}^N$, we observe the signal 
$\mathbf{y}\in\bra{0,1}^N$ as $\mathbf{x}$ with 
independent bit flips occurring with probability $p$. This gives the following
likelihood at each observation,
\begin{equation}
  P(\mathbf{y}|\mathbf{x}) = (1-p)\prod_{i} \p{\frac{p}{1-p}}^{\delta_{x_i,y_i}},
\end{equation}
which can be shown to have the equivalent representation as a Boltzmann 
distribution,
\begin{equation}
  P(\mathbf{y}|\mathbf{x}) = \frac{1}{Z(\mathbf{y})} e^{\sum_i D_i x_i},
\end{equation}
where $D_i \defas \ln \frac{p}{1-p}(2 y_i - 1)$. For a given prior distribution 
$P(\mathbf{x})$, the posterior distribution is given by Bayes' rule,
\begin{equation}
  P(\mathbf{x}|\mathbf{y}) = 
  % \frac{e^{\sum_i B_i x_i}}{Z(\mathbf{y})} P(\mathbf{x})\s{\prod_i P(y_i | x_i = 0) + P(y_i | x_i = 1)}^{-1}.
  \frac{e^{\sum_i D_i x_i}P(\mathbf{x})}{\sum_{\mathbf{x}} e^{\sum_i D_i x_i}P(\mathbf{x}) }.
\end{equation}
By assuming a factorized $P(\mathbf{x}) = \prod_i m_i^{x_i}(1-m_i)^{1-x_i}$, 
where $m_i$ might be per-site empirical averages obtained from available 
training data,
the posterior factorizes and
we can construct the Bayes-optimal pointwise estimator (OPE) as the average 
$\ang{x_i}_{P(x_i|y_i)}$ which is just $P(x_i = 1 | y_i)$ for our binary problem. Thus, the OPE at each site $x_i$ is given as 
\begin{align}
  P(x_i = 1| y_i) 
  &= \frac{1}{1 + \p{\frac{1 - m_i}{m_i}}\times\p{\frac{1-p}{p}}^{2y_i -1}},\notag \\
  &= {\rm sigm}\p{\ln \frac{m}{1-m} + (2y_i -1)\ln\frac{p}{1-p}}.
  \label{eq:ope}
\end{align}
For a given dataset, the OPE gives us the best-case performance using only 
pointwise statistics from the dataset, namely, empirical estimates of the 
magnetizations $m_i$. We can see from Eq. \eqref{eq:ope} that
the OPE either returns the observations, in the case of $p = 0$, or the 
prior magnetizations $m_i$, in the case $p = 0.5$. In this case of 
complete information loss, the worst case performance is bounded according to
the deviation of the dataset from its mean. We present the performance of the 
OPE in Fig. \ref{fig:bmnist-denoise} for the \emph{binary-MNIST} dataset.
This makes the OPE a valuable baseline comparison and
sanity check for the GRBM approximation of $P(\mathbf{x})$. As the GRBM model
takes into account both pointwise and pairwise relationships in the data, 
a properly trained GRBM should provide estimates \emph{at least as good as} the
OPE.

The $k$-Nearest-Neighbor ($k$-NN) algorithm represents a different
heuristic 
approach to the same problem \cite{GM2006}.
In this case, the noisy measurements are compared
to a set of exemplars: the training dataset. 
Then, according to some distance metric such as MSE or correlation, 
one finds the $k$ exemplars with minimal distance to the noisy 
observations to serve as a basis for recovering the original binary
signal.
One can use some arbitrary approach for fusing these exemplars together into
the final estimate, but the simplest case would be a simple average. In the 
case that $k \rightarrow \infty$, the performance when using averaging
is again bounded by the empirical magnetizations. 
In the other limit of $k = 1$, the estimate is 
simply the nearest exemplar. It is hard to show the limiting performance of
this approach, as it is dependent on the distances, in the chosen metric,
between the exemplars and the observations, as well as the interplay between
the noise channel and the distance metric. 

However, it can be seen directly
that this approach is non-optimal, as this approach will not
yield the true signal at $p = 0$ unless the true signal is itself 
contained within the training data.
We show the performance for $k = 1$ in Fig. \ref{fig:bmnist-denoise}. The
advantage of this approach is that it successfully regularizes against noise
as $p \rightarrow 0.5$, as the nearest exemplar is always noise-free and at 
least marginally correlated with the original signal, up to the distance 
metric. Additionally, we see that it performs better than the OPE in the
regime $ p > 0.2$. This can be explained since we can think of the $k$-NN
approach as implicitly, though indirectly, taking into account
higher-order correlations in the dataset by na\"ively returning data exemplars;
all the estimates trivially posses the same arbitrarily complex structure as
the unknown signal.

Using the GRBM, we can hope to capture the best points of both approaches. First,
we hope to perfectly estimate the original signal in the case $p = 0$. Second,
we hope to leverage the pairwise correlations present in the dataset, returning
estimates which retain the structure of the data even as $p \rightarrow 0.5$.
For GRBM denoising of the BSC, we no longer have a factorized posterior. 
Instead, we have the GRBM likelihood given in Eq. \eqref{eq:visgrbm} summed
over the hidden units. Using the definition of the binary prior given in
Appendix \ref{sec:apdx_binary},
\begin{equation}
  P(\mathbf{x}; W, \boldsymbol{U}) \propto
  \sum_{\mathbf{h}} e^{\sum_{ij} x_i \Wij h_j + \sum_i U_i x_i + \sum_j U_j h_j}. 
\end{equation} 
Since both the GRBM and the BSC channel likelihood are written as exponential
family distributions, 
$P(\mathbf{y}|\mathbf{x}) P(\mathbf{x}; W, \boldsymbol{U}) \propto \sum_{\mathbf{h}} e^{\sum_{ij} x_i \Wij h_j + \sum_i (D_i+U_i) x_i + \sum_j U_j h_j}$. Finding the averages $\ang{x_i}$ for this model simply consists in 
running the TAP-based inference of Alg. \ref{alg:rbmInf} for the modified
visible binary prior $\mathcal{B}(x_i;U_i + D_i)$. One heuristic caveat of this
approach is that we must take into account the multi-modal nature of the TAP
free energy. Since we must initialize somewhere, and the resulting inference 
estimate is dependent upon this initialization, we initialize the inference
with the OPE result. 

We can see that as $p \rightarrow 0$, $D_i \rightarrow \pm \infty$ and 
the highest probability configuration becomes observations. So, in the limit, we
are able to obtain the true signal, just as the OPE, especially since we 
initialize within the well of this potential. This is shown for binary RBMs 
trained with varying numbers of hidden units in Fig. \ref{fig:bmnist-denoise}.
In every case, for $p = 0$, the true signal is recovered. In the case of 
$N_h = \bra{25,50,100}$, we see that the TAP inference on the binary RBM always
outperforms the OPE. Additionally, we see that in each of these cases, the 
performance closely mirrors that of the $1$-NN as $p \rightarrow 0.5$. In the
limit $p = 0.5$, we see that the result of the TAP inference is, essentially,
uncorrelated with the original signal, as in this case, there is no extra
potential present to bias the inference, and the resulting estimate is simply
an arbitrary solution of the TAP free energy. As this closely mirrors the 
exemplar selection in $1$-NN, the MCC curves for the two approaches are similar.

\begin{center}
\begin{figure}
  \centering
  \includegraphics[width=0.45\textwidth]{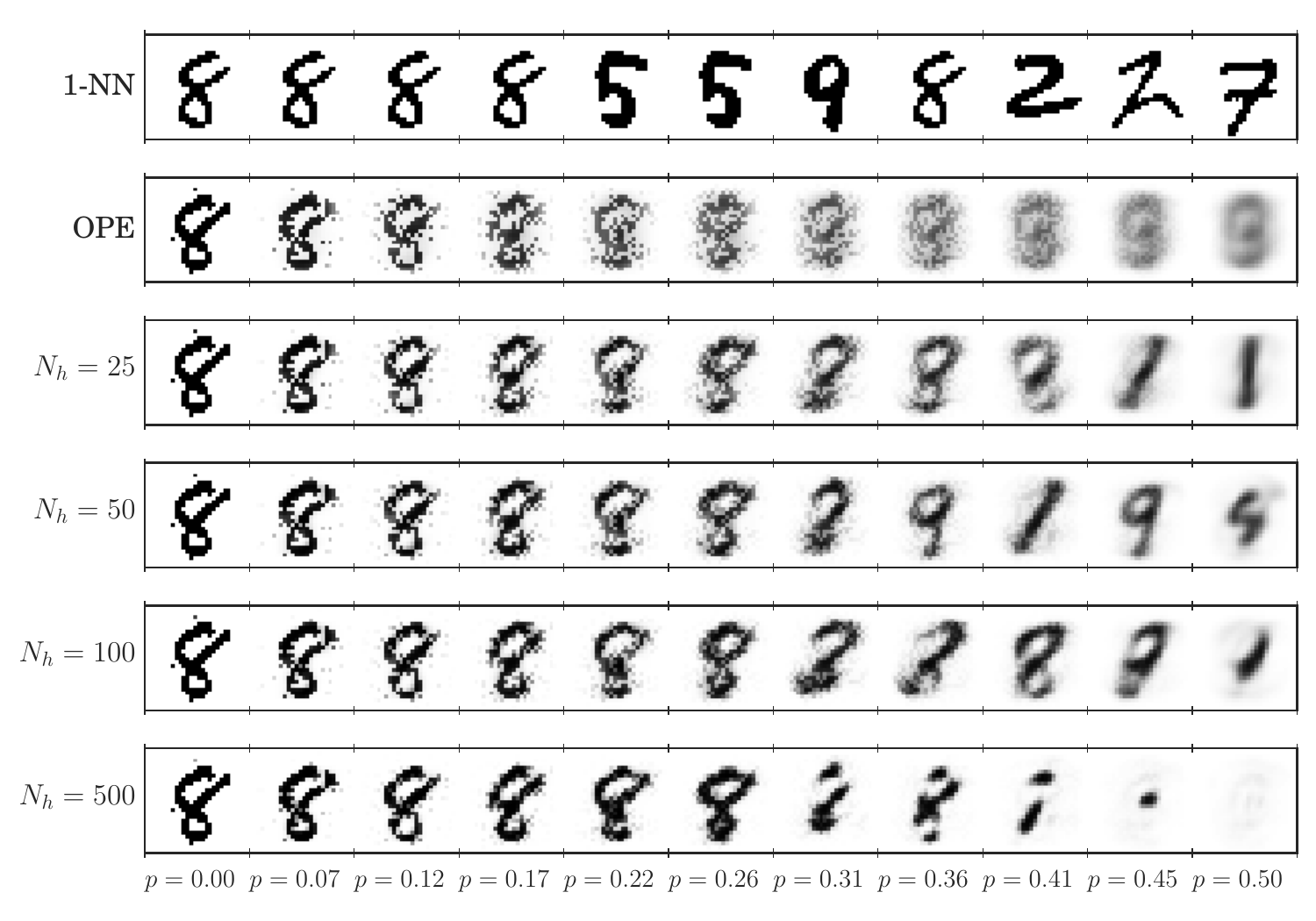}  
  \caption{\label{fig:bmnist-denoise-images}
    Subjective comparison of denoising estimates for a single digit image for 
    $p = \s{0,0.5}$. For the OPE and RBM approaches, the inferred posterior 
    averages $\ang{x_i}$ are shown, rather than the final configurations,
    where white and 
    black represent the values $0$ and $1$, respectively. At each tested value
    of $p$, the same noise realization is used for each method. 
    As $p$ increases,
    for $N_h = \bra{25,50,100}$,
    the TAP inference for the RBM provides estimates which still possess digit
    structure. In the case $N_h = 500,$ the TAP inference gets caught in 
    spurious and undesirable minima as $p$ increases.}
\end{figure}
\end{center}

In the case of $N_h = 500$, we can see that an over-training effect occurs. 
Essentially, at low values of $p$, the TAP inference over the binary RBM is
able to more accurately identify the original signal. However, at a certain
point, owing to the increased number of solutions in the TAP free energy, there
exist many undesirable minima around the noisy solutions, leading to poor
denoising estimates. One can observe this subjectively in Fig. 
\ref{fig:bmnist-denoise-images}, where in the case of $N_h = 500$, the TAP 
inference results in either nearly zero-modes, or in very localized ones. This
would seem to indicate
that landscape of the TAP free energy around the initializations is
becoming more unstable as the density of solutions increases around it.
Additionally, since the TAP free energy landscape was only probed using data
points during training, the clustering of solutions around noisy samples remains
ambiguous. Augmenting the initializations used when calculating the TAP solutions
for the gradient estimate with noisy data samples could help alleviate this 
problem and regularize the TAP free energy landscape in the space of noisy
data samples.

\section{Discussion}
\label{sec:discussion}
In this paper, we have proposed a novel interpretation of the 
RBM within a fully tractable and deterministic framework of learning 
and inference via TAP approximation. This deterministic construction
allows novel tools for scoring unsupervised models, investigation of
the memory of trained models, as well as allowing their efficient use
as structured joint priors for inverse problems.
While deterministic methods 
based on NMF for RBM training were shown to be inferior to CD-k in \cite{WH2002}, 
the level of approximation accuracy afforded by TAP finally makes the deterministic
approach to RBMs effective, as shown in the case of 
binary RBMs in \cite{GTZ2015}.

Additionally, our construction is generalized over the distribution of
both the hidden an visible units. This is unique to our work, as other
works propose unique training methods and models when changing the 
distribution of the visible units. For example, this can be seen
in the modified Hamiltonians used for real-valued data \cite{WMW2014,CIR2011}. 
This construction allows us to consider binary, real-valued,
and sparse real-valued datasets within the same framework. Additionally,
one can also consider other architectures by changing the distributions
imposed on the hidden unit. Here, we present experiments using only
binary hidden units, but one could also use our proposed framework
for Gaussian-distributed hidden units, thus mimicking a Hopfield
network \cite{Hop1982}. Or, also, sparse Gauss-Bernoulli distributed
hidden units could mimic the same functionality as that proposed by
the spike-and-slab RBM \cite{CBB2011}. We have left these 
investigations to further works on this topic.

Our proposed framework also offers a possibility to explore the 
statistical mechanics of these latent variable models at the level of
TAP approximation. Specifically, for a given statistical model of the 
weights $W$, both the cavity method and replica can begin to make 
predictions about these unsupervised models. 
Analytical understanding
of the complexity of the free energy landscape, and its transitions 
as a function of model hyper-parameters, can allow for a richer 
understanding of statistically optimal network construction for 
learning tasks. In the case of random networks, there has already been some 
progress in this area, as shown in \cite{TM2016} and \cite{BGS2016}. 
However, similar comprehensive studies conducted on learning in a realistic 
setting are still yet to be realized.
Finally, as our framework can be applied to deep Boltzmann machines 
with minimal alteration, it can also potentially lead to a 
richer understanding of deep networks and the role of hierarchy in 
regularizing the learning problem in high dimensionality.

\begin{acknowledgments}
This research was funded by
European Research Council under the European Union's $7^{th}$
Framework Programme (FP/2007-2013/ERC Grant Agreement 307087-SPARCS).
M.G. acknowledges funding from "Chaire de
recherche sur les mod\`eles et sciences des donn\'ees", Fondation CFM pour la Recherche-ENS.
\end{acknowledgments}
\vspace{1ex}

% % Appendices
% \newpage
% \pagebreak
\appendix
\section{Belief Propagation for Pairwise Models}
\label{sec:bp}
In order to estimate the derivatives of $\F$, we must first construct a BP 
algorithm on the factor graph representation given in Fig. \ref{fig:visFactorGraph}. 
We note that this graph, in terms of the variables $\mathbf{x}$, does not make
an explicit distinction between the latent and visible variables. We instead 
treat this graph in full generality so as to clarify the derivation and notation.
This graph corresponds to the following joint distribution over $\mathbf{x}$,
\begin{equation}
    P(\mathbf{x};W,\allparams) \propto e^{\sum_{(i,j)} \phi(x_i,x_j;\Wij) + \sum_i \phi(x_i;\theta_i)},
\end{equation}
where $\sum_{(i,j)}$ is a sum over the edges in the graph.
In the case of a Boltzmann machine, 
any two variables are connected via pairwise factors,
\begin{align}
\phi(x_i,x_j;\Wij) = e^{\Wij x_i x_j},
\end{align}
and all variables are also influenced by univariate factors written trivially as
$\phi(x_i;\theta_i) = P_i(x_i;\theta_i)$. 

A message-passing can be constructed on this factor graph by writing messages
from variables to factors and also from factors to variables. Since all 
factors are at most degree 2, we can write the messages for this system 
as variable to variable messages \cite{MM2009},
\begin{equation}
\msg_{i\to j}^{(t+1)}(x_i) =
    \frac{\phi(x_i;\theta_i)}{Z_{i\To j}}    
    \prod_{l \in \nbr{i}\except j }
    \integ{x_l} \phi(x_i,x_l;W_{il})~\msg^{(t)}_{l\To i}(x_l).
    \label{eq:motherMsg}
\end{equation}
Here, the notation $i\To j$ represents a message \emph{from} variable index $i$
\emph{to} variable index $j$, and $\nbr{i}\except j$ refers to all neighbors of 
variable index $i$ \emph{except} variable index $j$. We denote neighboring 
variables as those which share a pairwise factor. Finally, the super-scripts
on the messages refer to the time-index of the BP iteration, which implies the
successive application of \eqref{eq:motherMsg} until convergence on the set of
messages $\allmsg = \left\{ \msg_{i\To j}(x_i) : (i,j)\in E\right\}$, where
$E$ is the set of all pairs of neighboring variables. We also note the inclusion of the
message normalization term $Z_{i\To j}$ which ensures that all messages are 
valid PDFs. Additionally, it is possible to write the marginal beliefs at each
variable by collecting the messages from all their neighbors,
\begin{equation}
\marg^{(t)}_{\To i}(x_i) =
    \frac{\phi(x_i;\theta_i)}{Z_{\To i}}    
    \prod_{j \in \nbr{i}}
    \integ{x_j} \phi(x_i,x_j;\Wij)~\msg^{(t)}_{j\To i}(x_j).
    \label{eq:motherMarginal}
\end{equation}

Subsequently, the Bethe free energy can be written for a converged set of 
messages, $\allmsg^*$, according to \cite{MM2009} as

\begin{widetext}
\begin{equation}
\F_{\rm B}[\allmsg^*]
= 
    \sum_i \F_i[\allmsg^*]
    - \sum_{(i,j) \in E}
    \F_{(i,j)}[\allmsg^*]
=
    \sum_i \ln Z_{\To i}^*
    -
    \sum_{(i,j)} \ln \s{\int {\rm d}x_i~{\rm d}x_j~~
                     \msg_{i\To j}^*(x_i)~
                     \phi(x_i,x_j;\Wij)~
                     \msg_{j\To i}^*(x_j)},    
\label{eq:betheFE}                                     
\end{equation}
\end{widetext}
where $Z_{\To i}^*$ refers to the normalization of the
set of the marginal belief at site $i$ derived from  $\allmsg^*$.

Unfortunately, the message-passing of \eqref{eq:motherMsg} cannot be written as
a computable algorithm due to the continuous nature of the PDFs. Instead, we 
must find some manner by which to parameterize the messages. In the case of 
binary variables, as in \cite{GTZ2015}, each message PDF 
can be exactly parameterized by its expectation. However, for this general case
formulation, we cannot make the same assumption. Instead, we turn to 
\emph{relaxed} BP (r-BP) \cite{Ran2010}, described in the next section, which 
assumes a two-moment parameterization of the messages.

\section{r-BP for Pairwise Models}
\label{sec:rbp}
We will now consider one possible parametric approximation of the message
set via r-BP \cite{Ran2010}. This approach has also gone by a number of
different names in parallel re-discoveries of the approach, e.g.
moment matching \cite{OW2005} and non-parametric BP \cite{SIF2010}. In 
essence, we will be assuming that all messages $\allmsg$ can be 
well-approximated by their mean and variance, a Gaussian assumption. 
This approximation arises from a second-order expansion assuming small
weights $W_{ij}$. By making this assumption, we will ultimately be able
to close an approximation of the messages on their two first moments,
$a_{i\To j} \defas \an{x_i}_{\msg_{i\To j}}$ and 
$c_{i\To j} \defas \an{x_i^2}_{\msg_{i\To j}}  - \an{x_i}^2_{\msg_{i\To j}}$.

\subsection{Derivation via Small Weight Expansion}
Considering the marginalization taking place in \eqref{eq:motherMsg},
we will perform a second order expansion assuming that $W_{il}\To 0$. We start
by taking the Taylor series of the incoming message marginal for negligible 
weights,
\begin{align}
&\integ{x_l} e^{\Wil x_i x_l}\cdot\msg^{(t)}_{l\To i}(x_l) = \notag\\
  &\quad  1 + 
  \Wil\integ{x_l} \pde{\Wil}\s{e^{\Wil x_i x_l}}_{\Wil = 0}\msg^{(t)}_{l\To i}(x_l)
  \notag \\
  &\quad+ \frac{1}{2} \Wil^2 \integ{x_l} \pde{\Wil^2}\s{e^{\Wil x_i x_l}}_{\Wil = 0}\msg^{(t)}_{l\To i}(x_l)\notag \\ 
  &\quad+ O(\Wil^3).
\end{align}
Now, we approximate the series by dropping the terms less than $O(W_{il}^3)$. 
This approximation can be justified in the event that all weight values satisfy
$|W_{il}| < 1$. Identifying the integrals from the expansion as moments, we see
the following approximation
\begin{align}
  &\approx
  1 + x_i \Wil a_{l\To i}^{(t)} + \frac{1}{2} x_i^2 \Wil^2 \an{x_l^2}_{\msg_{l\To i}^{(t)}}.
\end{align}
However, we would like to write this approximation in terms of the central 
second moment. Through a second approximation that neglects $O(W_{il}^3)$ terms
we arrive at our desired parameterization of the incoming message marginalization
in terms of the message's two first \emph{central} moments,
\begin{align}
  &= \exp\br{ \log \s{
  1 + x_i \Wil a_{l\To i}^{(t)} + \frac{1}{2} x_i^2 \Wij^2 \an{x_l^2}_{\msg_{l\To i}^{(t)}}}}, \notag \\
  &\approx
  e^{ x_i \Wil a_{l\To i}^{(t)} + \frac{1}{2} x_i^2 \Wil^2 c_{l\To i}^{(t)} 
           + O(\Wil^3)}, \notag \\
  &\approx
  e^{ x_i \Wil a_{l\To i}^{(t)} + \frac{1}{2} x_i^2 \Wil^2 c_{l\To i}^{(t)}}.  
  \label{eq:smallw-expansion}         
\end{align}
We now substitute this approximation back into \eqref{eq:motherMsg} to get
\begin{align}
\msg_{i\to j}^{(t+1)}(x_i) \approx 
\widetilde{\msg}_{i\to j}^{(t+1)}(x_i) =
    \frac{\phi(x_i;\theta_i)}{Z_{i\To j}}    
    e^{ x_i B_{i\To j}^{(t)} - \frac{1}{2} x_i^2 A_{i\To j}^{(t)}},
    \label{eq:rbp_msg}
\end{align}
where
\begin{align}
  B_{i\To j}^{(t)} &\defas \sum_{l \in \nbr{i}\except j } \Wil a_{l\To i}^{(t)},
  \label{eq:Bij}\\
  A_{i\To j}^{(t)} &\defas - \sum_{l \in \nbr{i}\except j } \Wil^2 c_{l\To i}^{(t)}.
  \label{eq:Aij}
\end{align}
From here, we can see that we have now a set of closed equations due to the
dependence of $\mathbf{A}^{(t)}$ and $\mathbf{B}^{(t)}$ on the moments 
$\mathbf{a}^{(t)}$ and $\mathbf{c}^{(t)}$, and vice versa. 
The values of these moments can be written as a function of $\mathbf{A}^{(t)}$ 
and $\mathbf{B}^{(t)}$ which is dependent upon the form of the local potentials
$\phi(x_i;\theta_i)$, i.e. the prior distribution we assign to the variables
themselves,
\begin{align}
a^{(t)}_{i \To j}
  &= \fa\p{B^{(t-1)}_{i\To j},A^{(t-1)}_{i\To j} ; \theta_i},
  \label{eq:aij}\\
c^{(t)}_{i \To j}
  &= \fc\p{B^{(t-1)}_{i\To j},A^{(t-1)}_{i\To j} ; \theta_i},
  \label{eq:cij}
\end{align}
where
\begin{align}
\fa\p{B,A;\theta} 
  \defas& 
  \integ{x}x~\frac{\phi(x;\theta)}{Z}~e^{x B - \frac{1}{2}x^2 A},
  \label{eq:fa}\\
\fc\p{B,A;\theta} 
  \defas&
  -\fa\p{B,A;\theta}^2  + \notag\\
  \quad\quad& \integ{x}x^2~\frac{\phi(x;\theta)}{Z}~e^{x B - \frac{1}{2}x^2 A},
  \label{eq:fc}
\end{align}
and $Z$ is simply the normalization 
$\integ{x} \phi(x;\theta)~e^{x B - \frac{1}{2}x^2 A}$. The inferred marginal
distributions at each site can be calculated via the same functions, but 
instead using all of the incoming messages, i.e. 
$a^{(t)}_{i} = \fa\p{B^{(t)}_{\To i},A^{(t)}_{\To i}; \theta_i}$ and
$c^{(t)}_{i} = \fc\p{B^{(t)}_{\To i},A^{(t)}_{\To i}; \theta_i}$.
In Appendix \ref{sec:apdx_tunc_gaussian} we give the closed-forms of
these moment calculations for a few different choices of $\phi(x;\theta)$.

If one wants to obtain an estimate of the free energy for a given set of 
parameters $\allparams$, it is possible to iterate between 
\eqref{eq:Aij}, \eqref{eq:Bij} and \eqref{eq:aij}, \eqref{eq:cij} until, 
ideally, convergence. It is important to note, however, due to both the 
potentially loopy nature of the network as well as small-weight expansion, that
the BP iteration is not guaranteed to converge \cite{WT2003, MM2009}. 
Additionally, while we retain the time-indices in our derivation, it is not
clear whether one should attempt to iterate these message in fully sequential
or parallel fashion, or if some clustering and partitioning of the variables
should be applied to determine the update order dynamically. 

\subsection{r-BP Approximate Bethe Free Energy}
Additionally, we can write the specific form of the Bethe free energy under 
the r-BP two-moment parameterization of the messages. In this case, we can 
simply apply the small weight expansion of \eqref{eq:smallw-expansion} to the
Bethe free energy for pairwise models given in \eqref{eq:betheFE},
\begin{align}
\widetilde{\F}_{\rm B}[\widetilde{\allmsg}^*]
&=
    \sum_i \ln \widetilde{Z}_{\To i}^*\notag\\
    &\quad-
    \sum_{(i,j)} \ln \s{\int {\rm d}x_i~
                \widetilde{\msg}_{i\To j}^*(x_i)~
                 e^{x_i W_{ij} a_{j\To i}^*
                     + \frac{1}{2} x_i^2 W_{ij}^2 c_{j\To i}^*} }. 
\end{align}
Subsequently, using the final message definition given in \eqref{eq:rbp_msg},
we can see that 
\begin{align}
\widetilde{\F}_{\rm B}[\widetilde{\allmsg}^*]
&=
    \sum_i \ln \widetilde{Z}_{\To i}^*
    -
    \sum_{(i,j)} \bra{ \ln \widetilde{Z}_{\To i}^*
    - \ln \widetilde{Z}_{i\To j}^*},
\end{align}
which, correcting for double counting, can also be written as
\begin{equation}
\widetilde{\F}_{\rm B}[\widetilde{\allmsg}^*]
=
    \sum_i (1-\frac{1}{2} d_i) \ln \widetilde{Z}_{\To i}^*
    +
    \frac{1}{2}\sum_{i,j} \ln \widetilde{Z}_{i\To j}^*,
\end{equation}
where $d_i$ is the degree at site $i$, $|\partial_i|$.

\subsection{Enforcing Bounded Messages}
\label{sec:boundMsg}

While we 
write the r-BP messages \eqref{eq:rbp_msg} as though they are Gaussian 
distributions, this is a slight, since $A_{i\To j}^{(t)} \leq 0$ 
as $\Wil^2 c_{l\To i}^{(t)} \geq 0~~\forall i,l$. The implication of the 
expansion is that, in general, the messages are in fact unbounded. This 
unboundedness is a direct result of the form of the conventional RBM pairwise
factor, $e^{x_i \Wij x_j}$. 

There are a few avenues available to us to address these unbounded messages and 
produce a meaningful message-passing for generalized RBMs. Let us consider the 
cases for which the messages are unbounded \emph{given a specific variable 
distribution}. Assume that site $x_i$ is assigned a Gaussian prior,
$\phi(x_i;\theta_i = \bra{V_i,U_i}) \propto e^{x_i U_i - \frac{1}{2}x_i^2 V_i}$. 
In this case the r-BP message reads
\begin{align}
  \widetilde{\msg}_{i\to j}^{(t+1)}(x_i) 
    &=
    \frac{1}{\widetilde{Z}_{i\To j}}
    e^{ x_i (U_i + B_{i\To j}^{(t)}) - \frac{1}{2} x_i^2 (V_i + A_{i\To j}^{(t)})}.
\end{align}
In this case, the message is unbounded in the event that 
weighted sum of all incoming neighbor variances at $i$ exceeds the inverse 
variance of Gaussian prior on $x_i$,
\begin{align}
  V_i + A_{i\To j}^{(t)} &< 0,\\
  V_i &<  \sum_{l \in \nbr{i}\except j } \Wil^2 c_{l\To i}^{(t)},\\
  \sigma_i^2 &> \s{\sum_{l \in \nbr{i}\except j } \Wil^2 c_{l\To i}^{(t)}}^{-1}, 
\end{align}
where $\sigma_i^2$ is the variance of the Gaussian prior.
Said another way, this condition is telling us that when the message-passing 
starts to tell us that if the variance at $x_i$ is smaller than that of its
prior, the messages become unbounded and fail to be meaningful probability 
distributions, and our expansion fails. The implication is that the r-BP 
message passing should be utilized in contexts where there exists some,
preferably strong, evidence at each site, or the weights in $W$ should be 
sufficiently small. The stronger this local potential, or the smaller the
weights,
the more favorable the model is to the r-BP inference. This observation mirrors
those made in \cite{WT2003}, however, here the authors make the observation
that in this setting, BP based on small-weight expansion for \emph{binary} 
variables fails to converge. In our case, without taking some form of 
regularization, the inference fails entirely. Thus, large magnitude 
couplings $W_{il}$ must be backed with a high degree of evidence at 
site $i$ and $l$. This property could be utilized for the \emph{inverse}
learning problem, where one must learn the couplings $W$ given a dataset,
in order to constrain the learning to parameters which are amenable to the 
r-BP inference.

One direct manner to create probability distributions from otherwise unbounded
continuous functions is via truncation. Specifically, we enforce a 
non-infinite normalization factor by restricting the support of the distribution
to some subset of $\mathbb{R}$. In this case, just slightly violating the 
bounded condition above will induce a uniform message distribution over the
distribution support, while a strong violation will cause the distribution to
concentrate on the boundaries of the support. Another approach might simply
be to fix a hard boundary constraint on $A_{i\to j}$, thus never permitting 
unbounded messages to occur.

\section{Calculations for Specific Variable Distributions}
\label{sec:distributions}
%!TEX root = main.tex
\subsection{Truncated Gaussian Units}
\label{sec:apdx_tunc_gaussian}
In general, the truncated Gaussian is defined in the following manner,
\begin{equation}
\mathcal{TG}(x;\mu,\sigma^2,[\alpha,\omega]) =
  \frac{1}{\sqrt{2\pi\sigma^2}} \cdot 
  \frac{1}{ \cdf{\frac{\omega-\mu}{\sigma}} - \cdf{\frac{\alpha-\mu}{\sigma}} } \cdot
  e^{-\frac{(x-\mu)^2}{2\sigma^2}},
\end{equation}
where $\mu$ and $\sigma^2$ are the mean and variance of the original Gaussian prior to truncation and
the range $[\alpha,\omega]$ defines the lower and upper bounds of the truncation, 
$-\infty\leq\alpha<\omega\leq\infty$, and $\Phi\s{\cdot}$ is the CDF for the Normal distribution. 
To make things easier for us later, we will define the 
prior in a little bit of a different manner by making the following definitions,
\begin{equation}
  V \defas \frac{1}{\sigma^2}, \quad\quad
  U \defas \frac{\mu}{\sigma^2},
\end{equation}
and writing the distribution for the parameters 
$\theta = \bra{U,V,[\alpha,\omega]}$ as
\begin{equation}
\mathcal{TG}_{+}(x; \theta) 
  =
  \frac{\frac{2}{e^{\frac{U^2}{2V}}} 
  \sqrt{\frac{V}{2\pi}}   
  e^{-\frac{1}{2}V x^2 + Ux}}{\erf{\sqrt{\frac{V}{2}}(\omega - \frac{U}{V})} - \erf{\sqrt{\frac{V}{2}}(\alpha - \frac{U}{V})}}, 
\end{equation}
where $\erf{\cdot}$ is the error function and the last step follows from the identity $\cdf{x} = \frac{1}{2} + \frac{1}{2}\erf{\frac{x}{\sqrt{2}} }$. Here,
the subscript $+$ is used to indicate $V > 0$.

In the case that $V < 0$, we will write $\mathcal{TG}$ in terms of the 
imaginary error function, $\erfi{x} \defas -i \erf{i x}$, and noting that
for $V < 0$, $V = -1 \cdot |V|$,
\begin{equation}
\mathcal{TG}_{-}(x;\theta) 
  =
  \frac{\frac{2}{e^{\frac{U^2}{2V}}} 
  \sqrt{\frac{|V|}{2\pi}}   
  e^{-\frac{1}{2}V x^2 + Ux}}{
  \erfi{\sqrt{\frac{|V|}{2}}(\omega - \frac{U}{V})} - 
  \erfi{\sqrt{\frac{|V|}{2}}(\alpha - \frac{U}{V})}}.
\end{equation}
We use this negative variance version of the truncated Gaussian 
to handle the special case of $A+V < 0$ first detailed in Sec. 
\ref{sec:boundMsg}.

We now detail the computation of the partition 
and first two moments of the Gaussian-product distribution of $\mathcal{TG}$,

$Q(x;\theta,A,B) = \frac{1}{Z} \mathcal{TG}(x;\theta) e^{-\frac{1}{2}Ax^2 + Bx}$.
The calculation of the moments as a function of $A$ and $B$ will provide the
definitions of $\fa$ and $\fc$, while the calculation of the normalization $Z$ 
will provide terms necessary for both the computation of the TAP free energy
as well as the gradients necessary for learning $\theta$ during RBM training.

First, we will calculate the normalization of $Q(x;\theta,A,B)$ in terms of all free 
parameters. To do this,
consider the truncated normalization of the following product of Gaussians,
\begin{equation}
Z_{Q} = \frac{1}{Z_{\mathcal{TG}}}\integl{x}{\alpha}{\omega} e^{-\frac{1}{2}(A+V)x^2 + (B+U)x},
\end{equation}
where $Z_{\mathcal{TG}}$ is defined such that 
$\mathcal{TG}(x;\theta) = \frac{1}{Z_{\mathcal{TG}}} e^{-\frac{1}{2}Vx^2 + Ux}$.
We will need to make note of the special case of $A + V < 0$, thus,
% \begin{widetext}
\begin{equation}
\label{eq:zq-tg}
Z_{Q} =  
  \frac{1}{Z_{\mathcal{TG}}}
  \sqrt{\frac{\pi}{2|A+V|}} e^{\frac{(B+U)^2}{2(A+V)}}
  \times\left\{
\begin{array}{lr}
    d_+,   & A+V > 0\\
    d_-,  & A+V < 0
\end{array}
\right. ,
\end{equation}
where $d_+ \defas \erf{h_{\omega}} - \erf{h_{\alpha}}$,
$d_- \defas \erfi{h_{\omega}} - \erfi{h_{\alpha}}$,
and $h_{x} \defas \sqrt{\frac{|A+V|}{2}}\p{x - \frac{B+U}{A+V}}$.

Since $Q(x;\theta,A,B)$ is simply a truncated Gaussian with updated
parameters, the first moment is given according to the well 
known truncated 
Gaussian expectation. While this expectation is usually written in
terms of a mean and variance of the \emph{un}-truncated Gaussian
distribution, and for the case of positive mean, we will instead 
write the expectation in terms of the exponential 
polynomial coefficients and note the special case $A + V < 0$,
\begin{align}
\fa(B,A;\theta) &=
\frac{B+U}{A+V} + \sqrt{\frac{2}{\pi |A+V|}} \notag \\
&\quad \times\left\{
  \begin{array}{lr}
  \frac{e^{-h_{\alpha}^2} - e^{-h_{\omega}^2}}{d_+}, &A+V > 0 \\
  \frac{e^{h_{\alpha}^2} - e^{h_{\omega}^2}}{d_-}, &A+V < 0
  \end{array}
\right. .
\end{align}
 
Next, we will write the variance of $Q(x;\theta,A,B)$ as a function
of $A$ and $B$. As earlier, since $Q$ has the specific form of a 
truncated Gaussian distribution we can utilize the well-known 
variance formula for such a distribution. As in the case of $\fa$, we
modify this function for the special case of $A + V < 0$. Specifically,
\begin{align}
  &\fc(B,A;\theta) =
  \frac{1}{A+V}
  - \p{\fa(B,A;\theta) - \frac{B+U}{A+V}}^2 \notag \\
  &\quad\quad+ \frac{2}{\sqrt{\pi}(A+V)}
  \times
  \left\{
    \begin{array}{lr}
      \frac{h_{\alpha}e^{-h_{\alpha}^2} - h_{\omega}e^{-h_{\omega}^2}}{d_+}, &A+V > 0 \\
      \frac{h_{\alpha}e^{h_{\alpha}^2} - h_{\omega}e^{h_{\omega}^2}}{d_-}, &A+V < 0
    \end{array}
  \right. .
\end{align}

\subsubsection{Gradients of the Log-likelihood}
To determine the gradients of the log-likelihood w.r.t. the model parameters, 
it is necessary to calculate the gradients of both $\ln \mathcal{TG}_+(x;\theta)$
and $\ln Z_Q$ in terms of the 
distribution parameters $U$ and $V$. We assume that the boundary terms 
$\alpha$ and $\omega$ remain fixed.
Since both of these distributions are truncated Gaussians, we can treat them
both in terms of the derivatives of the log normalization of a general-case
truncated Gaussian, 
\begin{equation}
  \ln Z_{\mathcal{TG}} =
  -\frac{U^2}{2V} + \frac{1}{2}\ln |V| - \ln d_{+/-}.
\end{equation}
For Boltzmann measures of the given quadratic form, we know that 
$\pde{U}{\ln Z_{\mathcal{TG}}} = \ang{x}_{\mathcal{TG}}$ and
$\pde{V}{\ln Z_{\mathcal{TG}}} = -\frac{1}{2} \ang{x^2}_{\mathcal{TG}}$. From
these relations, we can write the necessary derivatives. For the case of 
$\ln \mathcal{TG}_+(x;\theta)$ we have
\begin{align}
\pde{U}{\ln \mathcal{TG}_+(x;\theta)} &= x - \pde{U}\ln Z_{\mathcal{TG}}
=x - \ang{x}_{\mathcal{TG}},
\end{align}
and 
\begin{align}
\pde{V}{\ln \mathcal{TG}_+(x;\theta)} 
  &= -\frac{1}{2}x^2 - \pde{V}\ln Z_{\mathcal{TG}},\notag\\
  &= -\frac{1}{2}\p{x^2 - \ang{x^2}_{\mathcal{TG}}},
\end{align}
which we can see as just the difference between the data and the moments of 
the truncated Gaussian distribution. 

Next, the derivatives of $Z_Q$, for fixed $A$ and $B$, can be written in terms 
of $\fa$ and $\fc$, the moments of $Q$,
\begin{align}
  \pde{U}\ln Z_Q &= f_a(B,A;\theta) - \ang{x}_{\mathcal{TG}},\\
  \pde{V}\ln Z_Q &= -\frac{1}{2}\p{f_c(B,A;\theta) + f_a(B,A;\theta)^2 
                                   - \ang{x^2}_{\mathcal{TG}}}.
\end{align}

Finally, we are ready to write the gradients of the log-likelihood 
to be used for both hidden and visible updates with the truncated Gaussian 
distribution. For a given set of mini-batch data indexed by $m \in \bra{1,\dots,M}$,
and a number of TAP solutions indexed by $p \in \bra{1,\dots,P}$, 
the gradients of a visible variable are given by
\begin{align}
\Delta U_{i} &=
  \frac{1}{M} \sum_m \p{x_i^{(m)} - \ang{x_i}_{\mathcal{TG}}}\notag\\
  &\quad-\frac{1}{P} \sum_p \p{a_i^{(p)} - \ang{x_i}_{\mathcal{TG}}},\notag\\
  &= \ang{x_i}_M - \ang{a_i}_P ,
\end{align}
and
\begin{align}
\Delta V_{i} &=
-\frac{1}{2}\s{\ang{x_i^2}_M - \ang{c_i + a_i^2}_P },
\end{align}
where $\ang{\cdot}_M$ and $\ang{\cdot}_P$ are averages over the mini-batch and
TAP solutions, respectively.

For updates of hidden side variables using the truncated Gaussian distribution
we have the following gradients for updating their parameters,
\begin{equation}
  \Delta U_j = \ang{\widetilde{a}_j}_M - \ang{a_j}_P
\end{equation}
and
\begin{equation}
  \Delta V_j = -\frac{1}{2}
  \s{\ang{\widetilde{c}_j + \widetilde{a}_j^2}_M - \ang{c_j + a_j^2}_P},
\end{equation}
where the moments $\widetilde{a_j}^{(m)} \defas \fa(\sum_i \Wij x_i^{(m)},0;\theta_j)$ and 
$\widetilde{c_j}^{(m)} \defas \fc(\sum_i \Wij x_i^{(m)},0;\theta_j)$ are 
defined for convenience. Using 
these gradients, we can now update the local biasing distributions for 
truncated Gaussian variables.

\subsubsection{Numerical Considerations}
Using the truncated Gaussian prior comes with a few numerical issues which
must be carefully considered. While we have already addressed the cases
when $A+V<0$, we have not addressed the case where 
\begin{equation}
  \left| \frac{B+U}{A+V}\right| \gg |\omega|, |\alpha|.
\end{equation}
One can see how this complicates matters by observing the term
$\frac{1}{d_{+/-}}$ which occurs in both the first and second moment
computations.
When the magnitude of the limits $\omega$ and $\alpha$ become vanishingly 
small in comparison to the scaled joint mean term, then we see that
$d_{+/-} \rightarrow 0$. However, the numerators of both $\fa$ and 
$\fc$ also go to zero, which implies that we may be able to find some 
method of approximation to find estimates of these moments without,
up to numerical precision, dividing by zero.

In order to handle this eventuality in our implementation, we make a 
Taylor series expansion of the $\frac{1}{d_{+/-}}$ term in the 
following way. First, we note that $d_{+/-}$ has the form
$\erf{\eta(\omega - \mu)} - \erf{\eta(\alpha - \mu)}$, which can 
be rewritten as $\erf{z} - \erf{z - \epsilon}$ for 
$z \defas \eta(\omega - \mu)$ and 
$\epsilon \defas \eta(\alpha - \omega)$ is a multiple of the difference
between the two boundaries of the truncated Gaussian distribution. 
Since we wish to consider the case that $\mu\rightarrow\infty$,
we will take the Taylor expansion centered at $z = \infty$, since the 
value of $\mu$ dominates. From here, we find that the following approximation
works well in practice,
\begin{align}
  \frac{e^{h_\alpha^2} - e^{h_\omega^2}}{d_+}
  &\overset{n}{\approx}
  \frac{\mathcal{S}^{(n)}(h_\omega^2)}{(\alpha - \frac{B+U}{A+V})(1 - e^{h_\alpha^2-h_\omega^2})} \notag\\
  &\quad+ \frac{\mathcal{S}^{(n)}(h_\alpha^2)}{(\omega - \frac{B+U}{A+V})(1 - e^{h_\omega^2-h_\alpha^2})},
\end{align}
where $\mathcal{S}^{(n)}(\cdot)$ is the $n$-term power series representation of the
error function. In our experiments, we use $n = 11$. A similar approximation
can be used for the variances in the same situation. While this approximation 
could potentially be computationally costly for large $n$, 
we note that it is only used for updates on 
variables for which a small value of $d_{+/-}$ has been detected.

\subsection{Truncated Gauss-Bernoulli Units}
\label{sec:apdx_trunc_gb}

For truncated Gauss-Bernoulli distributed units, we form the distribution as a
mixture between a delta function and $\mathcal{TG}$, with an extra term $(1-\rho)$
which controls the density at $x = 0$, thus, for 
$\theta = \bra{\rho,U,V,\s{\alpha,\omega}}$ we have 
\begin{align}
  \mathcal{TGB}(x;\theta) 
  &\defas 
  (1 - \rho)\delta(x) + \rho \mathcal{TG}(x;\theta)\\
  &=
  \frac{1}{Z_{\mathcal{TGB}}}
  \p{(1-\rho)\delta(x) + \rho e^{-\frac{1}{2}Vx^2 + Ux}}
\end{align}
where 
$Z_{\mathcal{TGB}} \defas (1-\rho) + \rho Z_{\mathcal{TG}}$ and
$\delta(x)$ is the Dirac delta function such that $\delta(0) = 1$ and
$0$ everywhere else. By using a construction such that the truncation is done 
on the Gaussian mode alone, and not across the entire distribution, we can 
easily write the necessary functions of this distribution in terms of the values
we have already calculated in Appendix \ref{sec:apdx_tunc_gaussian} as long as
$0 \in \s{\alpha,\omega}$. 
Additionally, it will be useful for our calculations to define the probability
of $x$ to be non-zero according to $\mathcal{TGB}$,
\begin{equation}
  {\rm P}\s{x \neq 0} 
  = \frac{\rho Z_{\mathcal{TG}}}{Z_\mathcal{TGB}}.
\end{equation}

We now continue as in the previous appendices and write the normalization and
first two moments of the Gaussian product distribution 
$Q(x;\theta,A,B) = \frac{1}{Z_Q} \mathcal{TGB}(x;\theta)e^{-\frac{1}{2}Ax^2 + Bx}$. First, the normalization can be written simply as a function of the truncated
Gaussian normalization modified by $A$ and $B$, as in \eqref{eq:zq-tg},
\begin{equation}
  Z_Q = (1-\rho) + \rho Z_{Q,\mathcal{TG}}.
\end{equation}
Next, the first moment of $Q$ can be found by recalling the relation 
$\pde{U}\ln Z_{Q} = \ang{x}_Q$ and consequently, that 
$\pde{U} Z_{Q} = Z_Q \ang{x}_Q$. Thus,
\begin{equation}
\fa(B,A;\theta) 
= \fa^{\mathcal{TG}}(B,A;\theta)\cdot{\rm P}\s{x\neq 0},
\end{equation}
where the non-zero probability is calculated according to $A$ and $B$.
For the second moment, we note that $\pde{V}Z_Q = -\frac{1}{2}Z_Q \ang{x^2}_{\mathcal{TG}}$ to find
\begin{align}
  \fc(B,A;\theta) 
    &= {\rm P}\s{x\neq 0}\cdot
       \p{\fa^{\mathcal{TG}}(B,A;\theta)^2 + \fc^{\mathcal{TG}}(B,A;\theta)}
       \notag \\
    &\quad - \fa(B,A;\theta)^2,\notag\\
    &= 
    \p{{\rm P}\s{x\neq 0} - {\rm P}\s{x\neq 0}^2} 
    \fa^{\mathcal{TG}}(B,A;\theta) \notag\\
    &\quad + {\rm P}\s{x \neq 0} \fc^{\mathcal{TG}}(B,A;\theta).
\end{align}

Next, we turn our attention to the log-likelihood 
gradients necessary for updating the  parameters $U$, $V$, and $\rho$ 
during training. First, we will look at the derivatives of $\ln\mathcal{TGB}$
required for updates on visible units. In order to calculate these derivatives,
we will split the log probability into two cases,
\begin{equation}
  \ln\mathcal{TGB}(x;\theta) 
  = 
    \left\{
    \begin{array}{lr}
    -\ln Z_{\mathcal{TGB}}, &x = 0 \\
    \ln\rho - \frac{1}{2}Vx^2 + Ux - \ln Z_{\mathcal{TGB}}, &x\neq 0
\end{array}
\right. 
\end{equation}
Consequently, the derivatives of the log probability can are written as
the following,
\begin{equation}
  \pde{U}\s{\ln\mathcal{TGB}(x;\theta)} =
  x - \ang{x}_{\mathcal{TGB}}, 
\end{equation}
and
\begin{equation}
  \pde{V}\s{\ln\mathcal{TGB}(x;\theta)} =
  -\frac{1}{2}\p{x^2 - \ang{x^2}_{\mathcal{TGB}}}. 
\end{equation}
The derivative w.r.t. $\rho$ is a bit more complicated, as we cannot use
the same identities. Additionally, we must also consider the two cases of
$x = 0$ and $x \neq 0$ separately. Thus,
\begin{align}
  \pde{\rho}\s{\ln\mathcal{TGB}(x;\theta)}
    &= 
    \left\{
    \begin{array}{lr}
      \rho Z_{\mathcal{TG}} - \frac{1}{\rho}P\s{x\neq 0}, &x = 0 \\
      \rho Z_{\mathcal{TG}} - \frac{1}{\rho}\p{P\s{x\neq 0}-1}, &x\neq 0
    \end{array}
    \right. ,
  \notag \\
  &=
  \rho Z_{\mathcal{TG}} - \frac{1}{\rho}\p{1 - \delta(x) - P\s{x\neq 0}},
\end{align}
which can be rewritten in the more concise form
\begin{equation}
  \pde{\rho}\s{\ln\mathcal{TGB}(x;\theta)}
  = 
  \frac{\delta(x) - \rho}{\rho(1-\rho)}
  \label{eq:pderho-concise}
\end{equation}
by noting the complement of the support probability 
$P[x = 0] = 1 - P[x\neq 0] = \frac{1-\rho}{Z_{\mathcal{TGB}}} = 
1 - \rho$ and making the appropriate substitution.

We now write the derivatives of the $\ln Z_Q$ in terms of 
$U$, $V$, and $\rho$. These take the same form as those written in 
Appendix \ref{sec:apdx_tunc_gaussian}. Subsequently, the equations for 
$\Delta U_i$, $\Delta V_i$, $\Delta U_j$, and $\Delta V_j$ all remain consistent,
just under the modification of all moments being taken w.r.t. the 
$\mathcal{TGB}$. We need only to write the gradients $\Delta \rho_i$, 
$\Delta \rho_j$. Starting from the log partition and applying the same identity
used to write Eq. \eqref{eq:pderho-concise}
\begin{equation}
  \pde{\rho}\s{\ln Z_Q} = 
  \frac{P_Q\s{x \neq 0} - \rho}{\rho(1-\rho)},
\end{equation}
which gives us the final gradients
\begin{equation}
\Delta\rho_i =  
  \frac{1}{\rho_i(1 - \rho_i)}\s{\ang{\delta(x_i)}_M - \ang{P_Q\s{x_i \neq 0}}_P}
\end{equation}
for visible units, and
\begin{equation}
\Delta\rho_j =  
  \frac{1}{\rho_j(1 - \rho_j)}\s{\ang{\widetilde{P}_Q\s{x_j \neq 0}}_M - \ang{P_Q\s{x_j \neq 0}}_P}
\end{equation}
for hidden units, where $\widetilde{P}_Q$ is calculated in the naive-mean-field 
manner described in Appendix \ref{sec:apdx_tunc_gaussian}, where 
$\widetilde{A}^{(m)}_j = 0$ and $\widetilde{B}^{(m)}_j = \sum_i \Wij x^{(m)}_i$.

\subsection{Binary Units}
\label{sec:apdx_binary}
We define the distribution for binary units to be the Bernoulli
distribution such that $x \in \bra{0,1}$ 
\begin{equation}
  \mathcal{B}(x;m) = (1-m)^{1-x}m^{x},
\end{equation}
where $m \defas {\rm Prob}[x = 1] = \ang{x}_{\mathcal{B}}$. We can also write
$\mathcal{B}$ as the Boltzmann distribution
\begin{equation}
  \mathcal{B}(x;U) = \frac{1}{Z_\mathcal{B}} e^{Ux}
\end{equation}
where $U\defas \ln\frac{m}{1-m}$ and $Z_{\mathcal{B}} \defas 1 + e^{U}$. Next,
we calculate the normalization and moments of the distribution 
$Q(x;A,B,U) = \frac{1}{Z_Q Z_{\mathcal{B}}} e^{-\frac{1}{2}Ax^2 + (B+U)x}$. 
For the normalization we have,
\begin{align}
Z_Q &= \frac{1}{Z_{\mathcal{B}}} 
       \sum_{x = \bra{0,1}} e^{-\frac{1}{2}Ax^2 + (B+U)x} \notag\\
    &= \frac{1}{Z_{\mathcal{B}}}\s{1 + e^{-\frac{1}{2}A + B + U}}.
\end{align}
Subsequently, for the moments of $Q$ we have
\begin{align}
  \fa(B,A;U) = \pde{U}\s{\ln Z_{\mathcal{B}}} =
  {\rm sigm}\p{U + B - \frac{1}{2}A},
\end{align}
where ${\rm \sigm}$ is the logistic sigmoid function. Subsequently, the 
variance for the binary unit can be calculated directly as 
\begin{equation}
  \fc(B,A;U) = \fa(B,A;U) - \fa(B,A;U)^2.
\end{equation}

Next, if we wish to define the learning gradients on $U$, we write the 
derivatives of $\ln B(x;U)$ and $\ln Z_Q$ w.r.t. $U$,
\begin{equation}
  \pde{U}{\ln B(x;U)} = \pde{U}\s{Ux - \ln Z_{\mathcal{B}}} = x - \ang{x}_{\mathcal{B}},
\end{equation}
and, for the log normalization,
\begin{equation}
  \pde{U}{\ln Z_{\mathcal{B}}} = \fa(B,A;U) - \ang{x}_{\mathcal{B}}.
\end{equation}
The resulting gradients for the distribution terms $U$ are 
\begin{equation}
  \Delta U_i = \ang{x_i}_M - \ang{a_i}_P
\end{equation}
for visible units, and
\begin{equation}
  \Delta U_j = \ang{\widetilde{a}_j}_M - \ang{a_i}_P
\end{equation}
for hidden units, where $\widetilde{a}_j \defas \fa(\sum_i \Wij x_i^{(m)},0;U)$.

\section{{Adaptive TAP}}
\label{sec:adatap}
When performing inference, one could employ instead of TAP a variant known
as adaptive TAP (or adaTAP) \cite{OW2001}, which gives in general more
accurate results, albeit being slower to iterate. We briefly investigate
here the performance of this method in the binary case.

The adaTAP algorithm is more generally presented without distinction
between visible and hidden variables. We thus write the algorithm for
generic weight matrix $J$ and bias vector $\mathbf{H}$, in practice here
defined by blocks
\begin{align*}
&J=\left( \begin{array}{cc}
            0 & W \\
            W^T & 0 \\
            \end{array}
    \right)  \in \mathbb{R}^{(N_v+N_h)\times (N_v+N_h)}
    \; , \\
&\mathbf{H} = \left( \begin{array}{c}
             \bf b \\
             \bf c 
             \end{array}
                 \right) \in \mathbb{R}^{(N_v+N_h)} \, ,
\end{align*}

The proposed implementation Alg. \ref{alg:rbmInfada} 
uses the recently introduced vector approximate message-passing (VAMP)
\cite{RSF2017} to find the adaTAP fixed points.  
After convergence, quantities with subscripts 1 and 2 are equal and 
identify with the outputs of the TAP inference algorithm Alg. \ref{alg:rbmInf}, 
\begin{align*}
&\bf{A_1} = \bf{A_2} = \left( \begin{array}{c}
           \bf{A^v}\\
            \bf{A^h}\\
            \end{array}
    \right)  \; ,
    &\bf{B_1} = \bf{B_2} = \left( \begin{array}{c}
           \bf{B^v}\\
            \bf{B^h}\\
            \end{array}
    \right)  
    \; , \\
&\mathbf{a_1} =\mathbf{a_2} = \left( \begin{array}{c}
             \bf a^v \\
             \bf a^h
             \end{array}
                 \right)  \, ,
&\mathbf{c_1} = \mathbf{c_2} = \left( \begin{array}{c}
             \bf c^v \\
             \bf c^h
             \end{array}
                 \right)  \, ,
\end{align*}
again compactly defined by blocks over the visible and hidden units.
Here $\mathbf{c_2}$ is defined as the diagonal of the $C_2$ matrix, which
gives an estimate of the correlation between different units and must
be computed at each step of the algorithm.
These quantities are then incorporated to our training algorithm Alg.
\ref{alg:rbmTrain}.

\begin{center}
\begin{figure}[ht]
\centering
  \includegraphics[width=0.4\textwidth]{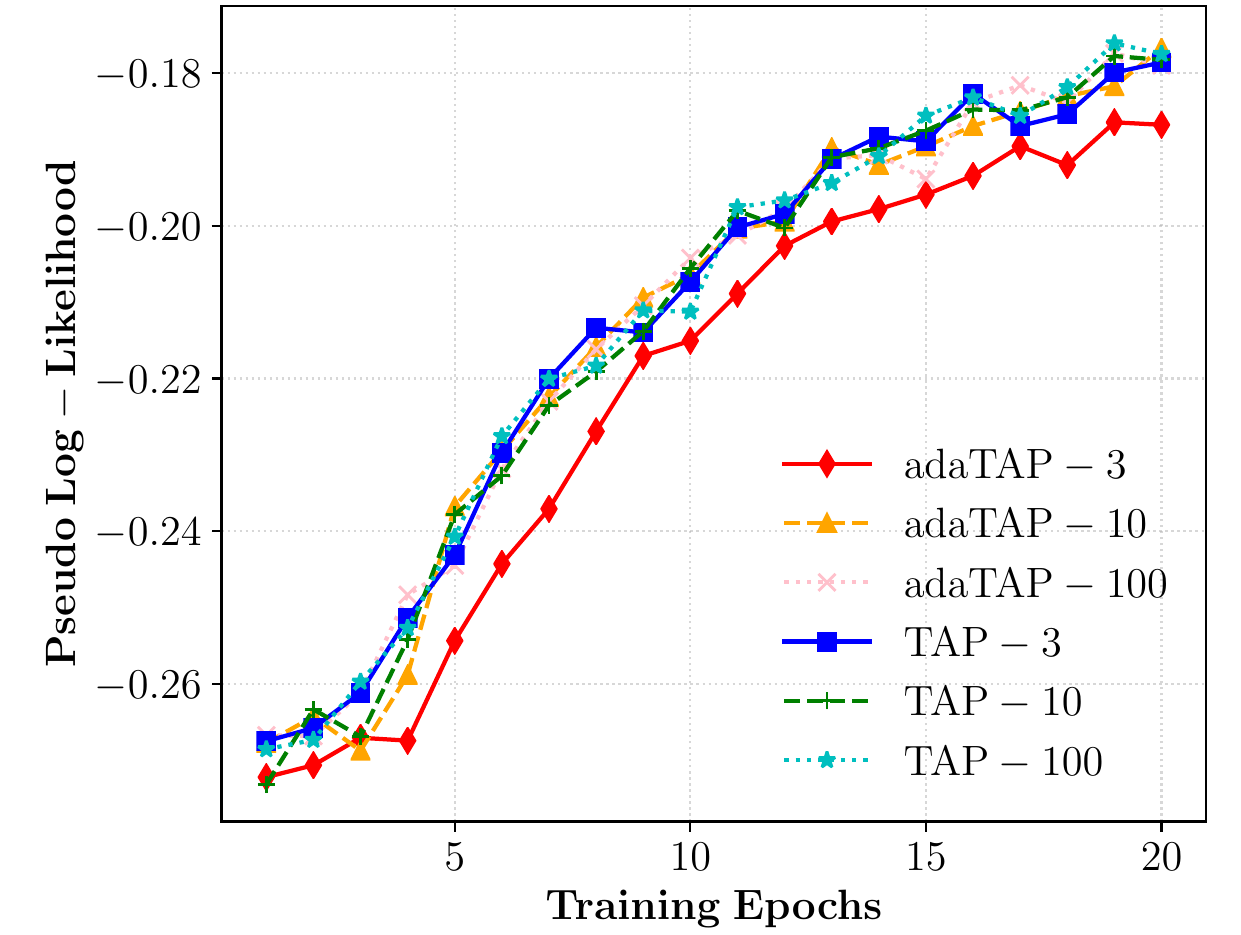}~
  \includegraphics[width=0.4\textwidth]{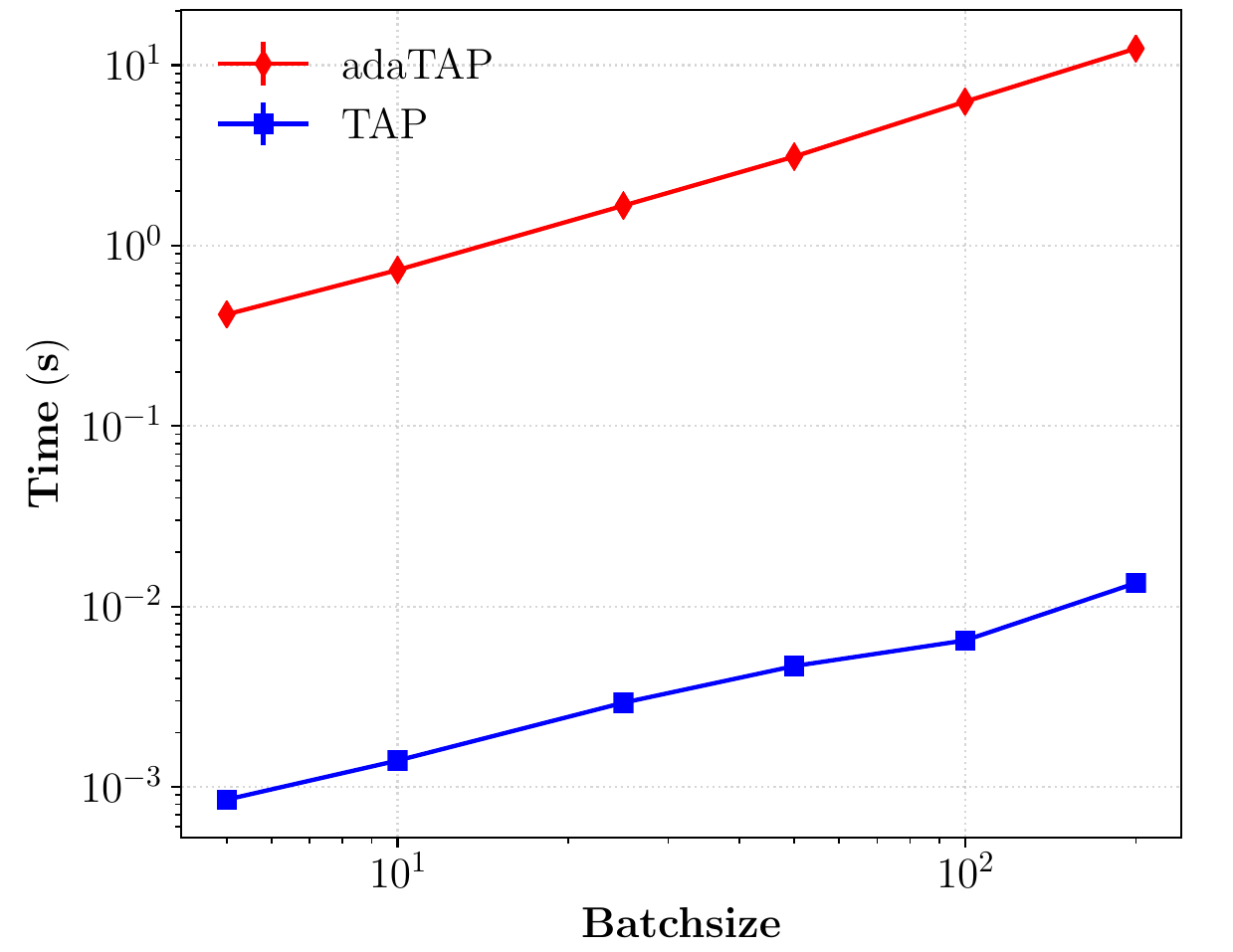}\\
  \caption{\label{fig:adatap} 
          \textbf{Left:} 
          Evolution of the pseudo-likelihood along the training of an
          RBM with 784 binary visible units and 500 binary hidden units.
          Training was performed using the 5000 first images of binarized
          MNIST, with a learning rate of 0.001 and batches of size 100.
          The different curves correspond to
          different strategies of estimation of the likelihood gradients, 
          either with TAP or adaTAP. Both
          algorithms were iterated for a fixed number of times (3, 10 and
          100). In all cases, a damping of 0.5 was used. 
          All methods yield comparable results in terms of
          training performance, except for adaTAP with only 3
          iterations, which shows poorer performance.
          \textbf{Right:} 
          Computation time for one iteration of the
          inference algorithm, as a function of batch size. % and number of hidden units.
          Time is reported in seconds for identical experimental settings. The
          need for a matrix inversion for each batch element makes VAMP 3
          orders of magnitude slower than TAP.}          
\end{figure}
\end{center}

The computational burden of Alg. \ref{alg:rbmInfada} lies in the matrix inversion
needed to evaluate $C_2$, which needs to be performed at each iteration.
In Fig. \ref{fig:adatap} (right), we compare the time needed to perform one
iteration of both the algorithms under identical experimental conditions.

% \newpage
% \onecolumngrid
% \begin{center}
% \begin{figure*}[b]
% \centering
%   \includegraphics[width=0.9\textwidth]{Comparing_trainings_with_adaTAP_and_TAP.pdf}\\
%   \caption{\label{fig:adatap} 
%           \textbf{Left:} 
%           Evolution of the pseudo-likelihood along the training of an
%           RBM with 784 binary visible units and 500 binary hidden units.
%           Training was performed using the 5000 first images of binarized
%           MNIST, with a learning rate of 0.001 and batches of size 100.
%           The different curves correspond to
%           different strategies of estimation of the likelihood gradients, 
%           either with TAP or adaTAP. Both
%           algorithms were iterated for a fixed number of times (3, 10 and
%           100). In all cases, a damping of 0.5 was used. 
%           All methods yield comparable results in terms of
%           training performance, except for adaTAP with only 3
%           iterations, which shows poorer performance.
%           \textbf{Right:} 
%           Computation time for one iteration of the
%           inference algorithm, as a function of batch size. % and number of hidden units.
%           Time is reported in seconds for identical experimental settings. The
%           need for a matrix inversion for each batch element makes VAMP 3
%           orders of magnitude slower than TAP.}          
% \end{figure*}
% \end{center}
% \twocolumngrid

This larger cost per-iteration should be compensated in principle by a more
accurate inference procedure. However, that does not seem to translate to
improvements in the training performance.
Fig. \ref{fig:adatap} (left)
presents a minimal test on 5000 MNIST training samples, where performances are 
reported in terms of the pseudo-likelihood.
We evaluate both algorithms for different numbers of iterations.
Results suggest that all strategies are roughly equivalent, except for running
adaTAP for a very small number of iterations, which always leads to a poorer result.

We thus conclude that, as far as proposing a tractable and efficient training
algorithm for RBMs, the TAP inference seems to serve the purpose more
appropriately. 

\section{Deep Boltzmann Machines}
\label{sec:dbm}
\subsection{Model and Inference}
It is possible to define as well \emph{deep} models of Boltzmann machines 
by considering several stacked hidden layers. 
These Deep Boltzmann Machines (DBM) \cite{SH2009} consist
in a straightforward extension of RBMs.
The distribution corresponding to a DBM with $L$ hidden layers indexed
by $l$ is 
% \onecolumngrid
\begin{widetext}
\begin{align}
P(\vecx, \vech^{(1)},..,\vech^{(L)};W^{(1)}, .., W^{(L)},\allparams) &= 
    \frac{1}{\Z\s{W^{(1)}, .., W^{(L)},\allparams}} e^{\sum_{i,j} x_i \Wij h^{(1)}_j + \sum_l \sum_{i,j} h^{(l)}_i \Wij^{(l)} h_j^{(l+1)} } 
      \notag\\ 
    &\quad\times \prod_{i}
     P_i^{\rm v}(x_i;\visparam{i}) 
    \prod_{l} \prod_{j} 
    P_j^{\mathrm{h} (l)}(h_j^{(l)};{\hidparam{j}}^{(l)}).
\label{eq:visgdbm}                          
\end{align}
\end{widetext}
% \twocolumngrid
Similarly to RBMs, the distribution of
visible variables is obtained by marginalizing out the latent variables, 
\begin{widetext}
\begin{align}
    P(\mathbf{x};W^{(1)}, .., W^{(L)},\allparams) = %\notag \\
    \int\p{\prod_l \prod_{j} {\rm d}h^{(l)}_j}   P(\vecx, \vech^{(1)},..,\vech^{(L)};W^{(1)}, .., W^{(L)},\allparams) ,
\end{align}
\end{widetext}
yielding the following log-likelihood
% \onecolumngrid
\begin{widetext}
\begin{align}
\ln P(\mathbf{x};W^{(1)}, .., W^{(L)},\allparams) &= 
    -\ln \Z\s{W^{(1)}, .., W^{(L)},\allparams}
    + \sum_{i}\ln P_i(x_i;\visparam{i})   \notag \\
    &\quad+ \ln \int \p{\prod_ l \prod_j {\rm d} h^{(l)}_j}
    e^{\sum_{i,j} x_i \Wij h^{(1)}_j + \sum_l \sum_{i,j} h^{(l)}_i \Wij^{(l)} h_j^{(l+1)} } 
    \prod_{l} \prod_{j} P_j^{\mathrm{h} (l)}(h_j^{(l)};{\hidparam{j}}^{(l)}) .
    \label{eq:genloglike_dbm}
\end{align}
\end{widetext}
% \twocolumngrid
The major difference between RBMs and DBMs lies in the 
complexity of evaluating the above expression. Whereas for RBMs
only the log-partition features a problematic multidimensional integral 
Eq. \eqref{eq:genloglike}, here both the log-partition and the last term 
are intractable. This additional complication carries through to the computation
of the gradients necessary for training, since the data-dependent 
term deriving from the last term of \eqref{eq:genloglike_dbm} is no longer tractable.

% \pagebreak
% \onecolumngrid
% \begin{center}
% \begin{figure*}[ht]
% \centering
%   \includegraphics[width=0.95\textwidth]{DBM_trainings.pdf}\\
%   \caption{\label{fig:dbmtraining} 
%            Training performances over 3000 training epochs for a 2-hidden
%            layer (left) and 3-hidden layer (right) deep Boltzmann machines
%            (DBMs) on the binarized MNIST datasets. Both models were
%            pretrained for 50 epochs with the same learning rate of 0.001.
%            The training performance is measured as the normalized (per unit) TAP
%            log-likelihood for the test images (blue) and the train images
%            (orange).}          
% \end{figure*}
% \end{center}
% \twocolumngrid

This intractability follows from the fact that hidden units in
neighboring layers are now connected to each other,
and are thus no longer conditionally independent. 
Interestingly, the first proposal
to deal with the data-dependent terms of DBMs consisted
in using a naive mean-field approximation \cite{SH2009}, 
while keeping a Monte Carlo based strategy to compute gradients deriving
from the log-partition. In this work, we propose instead to use the TAP 
approximation for both of them, hence improving on the NMF approximation 
and avoiding any sampling of the rather complicated RBMs.

The TAP equations related to the log-partition $\Z\s{W^{(1)}, .., W^{(L)},\allparams}$
follow directly from the general derivation of Sec. \ref{sec:tap} for fully connected models, 
with however a different weight matrix being used.
For instance, the effective weight matrix of a
DBM with 2 hidden layers is defined by blocks as
\begin{align}
W = \left( \begin{array}{ccc}
                    0 & W^{(1)} & 0                    \\
                    {W^{(1)}}^T & 0 &  W^{(2)}  \\
                    0 & {W^{(2)}}^T & 0 \\
                    \end{array}
                       \right). % \in \mathbb{R}^{(N_v+N_{h^{(1)}}+N_{h^{(2)}}) \times (N_v+N_{h^{(1)}}+N_{h^{(2)}})}.
\end{align}
Thus, implementing the GRBM inference algorithm Alg. \ref{alg:rbmInf} with the 
proper weights outputs TAP solutions $\{\mathbf{a}, \mathbf{c}, \mathbf{B}, \mathbf{A}\}$, 
each of them a vector with components corresponding to the different units in the DBM.

For the last term of \eqref{eq:genloglike_dbm}, we recognize 
the log-partition of a model closely related to the considered DBM, where 
visible units are not anymore variable but fixed, 
or \emph{clamped}, at values $x_i$, and the original interaction between visible and
first hidden layer units is replaced by an additional local field on each
$h^{(1)}_j$ equal to $\sum_i W_{ij}^{(1)} x_i$. 
Finally, under this simple modification of the Hamiltonian,
TAP equations follow again from the general derivation in Sec. \ref{sec:tap}. 
The resultant TAP solutions depending on data points $\vecx$ are said to be \emph{clamped} and
denoted as $\{\mathbf{\bar a(x)}, \mathbf{\bar c(x)}, \mathbf{\bar B(x)}, \mathbf{\bar A(x)}\}$. 

% Hence modifying 
% \begin{align}
% P_j^{\mathrm{h} (1)}(h_j^{(1)};{\hidparam{j}}^{(l)}) \leftarrow 
% e^{\sum_i W_{ij}^{(1)} x_i h_j^{(1)}}P_j^{\mathrm{h} (1)}(h_j^{(1)};{\hidparam{j}}^{(l)})
% \end{align}

\subsection{Training Algorithm and Experiments}
The gradients of the log-likelihood with respect to the model parameters 
$\allparams$ are similar to the RBM ones, given by \eqref{eq:paramGradVis}, \eqref{eq:paramGradHid}  
and \eqref{eq:Wgrad}. However, the first data-dependent term cannot be 
analytically computed anymore, and we use the clamped TAP solutions to approximate it.
The second term is evaluated using the data-independent TAP solutions,
similarly to our strategy for RBMs. 
The corresponding expressions of the gradients are
\begin{align}
\Delta \visparam{i} \approx
  \frac{1}{M} &\sum_m \pde{\visparam{i}}\s{\ln P_i^{\rm v}(x_i^{(m)};\visparam{i})} \notag\\
  -
  \frac{1}{K} &\sum_k \pde{\visparam{i}}\s{\ln Z_i^{\rm v}(B_{i,k}^{\rm v}, A_{i,k}^{\rm v}; \visparam{i})}\, , \\
\Delta {\hidparam{j}}^{(l)} \approx 
  \frac{1}{M} &\sum_m  \pde{{\hidparam{j}}^{(l)} }\Big[ \notag \\
  						& \quad \quad \quad\ln Z_j^{\rm h}
  					(\bar B_{j}^{\rm h}(\vecx^{(m)}) , \bar A_{j}^{\rm h}(\vecx^{(m)}); {\hidparam{j}}^{(l)} 
  						)\Big]\notag\\
  -
  \frac{1}{K} &\sum_k \pde{{\hidparam{j}}^{(l)} }\s{\ln Z_j^{\rm h}(B_{j,k}^{\rm h}, A_{j,k}^{\rm h}; {\hidparam{j}}^{(l)} )},
\end{align}
\begin{align}
\Delta W^{(1)}_{ij} \approx
  \frac{1}{M}&\sum_{m}x_i^{(m)} \bar a_{j}^{\rm h^{(1)}} (\vecx^{(m)})\notag\\
  -
  \frac{1}{K}&\sum_{k} \bra{a_{i,k}^{\rm v} a_{j,k}^{\rm h^{(1)}}
                       + \Wij^{(1)} c_{i,k}^{\rm v} c_{j,k}^{\rm h^{(1)}}},\\
\Delta W^{(l)}_{ij} \approx
  \frac{1}{M}&\sum_{m} \left\{
                                 \bar  a_{i}^{\rm h^{(l-1)}}(\vecx^{(m)})  \bar a_{j}^{\rm h^{(l)}} (\vecx^{(m)}) \right. \notag\\
                                & \left. \quad \quad + \Wij^{(l)} \bar c_{i}^{\rm h^{(l-1)}}(\vecx^{(m)})  \bar c_{j}^{\rm h^{(l)}} (\vecx^{(m)}) 
                                \right\}
                                \notag\\
  -
  \frac{1}{K}&\sum_{k} \bra{a_{i,k}^{\rm h^{(l-1)}} a_{j,k}^{\rm h^{(l)}}
                       + \Wij^{(l)} c_{i,k}^{\rm h^{(l-1)}} c_{j,k}^{\rm h^{(l)}}} \text{ for } l\geq2 \, .
\end{align}

These expressions can be plugged to a gradient ascent algorithm,
as in the RBM training algorithm Alg. \ref{alg:rbmTrain}. 
Nevertheless, this simple strategy of simultaneous training of
all the parameters of the model (\emph{joint} training) usually fails 
as the magnitude of weights of deep layers typically remains
very small and the model eventually resembles a mere RBM.
Several regularizations have been proposed to tackle this
well-known problem of DBM training \cite{SH2009,SL2010,Montavon2012,Desjardins2012,Melchior2016}. 
In our experiments, we used a greedy layerwise \emph{pre}-training \cite{SH2009}, 
which consists in computing a meaningful initialization of the weights
by training the RBMs layer-by-layer, before performing the joint training.
The complete algorithm is described in Alg. \ref{alg:dbmTrain}.

\begin{center}
\begin{figure}[hb]
\centering
  \includegraphics[width=0.4\textwidth]{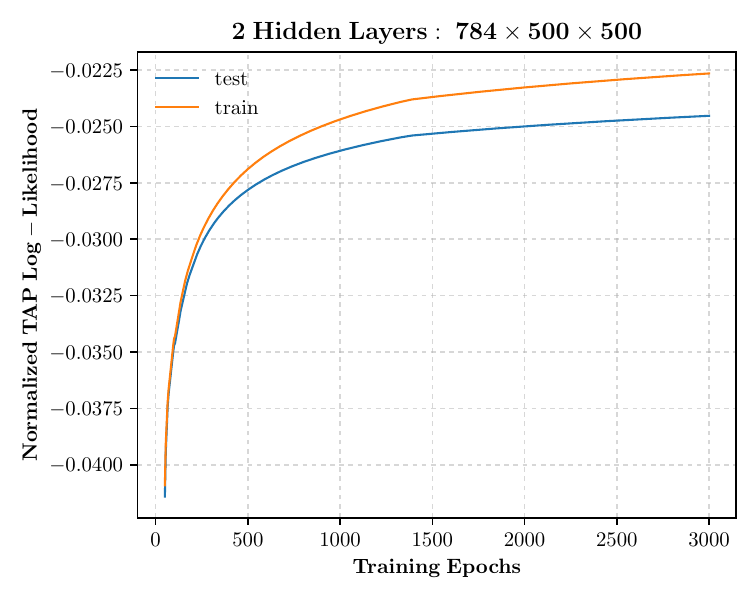}~
  \includegraphics[width=0.4\textwidth]{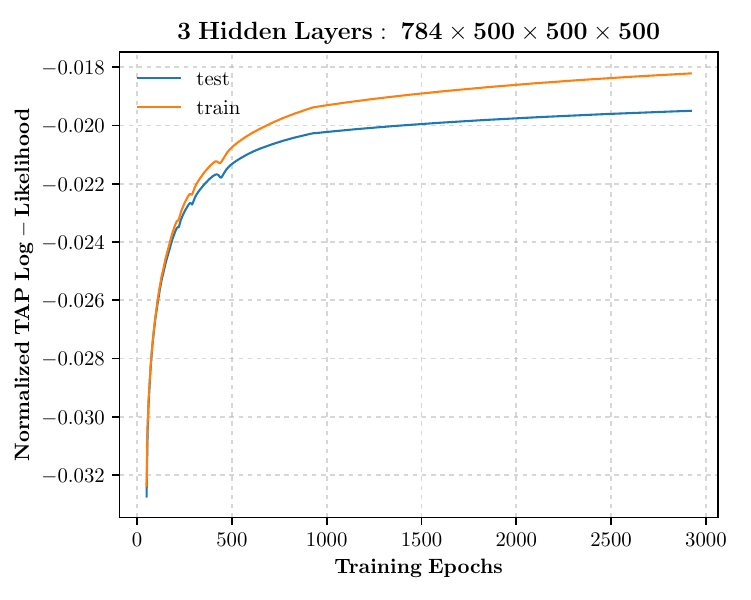}\\
  \caption{\label{fig:dbmtraining} 
           Training performances over 3000 training epochs for a 2-hidden
           layer (\textbf{left}) and 3-hidden layer (\textbf{right}) deep Boltzmann machines
           (DBMs) on the binarized MNIST datasets. Both models were
           pretrained for 50 epochs with the same learning rate of 0.001.
           The training performance is measured as the normalized (per unit) TAP
           log-likelihood for the test images (\textit{blue}) and the train images
           (\textit{orange}).}          
\end{figure}
\end{center}

Fig. \ref{fig:dbmtraining} shows the evolution of the TAP log-likelihood for a 2-hidden layer and a 3-hidden layer DBMs, trained with the above described algorithm.

%%% Inserting appendix figures here because im losing my mind: thanks RevTeX
% \clearpage
\begin{figure}
\begin{algorithm}[H]
  \caption{AdaTAP inference for binary-binary RBMs \label{alg:rbmInfada}}
  \begin{algorithmic}
    \STATE \emph{Input}: $J$, $\mathbf{H}$
    \STATE \emph{Initialize}: $t=0$, $\mathbf{A_1}^{\rm (0)}$, $\mathbf{B_1}^{\rm (0)}$
    \REPEAT
      \STATE \emph{Prior Updates}\Vhrulefill
      \STATE $a_{1,i}^{(t+1)} = \sigm(B_{1,i}^{(t)} - A_{1,i}^{(t)}/2)$ 
      \STATE $c_{1,i}^{(t+1)} = a_{1,i}^{(t+1)} (1-a_{1,i}^{(t+1)})$
      \STATE $A_{2,i}^{(t+1)} = 1/c_{1,i}^{(t+1)} - A_{1,i}^{(t+1)}$
      \STATE $B_{2,i}^{(t+1)} = 1/(1-a_{1,i}^{(t+1)} ) - B_{1,i}^{(t+1)}$
      \STATE 
      \STATE \emph{Interaction Updates}\Vhrulefill
      \STATE $C_2^{(t+1)} = (\operatorname{diag}(A_2^{(t + 1)})-J)^{-1}$
      \STATE $a_{2,i}^{(t+1)} = \displaystyle \sum_j \left(C_2^{(t+1)}\right)_{ij} \left(B_{2,j}^{(t+1)} + H_j \right)$
      \STATE $A_{1,i}^{(t+1)} = 1/\left(C_2^{(t+1)} \right)_{ii}  - A_{2,i}^{(t+1)}$
      \STATE $B_{1,i}^{(t+1)} = a_{2,i}^{(t+1)}/\left(C_2^{(t+1)} \right)_{ii} - B_{2,i}^{(t+1)}$
      \STATE $t = t+1$
    \UNTIL{Convergence}
  \end{algorithmic}
\end{algorithm}
\end{figure}
\newpage
\pagebreak
\begin{figure}
\begin{algorithm}[H]
  \caption{GDBM Training \label{alg:dbmTrain}}
  \begin{algorithmic}
    \STATE \emph{Input}: $\mathbf{X}$, $T_{\rm pre train}$,$T_{\rm joint \; train}$, $M$, $K$, $R(\cdot)$
    \STATE \emph{Pretraining}\Vhrulefill
    \STATE $W^{(1)}, \allparams^{\rm v} , \allparams^{\mathrm{h}^{(1)}} \leftarrow $ Alg. \ref{alg:rbmTrain}($\mathbf{X}$, $T_{\rm pre train}$, $M$, $K$, $R(\cdot)$)
      \FOR{All hidden layers $l\geq2$}
        \STATE $\mathbf{H}^{(l-1)} \sim P\p{\vech^{(l)} | \mathbf{H}^{(l-1)}, ... , \mathbf{H}^{1}, \mathbf{X}}$ 
        \STATE $W^{(l)}, \allparams^{\mathrm{h}^{(l)}} \leftarrow $ Alg. \ref{alg:rbmTrain}($\mathbf{H}^{(l-1)}$, $T_{\rm pre train}$, $M$, $K$, $R(\cdot)$)
        \ENDFOR
    \STATE \emph{Joint training}\Vhrulefill  
    \STATE \emph{Initialize:} t = 0
    \REPEAT
      \FOR{All mini-batches $\mathbf{X}^{B}$ of size $M$}
        \STATE $\mathbf{a}^{(t+1)}, \mathbf{c}^{(t+1)}, \mathbf{B}^{(t+1)}, \mathbf{A}^{(t+1)} \leftarrow$ Alg. \ref{alg:rbmInf}$(W^{(t)},\allparams^{(t)}) \times K$
        % \FOR{All data point $\vecx^{(m)}$}
            \STATE $\mathbf{\bar a}^{(t+1)}, \mathbf{\bar c}^{(t+1)}, \mathbf{\bar B}^{(t+1)}, \mathbf{\bar A}^{(t+1)} \leftarrow$ Alg. \ref{alg:rbmInf}$(\mathbf{X}^{B},W^{(t)},\allparams^{(t)}) $
        % \ENDFOR
        \STATE $W_{ij}^{(l),(t+1)} \leftarrow W_{ij}^{(l),(t)} + \gamma \Delta W_{ij} ^{(l),(t)} $ %+ \gamma\epsilon {\rm R}(W_{ij} ^{(l),(t)}) + \eta \Delta W_{ij}^{(l),(t-1)}$
        \STATE ${\hidparam{j}}^{(l),(t+1)} \leftarrow {\hidparam{j}}^{(l),(t)} + \gamma \Delta {\hidparam{j}}^{(l),(t)}$
        \STATE ${\visparam{i}}^{,(t+1)} \leftarrow {\visparam{i}}^{,(t)} + \gamma \Delta {\visparam{i}}^{,(t)}$        
      \ENDFOR
      \STATE $t \leftarrow t+1$
    \UNTIL{$t > T_{\rm joint \; train}$}
  \end{algorithmic}
\end{algorithm}
\end{figure}

\clearpage

\onecolumngrid

\twocolumngrid

\section{Comparison of the TAP log-likelihood with other surrogates}

As detailed in the main text, we consider the TAP log-likelihood as a
\emph{surrogate} to the true and intractable log-likelihood, as there are no guarantees to
how close it lies to the true value, nor does it provide any bound on the value of the
true log-likelihood -- as discussed in Section \ref{subsec:tapsolutions}, 
the TAP estimate may fall either above or below. Surrogates, such as the 
pseudo-likelihood \cite{besag1975statistical}, are widely used in Boltzmann 
machine learning \cite{Tie2008}.As a surrogate for the true log-likelihood,
besides its fast convergence and deterministic calculation, the TAP 
log-likelihood possesses many interesting properties which we study in our 
experiments. For instance,
% Marylou: we do not show samples in our experiments, so I think we should put it this way.
% samples of the resulting generative model are qualitatively similar to those % in the training set.
TAP machines can be used for denoising, as demonstrated in Section \ref{subsec:denoising},
or as priors for other applications of statistical inference.

\begin{figure}
    \begin{center}
        \includegraphics[width=0.5\textwidth]{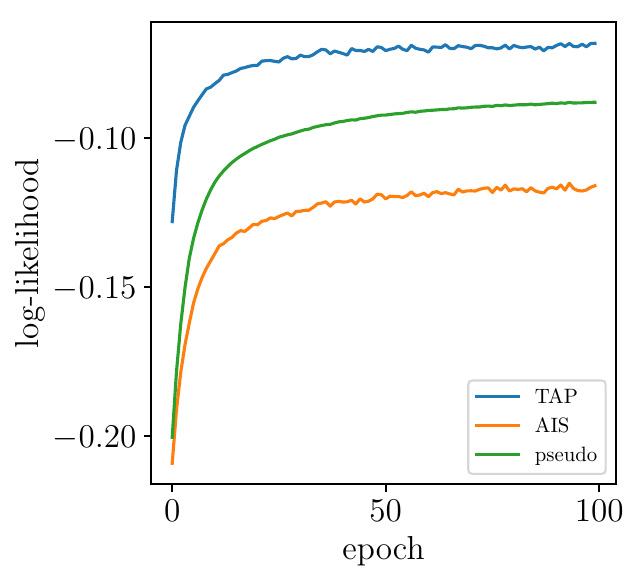}
        \caption{\label{fig:tap-ais-psuedo}%
        Comparison of the TAP log-likelihood to two other measures: the
        estimate provided by annealed importance sampling (AIS) and the
        (log) pseudo-likelihood (PL). All three measures are evaluated for the
        same sets of parameters, which correspond to the training of a
        binary-binary RBM using the TAP log-likelihood (see Fig.
        \ref{fig:bmnist-tapLL}(a)). 
        The AIS results are averaged over 100 runs. For each run,
        14,500 intermediate distributions are generated between the initial 
        and target distribution using the same schedule as in \cite{aismatlab}.
        The pseudo-likelihood is computed using a stochastic approximation
        over 100 of the 784 pixels.
        Even though the training is performed so as to maximize the TAP
        log-likelihood, the AIS and PL estimates also increase, 
        indicating the measures to be consistent.}
    \end{center}
\end{figure}

Another piece of supporting evidence for the quality of the TAP log-likelihood as a 
surrogate comes from studying how consistent it is with other estimators
and surrogates. We present two of them here:
the pseudo-likelihood (PL), and the log-likelihood estimate 
provided by annealed importance sampling (AIS) \cite{Nea2001}. 
For this experiment, we compare log-likelihood estimates from TAP, PL, and AIS
over the course of 100 epochs of training performed on a single binary-binary 
RBM with 500 hidden units using the binary-MNIST dataset as training data.
The training is performed by maximizing
the TAP log-likelihood, and each of the three approaches produces either an estimate
or a surrogate to the true log-likelihood at the end of each epoch (as in
Fig.~\ref{fig:bmnist-tapLL}(a)). We show the comparison of these surrogates
over training in Fig.~\ref{fig:tap-ais-psuedo}.

% Eric: This is just a placeholder for the final text.

All three measures have the same qualitative behavior as a function of epochs,
indicating they are consistent to each other in this particular experiment.
Notably, over training,
one does not need to accurately estimate the \emph{value} of the 
underlying true log-likelihood, but rather provide the same quality as a metric,
that is, if one set of model parameters has a larger log-likelihood than another
according to the true log-likelihood, then it should also be larger in the 
surrogate. As long as the ``landscape'' of the log-likelihood is preserved by
the surrogate, then it should be sufficient for training accurate models. 
While the true log-likelihood is unknowable in the context of this experiment, 
comparing to the AIS estimate shows similar growth in log-likelihood between 
the TAP surrogate and AIS. Thus, we can observe that maximizing
the TAP log-likelihood does indeed appear to produce models which improve
the true log-likelihood overtraining, indicating some correspondence between 
the TAP log-likelihood and the true log-likelihood, up to the AIS estimate.

% \begin{center}
% \begin{figure}[hb]
% \centering
%   \includegraphics[width=0.95\textwidth]{fig12.pdf}\\
%   \caption{\label{fig:aispseudo}}          
% \end{figure}
% \end{center}

\clearpage
% \vskip 0.2in
\small{ \bibliography{references} }

\end{document}